\documentclass[final]{clv2025}

\jvol{vv}
\jnum{nn}
\jyear{2025}

\pageonefooter{Action editor: \{action editor name\}. Submission received: DD Month YYYY; revised version received: DD Month YYYY; accepted for publication: DD Month YYYY.}

\usepackage[commandnameprefix=ifneeded,todonotes={textsize=small}]{changes}

\usepackage{amsmath}
\usepackage{booktabs}
\usepackage{longtable}
\usepackage{xcolor}
\usepackage{multirow}
\usepackage{wrapfig}
\usepackage{tabularx}
\usepackage{tipa}

\usepackage{tikz}
\usepackage{xspace}

\newcommand{\llama}{\textsc{LLaMA3.1-8B}\xspace}
\newcommand{\llamalarge}{\textsc{LLaMA3.1-70B}\xspace}
\newcommand{\anita}{\textsc{Anita-8B}\xspace}
\newcommand{\minerva}{\textsc{Minerva-7B}\xspace}

\newcommand{\ap}[1]{{\textcolor{black}{#1}}}
\newcommand{\mr}[1]{{\textcolor{black}{#1}}}

\newcommand{\posscite}[1]{\citeauthor{#1}’s (\citeyear{#1})}

\newcommand{\model}[1]{%
  \ifthenelse{\equal{#1}{anita}}{%
    \textsc{\anita}\xspace
  }{%
    \ifthenelse{\equal{#1}{llama8}}{%
      \textsc{\llama}\xspace
    }{%
      \ifthenelse{\equal{#1}{llama70}}{%
        \textsc{\llamalarge}\xspace
      }{%
        \ifthenelse{\equal{#1}{minerva}}{%
          \textsc{\minerva}\xspace
        }{%
        }%
      }%
    }%
  }%
}

\newcommand{\mnbox}[2]{
\tikzstyle{mybox} = [draw=black, fill=white, thick,
    rectangle, rounded corners, inner sep=5pt, inner ysep=10pt]
\tikzstyle{fancytitle} =[fill=gray!20, text=black, draw=black]

\begin{center}
\begin{tikzpicture}
\node [mybox] (box){%
    \begin{minipage}{0.6\textwidth}
{#1}
    \end{minipage}
};
\node[fancytitle, right=10pt] at (box.north west) {#2};
\end{tikzpicture}%
\end{center}
}

\runningtitle{Challenging the Abilities of LLMs in Italian}
\runningauthor{Nissim, Croce et al.}

\makeatletter
\def\authorfont{\normalsize\raggedright} 
\makeatother

\makeatletter
\def\affilfont{\small\raggedright} 
\makeatother

\begin{document}

\title{Challenging the Abilities of Large Language Models in Italian: a Community Initiative}

\author{
Malvina Nissim\thanks{Corresponding authors: \email{m.nissim@rug.nl}}$^{22}$,
Danilo Croce\thanks{Corresponding authors: \email{croce@info.uniroma2.it}}$^{27}$,
Viviana Patti\thanks{Corresponding authors: \email{viviana.patti@unito.it}}$^{31}$,
Pierpaolo Basile\thanks{Corresponding authors: \email{pierpaolo.basile@uniba.it}}$^{18}$,
Giuseppe Attanasio$^{11}$,
Elio Musacchio$^{18,26}$,
Matteo Rinaldi$^{31}$,
Federico Borazio$^{27}$,
Maria Francis$^{22,30}$,
Jacopo Gili$^{31}$,
Daniel Scalena$^{22,23}$,
Begoña Altuna$^{29}$,
Ekhi Azurmendi$^{29}$,
Valerio Basile$^{31}$,
Luisa Bentivogli$^{6}$,
Arianna Bisazza$^{22}$,
Marianna Bolognesi$^{19}$,
Dominique Brunato$^{10}$,
Tommaso Caselli$^{22}$,
Silvia Casola$^{15}$,
Maria Cassese$^{12}$,
Mauro Cettolo$^{6}$,
Claudia Collacciani$^{19}$,
Leonardo De Cosmo$^{2}$,
Maria Pia Di Buono$^{24}$,
Andrea Esuli$^{12}$,
Julen Etxaniz$^{29}$,
Chiara Ferrando$^{31}$,
Alessia Fidelangeli$^{19}$,
Simona Frenda$^{8}$,
Achille Fusco$^{13}$,
Marco Gaido$^{6}$,
Andrea Galassi$^{19}$,
Federico Galli$^{19}$,
Luca Giordano$^{24}$,
Mattia Goffetti$^{1}$,
Itziar Gonzalez-Dios$^{29}$,
Lorenzo Gregori$^{20}$,
Giulia Grundler$^{19}$,
Sandro Iannaccone$^{7}$,
Chunyang Jiang$^{9,21}$,
Moreno La Quatra$^{14}$,
Francesca Lagioia$^{4,19}$,
Soda Marem Lo$^{31}$,
Marco Madeddu$^{31}$,
Bernardo Magnini$^{6}$,
Raffaele Manna$^{24}$,
Fabio Mercorio$^{23}$,
Paola Merlo$^{9,21}$,
Arianna Muti$^{3}$,
Vivi Nastase$^{9}$,
Matteo Negri$^{6}$,
Dario Onorati$^{27}$,
Elena Palmieri$^{19}$,
Sara Papi$^{6}$,
Lucia Passaro$^{26}$,
Giulia Pensa$^{29}$,
Andrea Piergentili$^{6,30}$,
Daniele Potertì$^{23}$,
Giovanni Puccetti$^{12}$,
Federico Ranaldi$^{27}$,
Leonardo Ranaldi$^{27}$,
Andrea Amelio Ravelli$^{19,30}$,
Martina Rosola$^{17}$,
Elena Sofia Ruzzetti$^{27}$,
Giuseppe Samo$^{9}$,
Andrea Santilli$^{16}$,
Piera Santin$^{19}$,
Gabriele Sarti$^{22}$,
Giovanni Sartor$^{4,19}$,
Beatrice Savoldi$^{6}$,
Antonio Serino$^{23}$,
Andrea Seveso$^{23}$,
Lucia Siciliani$^{18}$,
Paolo Torroni$^{19}$,
Rossella Varvara$^{25}$,
Andrea Zaninello$^{6}$,
Asya Zanollo$^{13}$,
Fabio Massimo Zanzotto$^{27}$,
Kamyar Zeinalipour$^{28}$,
Andrea Zugarini$^{5}$}

\affilblock{
    \affil{Alpha Test S.r.l.} 
    \affil{ANSA} 
    \affil{Bocconi University} 
    \affil{European University Institute} 
    \affil{Expert.ai} 
    \affil{Fondazione Bruno Kessler} 
    \affil{Galileo Net} 
    \affil{Heriot-Watt University} 
    \affil{Idiap Research Institute} 
    \affil{ILC-CNR} 
    \affil{Instituto de Telecomunicações} 
    \affil{ISTI-CNR} 
    \affil{IUSS Pavia} 
    \affil{Kore University of Enna} 
    \affil{LMU Munich} 
    \affil{Sapienza University of Rome} 
    \affil{Universitat de Barcelona} 
    \affil{University of Bari Aldo Moro} 
    \affil{University of Bologna} 
    \affil{University of Florence} 
    \affil{University of Geneva} 
    \affil{University of Groningen} 
    \affil{University of Milano-Bicocca} 
    \affil{University of Naples "L'Orientale"} 
    \affil{University of Pavia} 
    \affil{University of Pisa} 
    \affil{University of Rome Tor Vergata} 
    \affil{University of Siena} 
    \affil{University of the Basque Country (UPV/EHU)} 
    \affil{University of Trento} 
    \affil{University of Turin} 
}

\maketitle

\begin{abstract}
The rapid progress of Large Language Models (LLMs) has transformed natural language processing and broadened its impact across research and society. Yet, systematic evaluation of these models, especially for languages beyond English, remains limited. "Challenging the Abilities of LAnguage Models in ITAlian" (CALAMITA) is a large-scale collaborative benchmarking initiative for Italian, coordinated under the Italian Association for Computational Linguistics. Unlike existing efforts that focus on leaderboards, CALAMITA foregrounds methodology: it federates more than 80 contributors from academia, industry, and the public sector to design, document, and evaluate a diverse collection of tasks, covering linguistic competence, commonsense reasoning, factual consistency, fairness, summarization, translation, and code generation. Through this process, we not only assembled a benchmark of over 20 tasks and almost 100 subtasks, but also established a centralized evaluation pipeline that supports heterogeneous datasets and metrics. We report results for four open-weight LLMs, highlighting systematic strengths and weaknesses across abilities, as well as challenges in task-specific evaluation. Beyond quantitative results, CALAMITA exposes methodological lessons: the necessity of fine-grained, task-representative metrics, the importance of harmonized pipelines, and the benefits and limitations of broad community engagement.
CALAMITA is conceived as a rolling benchmark, enabling continuous integration of new tasks and models. This makes it both a resource -- the most comprehensive and diverse benchmark for Italian to date -- and a framework for sustainable, community-driven evaluation. We argue that this combination offers a blueprint for other languages and communities seeking inclusive and rigorous LLM evaluation practices.
\end{abstract}

\section{Introduction and Motivation}
\label{sec:introduction}

Large Language Models (LLMs) have rapidly become a cornerstone of contemporary Natural Language Processing (NLP), demonstrating impressive abilities across a wide range of linguistic and more general downstream tasks. Concurrently, beyond their dominance in NLP, LLMs have now become pervasive in the daily life of many, and also in the research activities of scholars of a large variety of fields. 

Evaluating where these models excel and, crucially, where they fall short is therefore not only key to a better understanding of current developments in NLP and thus better guidance for future steps, but also a fundamental contribution to a more aware, responsible, and effective use of such models in research and society at large.

How to do model benchmarking in practice, however, remains a significant challenge. The question of which benchmarks to use to assess the acquired abilities of LLMs is far from trivial, especially when considering languages beyond English, Chinese, and programming code, which together dominate the training data of most modern LLMs (for instance, less than 0.1\% of training examples in models such as LLaMA2 are in Italian \citep{touvron2023llama}). The crux of the issue is not the inherent complexity of languages like Italian, but rather the limited international attention devoted to their evaluation, and the effort required to run language-specific campaigns. At large scale, the common trend to date for most languages other than English has been to rely on translated benchmarks \cite{thellmann2024towards,xuan2025mmlu}  or synthetic data approaches that fail to capture the true linguistic challenges faced by models in real-world scenarios and the true abilities of models in those languages \cite{plaza2024spanish,Wu2025TheBL}. Without native, authentic, human-crafted evaluation datasets, both fine-tuning and robust assessment of LLMs in underrepresented languages become academic exercises aimed at extracting leaderboard scores with no real understanding of the actual abilities of models in those languages and the cultures they express.

Faced with this state of affairs, and drawing on a long-standing tradition of Italian-focused evaluation campaigns \cite{AttardiBBCDMPSS15,basile2017evalita,passaro2020lessons,basile-etal-2022-italian}, the Italian NLP community has mobilized. Under the auspices of AILC (Associazione Italiana di Linguistica Computazionale)\footnote{\url{https://www.ai-lc.it/en/}}, the CALAMITA (\emph{Challenge the Abilities of Language Models in Italian}) initiative was launched as a collective, grassroots effort to create high-quality and broad-interest evaluation data for Italian LLMs and set up a dynamic benchmarking framework for the benefit of the research community at large. Leveraging AILC's expertise in NLP and evaluation campaigns, CALAMITA  
brings together the necessary sensitivity for language-sensitive challenges, the capacity to automate and centralize large-scale evaluation, and the access to national high-performance computing resources such as the Leonardo supercomputer\footnote{\url{https://leonardo-supercomputer.cineca.eu/}}, which remains out of reach for many individual researchers.

The community response has been extraordinary. More than 80 contributors from 31 different institutions both within and beyond NLP, including non-academic stakeholders such as journalists, publishers, and public sector professionals participated in defining tasks, annotating data, and supporting the automation of evaluation pipelines and the use of computational clusters. This distributed, self-organized effort enabled the evaluation of four LLMs across 22 distinct challenges and 95 sub-tasks, all orchestrated in a collaborative and centralized fashion. It also established an infrastructure which allows for the continuous addition of more challenges to the dynamic benchmark and the evaluation of other models (see Section~\ref{sec:outlook}).

Generating a large volume of evaluation results presents its own challenges. Within CALAMITA we have 
organized the challenges in a taxonomy to better see  generalizations of model behaviors, while we have also encouraged the development of task-representative metrics so that model performance can be studied at a more fine-grained level without reducing it to a single averaged score.
Indeed, CALAMITA does not aim to simply provide an abstract ranking of LLMs, nor to declare a definitive ``winner'' among models. Instead, it reflects the unique experience of a research association that has succeeded in federating multiple and diverse groups towards the creation of a unique resource. This resource is not limited to the current benchmark, which is composed of a large diversity of tasks, and has enabled a first round of experiments enriched with error analysis --- this resource is the methodology itself: a collaborative framework for dataset creation and evaluation that combines linguistic insight, computational efficiency, and community engagement.

CALAMITA is not conceived as a one-off experiment, but as an ongoing, sustainable initiative, grounded on a rolling mechanism which will enable the continuous addition of tasks and evaluation of models. Its dynamics will continue to be community-driven, and guided by the constantly evolving needs of the Italian, as well as the international research landscape.
Since this framework has proven successful in promoting a robust and meaningful evaluation of LLMs in languages beyond the current mainstream, we hope that our experience can serve as a blueprint for other research communities towards a more inclusive and comprehensive landscape for natural language processing research. As multilingual LLMs continue to advance, it is crucial to ensure that evaluation practices keep pace, especially for languages with limited resources and less international visibility.

For the reasons above, this paper contains not only a description of the challenges composing the CALAMITA benchmark at the time of writing, and results -- both quantitative and qualitative -- on four open-weight LLMs, but also a detailed explanation of the collaborative methodology we adopted both for data creation as well as for model evaluation.
We discuss the linguistic, practical, and technical challenges encountered, and the solutions developed collaboratively by the community. Finally, we outline ongoing and future directions, inviting continued participation and replication of our approach in other contexts.

\section{Related Work}
\label{sec:related}

Benchmarking has long served as a key contributor to empirical progress in NLP. In the past, many benchmarks were designed to test the general linguistic competence of language models through tasks such as summarization or paraphrase detection, which served as proxies for broader language understanding. Others focused on more narrowly defined tasks with a view to downstream applications, such as review classification, or hate speech detection, with datasets often designed both for model development and evaluation. This was reflected in long-standing evaluation campaigns such as SemEval\footnote{\url{https://semeval.github.io/}}, or--in the Italian context--EVALITA\footnote{\url{https://www.evalita.it/}}.

However, the rapid improvement of large language models (LLMs) is currently shifting the benchmarking landscape. More than focusing on downstream applications with models trained to solve specific tasks, there is a growing need for benchmarks that help to capture the diversity of LLMs’ emerging abilities, especially in light of the fact that they have not been explicitly trained for those. Example abilities are reasoning, factual consistency, behavioural characteristics, and alignment with human values. This section does not attempt to be a comprehensive survey of the literature on language model evaluation. Instead, in the first part we illustrate the diversity of LLM benchmarks and the types of tested abilities they include; in the second we discuss initiatives that are similar to ours in scope and methodological approach.

\paragraph{Diversity of tasks and tested abilities in recent  benchmarking}
Many benchmarks continue to focus on traditional language processing tasks, including question answering \cite{qa_kwiatkowski2019natural,joshi2017triviaqa, choi2018quac, lin2021truthfulqa}, reading comprehension \cite{dua2019drop, lai2017race, bandarkar2023belebele}, word prediction \cite{paperno2016lambada}, sentence completion \cite{zellers2019hellaswag}, summarization \cite{ladhak2020wikilingua}, paraphrase detection \cite{yang2019paws}, acceptability judgment \cite{warstadt2020blimp, suijkerbuijk2025blimp, zhang2023mela}, and dialogue modeling \cite{bai2024mt}. Such foundational tasks are often grouped into large-scale benchmarks, such as GLUE \cite{wang2018glue}, SuperGLUE \cite{wang2019superglue} or XTREME \cite{hu2020xtreme}, which have for years served as reference benchmarks for first generation pretrained models such as BERT, RoBERTa, and GPT-2. Differently from more recent ability-testing benchmarks, a feature of GLUE and XTREME is that they provide task-specific training datasets on which pretrained models should be fine-tuned, in addition to evaluation sets for testing.

Research into the emerging abilities of LLMs has attracted diverse lines of inquiry. We outline several below, including domain-specific knowledge, multiple types of reasoning, commonsense knowledge, cognitive aspects, and pragmatics, and fairness. These are also categories that we draw upon to characterize and classify our CALAMITA challenges (see Table~\ref{tab:categories} and Section~\ref{sec:results-detailed}).

Domain-specific benchmarks have been developed to test how well language models can operate within particular fields, either by assessing their factual knowledge or their ability to perform tasks that are part of real-world applications. Benchmarks have been designed to assess knowledge in science \cite{welbl2017crowdsourcing, guo2023can} or finance \cite{koncel2023bizbench}. AGIEval \cite{zhong2023agieval} compiles questions from standardized exams such as the SAT or legal entry exams. In professional contexts, ACI-Bench \cite{yim2023aci} evaluates models on clinical note generation for medical visits, and QMSum \cite{zhong2021qmsum} assesses automatic meeting summarization in multiple domains.

Recently, a central open question is the extent to which LLMs can perform formal reasoning -- that is, the ability to reason abstractly according to explicit rules. Popular tasks targeting formal reasoning include the solution of mathematical word problems (\citet{wang2017deep,cobbe2021training, li2024gsm}, see \citet{chen2022unigeo} for geometry problems) and code generation \cite{austin2021program, yan2023codescope, zhang2024benchmarking, lu2021codexglue, khan2023xcodeeval, kasner2024beyond}, but benchmarks have also been introduced testing performance on algorithmic problems \cite{fan2023nphardeval, parmar2024logicbench}, graph extraction \cite{du2024paged}, quantitative reasoning \cite{ravichander2019equate} and fallacy detection \cite{helwe2023mafalda}.  Commonsense reasoning benchmarks have emerged as a way to evaluate forms of reasoning that are not governed by formal logic but are typically intuitive for humans \cite{clark2018think, ponti2020xcopa, sakaguchi2021winogrande, frohberg2021crass}. These tasks aim to assess models’ ability to make plausible inferences about everyday situations. Interestingly, what is regarded as commonsense may be culture-specific, and this understanding may be evaluated independently \cite{sun2024benchmarking}. Related areas such as physical commonsense \cite{bisk2020piqa}, temporal reasoning, \cite{zhang2024analyzing}, and natural language inference \cite{multinli_williams2017broad, sadat2022scinli} are often included under this umbrella. To assess various forms of reasoning at once, using language models to solve puzzles is a creative branch of research; for example, \citet{todd2024missed} test lateral thinking by having LLMs solve The New York Times' \textit{Connections} puzzles.
A growing area of evaluation builds on insights from cognitive science and psychology to probe the extent to which LLMs exhibit human-like social and pragmatic reasoning. Theory-of-Mind (ToM) benchmarks test whether models can infer the mental states of others -- a fundamental component of social intelligence \cite{kosinski2023theory, hu2025re, shapirachallenging, gordon2016commonsense, le2019revisiting, shapira2024clever, xu2024opentom, chen2024tombench}. Social IQa \cite{sap2019socialiqa} targets commonsense reasoning about social interactions, and EmoBench \cite{sabour2024emobench} evaluates emotional intelligence. 
Related work in pragmatics explores figurative language understanding \cite{liu2022testing, liu2022figmemes, stowe2022impli,ma-etal-2025-pragmatics}, implicature and presupposition \cite{zheng2021grice, hu2022fine, jeretic2020natural}, intention recognition \cite{sakurai2024evaluating}, and neologism understanding \cite{zheng2024neo}. MM-SAP \cite{wang2024mm} assesses self-awareness, and \citet{hessel2022androids} test understanding of humor. 

With LLMs becoming increasingly prevalent in daily life, ethical behavior becomes an important dimension of LLM evaluation, such as testing a model’s ability to avoid social biases and toxic or harmful outputs \cite{hartvigsen2022toxigen, magooda2023framework, zhang2023safetybench, liu2023alignbench}. The assessment of a model’s ability to make appropriate moral judgements is an open problem, in part due to morality’s subjective nature - SCRUPLES \cite{lourie2021scruples} compiles community moral judgements with reasoning from Reddit’s \textit{AITA} forum. 

Recent benchmarking efforts reflect the fundamental shifts in NLP research culture and priorities brought about by the widespread adoption of LLMs. As the ability to distinguish between human- and machine-generated text becomes more pressing, benchmarks emerge to assess models’ ability to detect synthetic texts \cite{dugan2024raid} and evaluate LLM watermarks \cite{tu2023waterbench}. The increase in possible context window sizes has prompted the development of benchmarks designed to test LLM performance on inputs spanning over a hundred thousand tokens \cite{zhang2024infty}. At the same time, growing concerns about factual reliability have motivated benchmarks assessing hallucination tendencies \cite{liang2023uhgeval} or robustness \cite{siska2024examining}. The increasing popularity of chain-of-thought prompting to elicit intermediate reasoning steps has led to the development of datasets that focus on verifying reasoning chains \cite{jacovi2024chain}. Finally, the open-ended nature of LLM generation has increased the importance of format adherence \cite{jiang2023followbench,xia2024fofo}.

\paragraph{Large-scale and collaborative LLM benchmarking}

MMLU (Measuring Massive Multitask Language Understanding, \citep{hendrycks2020measuring}) has been the reference benchmark for years to test LLMs' abilities in commonsense and math reasoning as well as domain knowledge in academic fields like STEM, humanities, social sciences, and law, with the tasks framed as multiple-choice questions adapted from real exams and textbooks. One outstanding problem with LLM benchmarks such as MMLU is that with the rapid progress in LLM development they are quickly \textit{saturating}, i.e., they have become increasingly easy for advanced models to surpass, making it difficult to distinguish subtle differences in model abilities. The development of particularly difficult benchmarks, such as MMLU-Pro \citet{wang2024mmlu} and Humanity’s Last Exam (HLE, \citet{phan2025humanitysexam}) was aimed at addressing this issue. 

In the context of our initiative, the genesis of HLE, which features 2,500 expert-level questions across multiple disciplines, is particularly interesting. To build this resource, the organizers (Scale AI\footnote{\url{https://scale.com/}}, a company which produces curated data and infrastructure to AI companies which develop LLMs) sent out calls through academic communications channels such as faculty mailing lists, graduate programs, and conference communities for subject-matter experts from academia and industry, in multiple fields, such as neuroscience, law, mathematics, medieval history, etc. Eventually, they received contributions from a total of nearly 1,000 contributors across 50 countries and more than 500 institutions. All contributions were voluntary and were structured as fully specified exam-style questions submitted through a secure online platform. 

CALAMITA shares with HLE the community-driven spirit which drives development. 
However, not only does HLE still focus on English, but mostly its creation was motivated by saturating benchmarks, with the aim of creating truly \textit{difficult} questions, where variation is given by the presence of questions in multiple expert fields. Our aim is to create a diverse benchmark to test multiple abilities of LLMs. Lastly, while HLE is a community effort which is company-driven, CALAMITA is a community effort coordinated by the association representing the Italian NLP research community. 

The initiative that CALAMITA is closest to is BIG-bench, a community-sourced benchmark, focused on English and launched and coordinated by Google Research, featuring 204 tasks contributed from 442 researchers across 132 institutions \citep{srivastava2023beyond}. We share with this project attention to diversity and a collaborative, bottom-up strategy to benchmark creation, departing from the top-down approach typical of the more classic benchmarks mentioned above, including MMLU. Yet again, rather than born in the context of a company, CALAMITA is a non-English effort led by the national NLP Association in Italy, with a dynamic structure which enables the continuous addition and evaluation of new tasks (Section~\ref{sec:outlook}).

Lastly, we would like to mention some very recent efforts by the Iberian NLP community, who in 2024-2025 started two initiatives similar to CALAMITA, namely IberoBench \citep{baucells-etal-2025-iberobench} and IberBench \citep{gonzalez2025iberbench}. 
Researchers mostly took existing benchmarks from previous efforts such as IberEval and IberLEF \citep[e.g.]{ibereval-2018,iberspeakers-iberlef-2023}, and urged the NLP community to contribute their help to integrate them, correct them, etc. Also due to this strategy, the benchmarks  mostly revolve around traditional NLP tasks, such as Sentiment Analysis, Natural Language Inference, Named Entity Recognition. While we also plan to integrate in CALAMITA past datasets (and thus tasks) from previous EVALITA campaigns (coordinating with ongoing efforts \cite{magninietal25})\footnote{This is already partly happening with the integration of \textsc{ItaEval}, which contains some of the past EVALITA tasks, see Section~\ref{sec:itaeval}. For a more general discussion of the interaction of CALAMITA and EVALITA please see Section~\ref{sec:outlook}.}, with CALAMITA we aimed at building a broad and diverse community first, calling upon a large variety of often not yet tested challenges. While the very recent development of these two Iberian benchmarks provides further evidence that community-driven efforts are both a current need and a powerful strategy in NLP benchmarking, they are separate, and born of independent initiatives. \citet{baucells-etal-2025-iberobench} explicitly state that ``IberBench and IberoBench were developed independently and in parallel, without the involved parties being aware
of each other". CALAMITA, instead, is a fully centralized initiative launched and supported by the Italian Association for Computational Linguistics, to minimize dispersion of efforts, maximize harmonization, and truly create a sense of community. The core of this paper lies indeed with the collaborative methodology that we adopted, rather than an assessment and ranking of a wide range of LLMs. The latter effort is something that the CALAMITA benchmark enables as future research.

\section{Collaborative Methodology}
\label{sec:method}

\subsection{Procedure}

The CALAMITA organization collective included four chairs (senior academics with an NLP background, all members of the steering board of AILC, who conceived, launched, and coordinated the initiative) and an evaluation team, who took care of data collection and processing, and of model testing. The latter group was formed by three master students, three PhD students, and one postdoc who also served as coordinator of the whole evaluation team. This collective of eleven people represented seven distinct academic affiliations. 
Ten months elapsed from conception to the first workshop, which featured 22 challenge presentations, published proceedings, and a public leaderboard including most results for two LLMs. We outline below a more fine-grained description of the intermediate steps.

\paragraph{Pre-proposals} By means of a call for pre-proposals distributed through standard communication channels, as well as personal networks, potentially interested parties were informed of the initiative and asked to provide a short proposal with their idea for a challenge that would target specific abilities of LLMs. Proposers were asked to provide a motivation, and as many details as possible on the proposed dataset, with the requirement that it would be natively created in Italian. Acceptance was based on meeting this basic requirement, on the feasibility of dataset creation, and on the conceptual soundness of the proposed challenge. No assessment was based on the practical utility of the task, nor on its novelty. All accepted pre-proposals (22 out of 26) were then invited for a full submission to the CALAMITA benchmark, with further instructions \citep{attanasio-etal-2024-calamita}.

\begin{wrapfigure}{r}{8.6cm}
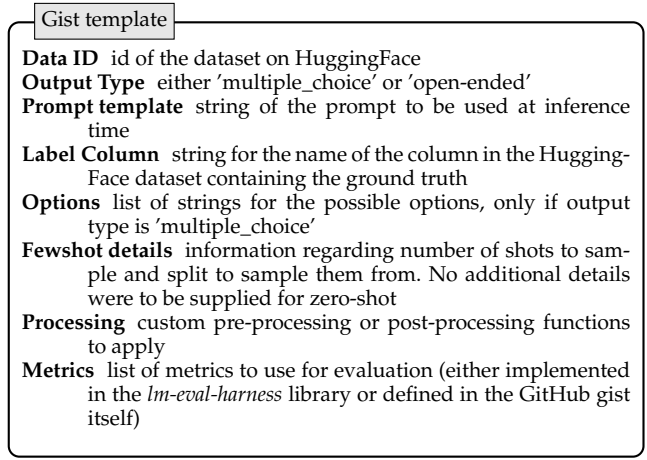

\centering
\footnotesize
\mnbox{
\begin{description}
\itemsep 0pt
    \item[Data ID] id of the dataset on HuggingFace
    \item[Output Type] either 'multiple\_choice' or 'open-ended'
    \item[Prompt template] string of the prompt to be used at inference time
    \item[Label Column] string for the name of the column in the HuggingFace dataset containing the ground truth
    \item[Options] list of strings for the possible options, only if output type is 'multiple\_choice'
    \item[Fewshot details] information regarding number of shots to sample and split to sample them from. No additional details were to be supplied for zero-shot
    \item[Processing] custom pre-processing or post-processing functions to apply
    \item[Metrics] list of metrics to use for evaluation (either implemented in the \textit{lm-eval-harness} library or defined in the GitHub gist itself) 
\end{description}
}{Gist template}
\caption{\label{fig:gist}Gist template.}
\end{wrapfigure}

\paragraph{Submission}
All final participants had to submit three artifacts: 1) a technical report of the data and tasks\footnote{These were eventually collected in the Proceedings of the CLiC-it Conference that the first CALAMITA event was co-located with \citep{DBLP:conf/clic-it/2024}.}; 2) the challenge dataset; 3) one GitHub gist for each task describing information required for its technical implementation.
The template of the technical report was provided by the CALAMITA team and included sections for the challenge and data description (including source and any copyright and privacy requirements). The submission of the technical reports was handled through the SoftConf platform, where at least two members of the CALAMITA team (from both the chairs and the evaluation team) reviewed the technical report providing suggestions also aimed at increasing coherence across reports. 
To standardize the access to resources and formatting and promote openness, we recommended participants to share their datasets through HuggingFace\footnote{\url{https://huggingface.co/}}.
However, in the case of issues related to making the data public (e.g., privacy requirements), it was also possible for participants to submit the data through SoftConf in a dedicated zip file.
For the specifics of the GitHub gist, see Section~\ref{sec:tech-aspects} and Figure~\ref{fig:gist}.
Each challenge could include more than one (sub)task. Each task could have a different metric, if necessary. The participants were also asked, in case of tasks having multiple metrics, to choose the main metric to be used for a given subtask to be used in the final CALAMITA's leaderboard; this is also the metric we use for the aggregated results in Section~\ref{sec:results-overview-agg} and in Table~\ref{tab:full-results} in the Appendix.

From the beginning of the submission phase, a Slack server was created to manage communication internally and facilitate the distribution of tasks across the evaluation team, but also with all participants. In fact, a dedicated channel was used to support all the challenge proposers who needed help in processing data, identifying the best metrics for their tasks, and creating the gist itself towards a successful submission.
Arguably, a submission procedure that requires a descriptive Gist and short pseudo-code was a winning aspect of the collaborative paradigm. Participants had not to implement challenges themselves but interact and monitor our evaluation team's efforts in implementing them in a streamlined, centralized, and standard way.
This was particularly helpful for those groups who joined the CALAMITA initiative as non-experts in NLP or data processing in the context of model evaluation.

\paragraph{Evaluation}
Each challenge was assigned to one member of the evaluation team, with the task of staying in touch with the proposers, help them with finalizing the data and the metrics when necessary, and then run the evaluation on the Leonardo HPC. The load was spread such that each member implemented at most eight subtasks.

A preliminary leaderboard for \llama{} and \anita{} was released to the participants, including all CALAMITA tasks.
Participants discussed the process and the obtained results with the member of the evaluation team responsible for their task to flag potential errors and identify necessary fixes. After further improving the codebase for all tasks in need of fixes, a new evaluation run was performed.

\paragraph{Workshop and community sharing}
On the occasion of the annual national conference of the Italian Association for Computational Linguistics, CLiC-it~2024\footnote{\url{https://clic2024.ilc.cnr.it}}, CALAMITA received a dedicated slot for presenting the whole framework, the 22 challenges, and the leaderboard. All tasks were presented directly by a representative of the proposing team. A discussion panel followed to devise a route to a sustainable future of CALAMITA.

\paragraph{Follow up}
Two additional models were added to the pool of tested LLMs, namely Minerva-7B-Instruct \cite{orlando-etal-2024-minerva}, which was released at the time of the workshop, and LLaMA-3.1-Instruct-70B \cite{llama3} to scale up model complexity. Task organizers were then provided with the outputs from all four models, and were asked to perform error analysis which should be included in a brief one-page report. The 22 resulting reports form the core of the Results section of this paper.

Finally, to ensure the growth of our benchmark, we have just recently initiated a rolling task submission procedure and a call for participation in the evaluation team. Further details are provided in Section~\ref{sec:outlook}.

\subsection{Technical Aspects}
\label{sec:tech-aspects}

While datasets were created by the proposing teams, the submitted data as well as the scoring procedures had to be further processed to enable fully automatic evaluation. 

\paragraph{Dataset processing}
As mentioned, participants were required either to upload their datasets on Huggingface or to send the datasets' zip file through Softconf, and submit a separate GitHub gist for each subtask. 
We provide the template used for each  GitHub gist in Figure~\ref{fig:gist}.

The evaluation team implemented each task in the \texttt{lm-evaluation-harness}\footnote{\url{https://github.com/EleutherAI/lm-evaluation-harness}} (\texttt{lm-eval}, for brevity) library \cite{eval-harness}, following the requirements outlined by participants in the GitHub gists. A fork of the library was created to implement the tasks. We chose \texttt{lm-eval} because of its focus on reproducibility and the ease with which the library allows for the inclusion of additional tasks. Each sub-task was implemented in a dedicated YAML  file, following the formatting expected by the library. This YAML file is used to setup common details in task implementation, like the prompt template or the label column, and was created following the information provided in the GitHub gist by the task proposers. 

Following standard \texttt{lm-eval} practices, each implemented task fell into one of two categories: Multiple-Choice Question Answering (MCQA) and Open-Ended Generation (OE). For MCQA tasks, we measured accuracy computed by comparing length‑normalized log-probabilities---each choice's log-probability is divided by the choice length (unit depends on the task: bytes, tokens, or characters) and the choice with the highest normalized score is picked and used for the aggregate metric defined by the proposers, e.g., F1 Macro. OE tasks required more careful considerations. In some cases, we iterated with the proposers to formalize output parsing, especially for structured outputs such as JSON, or task-specific decoding choices such as the end-of-sentence (EOS) token. Moreover, unless the challenge proposers requested otherwise, we opted not to use chat templates and set the decoding temperature to 0.

\paragraph{Model evaluation}
With all tasks implemented in the \texttt{lm-eval-harness}, the evaluation runs were performed on LEONARDO \cite{leonardo}, a supercomputer that we could access thanks to the acquisition of an ICRA-C project grant (10K~GPU/hours).\footnote{Each node in LEONARDO has 4 NVIDIA A100 64 GB VRAM GPUs.} We tested four different models. A first round of evaluation was run for LLaMAntino-3-ANITA-8B-Inst-DPO-ITA \cite{polignano2024advanced} and LLaMA-3.1-8B-Instruct \cite{llama3}, both sharing the same base model (\llama); the former is instruction-tuned with instructions (automatically) translated into Italian, while the latter is instruction-tuned with English data. In a second round we also tested Minerva-7B-Instruct \cite{orlando-etal-2024-minerva} and LLaMA-3.1-Instruct-70B \cite{llama3}. This allowed us to consider a model trained from scratch with a focus on the Italian language (Minerva-7B-Instruct's pretraining data comprises both Italian and English texts in equal proportions), and a multilingual model with a greater number of parameters to also test the influence of size. Outputs were checked with each task's organizers to ensure they were implemented correctly.

The codebase is publicly available\footnote{\url{https://github.com/CALAMITA-AILC/calamita-eval}}, including all CALAMITA (sub)tasks. We also plan to release all of the models' outputs (which are used for the error analyses presented in Section~\ref{sec:results}) on a public repository alongside this work.

\subsection{Hurdles and bottlenecks}

As the organizers of BIG-Bench and HLE also note, ensuring consistency and quality across so many datasets and contributors is a very difficult challenge. Centralizing data collection, actual task implementation, and evaluation runs in the hands of a small team of evaluators, and by means of regular interaction with all the challenge organizers, mitigated this problem. Still, some aspects proved particularly challenging to address, especially when custom implementations were necessary. 
In most cases, \texttt{lm-eval} supported the task implementation required by the authors, but for a few edge cases, some missing functionalities had to be added to \texttt{lm-eval}. For example, the library did not support task dependency and chained execution (e.g., Task Y requires the output of Task X), a case occurring in one of the CALAMITA challenges.

The framing of the tasks also presented complexities. While, for example, multiple choice questions were straightforward to implement thanks to the \texttt{lm-eval-harness}' built-in log-likelihood based evaluation, standardizing evaluation of more purely generative tasks was not trivial. On the one hand, autoregressive generation is computationally expensive, so that a significantly larger amount of time had to be allocated for evaluation and also debugging. On the other hand, the variability in models' generations, particularly when the difficulty of the tasks or the longer context caused models to struggle, complicated the design of post-processing strategies aimed at extracting the relevant content in the output. This was particularly problematic for tasks that required the models to answer with specific data structures, such as JSON objects. In these cases, it was often necessary to interact with the challenge organizers to establish the limits of acceptable or unacceptable answers. 
More generally, parsing the models' outputs can lead to significant variations in the final scores.

\section{CALAMITA Challenges: Descriptions and Results}
\label{sec:results}

The first iteration of CALAMITA resulted in 22 challenges articulated in a total of 95 subtasks, all of which appear on the current leaderboard.\footnote{Leaderboard accessible at \url{https://github.com/CALAMITA-AILC/calamita-eval}.}

The collaborative spirit at the heart of this initiative has fostered a considerable diversity in the proposed challenges, since researchers from multiple fields are interested in testing different model abilities. To make a better sense of the diverse challenges, we opted for categorizing them into a systematic taxonomy, since it enables a more informative identification of model weaknesses across distinct cognitive and linguistic domains, thereby helping with results and error analysis. Based on observations from previous benchmarking work (see Section~\ref{sec:related}), we defined eight categories spanning general abilities--commonsense knowledge, factual knowledge, linguistic knowledge, formal reasoning, and fairness and bias--alongside specialized NLP capabilities including code generation, machine translation, and summarization. 
Table~\ref{tab:categories} describes each top-level category and reports how many challenges and respective subtasks it includes. Challenges and tasks may belong to more than one category; an overview of challenges with associated ability categories is shown in Table~\ref{tab:challenge-abilities} in the Appendix. In the Appendix, we also report a detailed table with each category and its relative challenges and subtasks (Table~\ref{tab:category-parent-task}).

\begin{table*}[!t]
\footnotesize
\caption{Categories of abilities tested by CALAMITA challenges. Challenges (\textbf{\#Ch}) test general abilities such as knowledge about true facts, commonsense, and logical reasoning or specific NLP-oriented abilities such as code generation or machine translation, and are realized through a number of specific subtasks (\textbf{\#T}).\label{tab:categories}}
\begin{tabular}{>{\raggedright\arraybackslash}p{2.4cm} p{8cm} rr}
\toprule
\textbf{Ability tested} & \textbf{Description} & \textbf{\#Ch} & \textbf{\#T}\\
\midrule Commonsense reasoning & General knowledge about the world that is typically taken for granted in everyday life, e.g., everyday cause-and-effect relationships, situational judgments, physical properties, and basic social interactions. & 10 & 42\\ 
\midrule
Factual knowledge & Knowledge of concrete, verifiable facts about the world, e.g., definitions, historical events, or scientific concepts. & 8 & 29\\
\midrule
Linguistic knowledge & Linguistically motivated tasks that test specific language skills, e.g., word sense disambiguation, coreference resolution, or acceptability judgment. & 14 & 60\\
\midrule
Formal reasoning & Ability to understand and use formally logical principles to solve problems, e.g., mathematical problems. & 7 & 22 \\
\midrule
Fairness and bias & Evaluates a model's capacity to handle sensitive tasks, including exclusive and stereotyped language understanding and detecting offensive or biased language towards social groups. & 3 & 16\\
\midrule
Code generation & Ability to generate fully functioning code for a specific programming language. & 1 & 1\\
\midrule
Machine translation & Ability to translate a sentence from a source language into another language, with one of the two being Italian. & 2 & 15\\
\midrule
Summarization & Ability to create relevant summaries of a given excerpt, e.g., news headline generation or news reduction. & 2 & 4\\
\bottomrule
\end{tabular}
\end{table*}

Under \textit{commonsense knowledge}, we have included tasks such as ECWCA and EurekaRebus, both involving language puzzles calling upon reasoning over general knowledge, INVALSI and MultIT, multiple-choice challenges on general knowledge, and MACID, a challenge revolving around the identification of closely related action concepts. Under \textit{factual knowledge}, in addition to the already mentioned multiple-choice benchmarks, we have included challenges like BEEP, designed to evaluate LLMs on a simulated Italian driver’s license
exam, and VeryfIT, where LLMs are evaluated on their in-memory factual knowledge on data written by professional fact-checkers. \textit{Linguistic abilities} are of the kind tested in the TRACE-it challenge, where models are tested for their ability to comprehend a specific type of complex syntactic construction in Italian, in BLM-it which tests the representation and knowledge of abstract linguistic rules in LLMs, or in ABRICOT, where models are tested on their ability to understand and
assess the abstractness and inclusiveness of language. The linguistic puzzle challenges are also categorized as \textit{formal reasoning} challenges, together with, for example, GEESE, which assesses the impact of generated explanations on the predictive performance of LLMs in the Textual Entailment task, and GITA, where models are tested for their reasoning capabilities regarding the physical world. Two challenges probe LLMs for their \textit{fairness}, namely GFG, designed to assess and
monitor the recognition and generation of gender-fair language in both mono- and cross-
lingual scenarios, and \textsc{PejorativITy}, which investigates misogyny expressed through neutral
words that can acquire negative connotation when used as pejorative epithets. GFG is also one of the two tasks testing \textit{machine translation} abilities, together with MAGNET, where models are tested in their translation skills, focusing on Italian and English. Lastly, one challenge (TERMITE) focuses on testing \textit{coding} abilities in the context of SQL queries, and one (GATTINA) on the models' summarisation abilities by making them generate headlines for news articles. From the original \textsc{ItaEval} benchmark \citep{DBLP:conf/clic-it/AttanasioDQSS24,DBLP:conf/clic-it/AttanasioQSS24}, we have retained only those tasks which had been natively developed in and for Italian, and decided not to include in the official CALAMITA benchmark the datasets derived from translating original English datasets. Nevertheless, through the CALAMITA evaluation framework, we could easily evaluate all of the \textsc{ItaEval} tasks, including the translated ones. While these results are not included in any of the aggregated analyses that we present in Section~\ref{sec:results-overview-agg}, nor in the overview tables in the Appendix, they are discussed in the paragraph specifically devoted to \textsc{ItaEval} results (Section~\ref{sec:itaeval}) and reported for completeness in a separate table in the Appendix (Table~\ref{tab:itaeval-english-results}). Each retained \textsc{ItaEval} task was also assigned one or more of our taxonomic categories.\footnote{The \textsc{ItaEval} tasks which are not included in CALAMITA due to their being originally developed in and for English are: arc\_challenge\_ita, belebele\_ita, hellaswag\_ita, squad\_it, truthfulqa\_mc2\_ita, xcopa\_it; see \citep{DBLP:conf/clic-it/AttanasioDQSS24,DBLP:conf/clic-it/AttanasioQSS24} for more details on these tasks, and Table~\ref{tab:itaeval-english-results} in the Appendix for actual results on the four models tested within CALAMITA.}

Each challenge, with challenge-specific results and insights, is discussed in Section~\ref{sec:results-detailed}. In contrast, Section~\ref{sec:results-overview} presents a broader discussion with a higher-level analysis across challenges and categories based on aggregated results. Full detailed results per challenge and subtask are included in the Appendix (Table~\ref{tab:full-results}). Please note that details for each challenge regarding background, motivation, data collection and dataset composition, as well as metrics is available in the challenge description papers which have been collected and published as part of the introduction to the CALAMITA initiative. They are collected in \citep{DBLP:conf/clic-it/2024} and referred to specifically in each of the challenge-specific paragraphs below.

\subsection{Challenge-Specific Insights and Lessons Learned}
\label{sec:results-detailed}

This section\footnote{Each subsection was drafted by the corresponding challenge organizers based on the shared model outputs. As a result, prose style and reporting conventions naturally vary across sections. We have harmonized table formats and model naming, while postponing full unification of figure styles to the final version.
For several challenges Italian examples are shown, but not all of them are translated. In the final version, all such examples will be accompanied by an English translation to ensure clarity for all readers.} reports task-specific insights contributed by the organizers of each challenge after the evaluation campaign. Following the collective run on four open-weight models (\llama, \llamalarge, \anita, \minerva), we returned both scores and raw generations to all teams and asked them to perform a short, focused error analysis. Their one-page reports constitute the backbone of the aggregate analysis which follows, and make this the most collaborative portion of the paper. For each task we provide: (i) a concise task synopsis; (ii) a brief summary of results; and (iii) a critical analysis of typical errors and edge cases, so that aggregate numbers are complemented with linguistic evidence about strengths and weaknesses of each model. This procedure aims to turn a static and rather dry leaderboard into a more actionable understanding of model behavior across heterogeneous tasks, metrics, and prompts.

\subsubsection{ABRICOT - ABstRactness and Inclusiveness in COntexT} \posscite{DBLP:conf/clic-it/0002CREB24} challenge evaluates Italian language models on their ability to understand and assess the abstractness and inclusiveness of language, two nuanced features that humans naturally convey in everyday communication. Unlike binary categorizations such as abstract/concrete or inclusive/exclusive, these features exist on a continuous spectrum with varying degrees of intensity. The task is based on a manual collection of sentences that present the same noun phrase (NP) in different contexts, allowing its interpretation to vary between the extremes of abstractness and inclusiveness. This challenge aims to verify how LLMs perceive subtle linguistic variations and their implications in natural language.
One of the defining features of human language is its ability to convey both specific and general information using the same lexical item. 
Consider the following examples: (a)\textit{``The} \underline{turtle} \textit{escaped from the zoo.''} and (b) \textit{``The} \underline{turtle} \textit{is a reptile.''} In (a), \textit{turtle} designates a specific individual, whereas in (b), it generalizes to an entire category of reptiles. This feature of natural languages, called {genericity} \cite{genericityintro} is fundamental to linguistic economy, allowing speakers to efficiently express a wide range of meanings without requiring a separate lexical item for every level of genericity. 
This task was designed to evaluate the ability of Italian language models to grasp these two semantic properties of word meaning in context: {abstractness}, the degree to which a noun phrase (NP) refers to a concept rather than a tangible entity, and {inclusiveness}, the extent to which an NP denotes a broad category encompassing multiple subtypes.
The task consists of a dataset of 127 samples, each featuring a specific NP and its corresponding abstractness and inclusiveness scores, as annotated by five human raters on a continuous scale from 0 to 1. Notably, the dataset includes 20 unique NPs that recur across different contexts, allowing for a nuanced assessment of how context influences semantic interpretation.
ABRICOT is divided into two subtasks: {ABRICOT-ABS}, where language models assign a score from 1 to 5 to assess the abstractness of an NP in context. {ABRICOT-INC}, where models evaluate the NP's inclusiveness using the same scoring scale. Model performance is measured using Pearson correlation with human annotations, quantifying how well computational models approximate human semantic intuition.

\begin{table}[!ht]
\footnotesize
\centering
\caption{Results for the ABRICOT task, which evaluates model sensitivity to abstractness (ABS) and inclusiveness (INC) in context. Scores are reported as Pearson correlations with human annotations, comparing four Italian LLMs across the two subtasks.}
\label{tab:abricot}
\begin{tabular}{lcccc}
\toprule
 & \textbf{\llama} & \textbf{\llamalarge} & \textbf{\anita} & \textbf{\minerva} \\
\midrule
\textbf{ABRICOT--ABS} & 0.29 & 0.44 & 0.50 & -0.03 \\
\textbf{ABRICOT--INC} & 0.18 & 0.25 & 0.14 & -0.20 \\
\bottomrule
\end{tabular}
\end{table}

\paragraph{Performance across models}
Table~\ref{tab:abricot} shows the Pearson correlation scores for each of the four tested models \llama{}, \llamalarge{}, \minerva{} and \anita{}) on the two ABRICOT subtasks. \llamalarge{} outperforms other models on both subtasks, while \minerva{} Instruct performs the worst, showing negative correlation for the ABRICOT-INC subtask.

\paragraph{Error Analysis}
All models test are far from human performance. We report an error from the best model, \llamalarge{}, it assigns the same score, 3, to the word \emph{persona} (person) when seen in two contexts: (1) ``\textit{Le persone affette da eczema sono spesso sensibili ai prodotti aggressivi}''\footnote{``\textit{Persons affected by eczema are sensitive to aggressive products}''} and (2) ``\textit{Questa persona è davvero noiosa.}''\footnote{``\textit{This person is really boring.}''} However, the first occurrence of \emph{persona} encompasses a large number of individuals, while the second refers to a specific individual.

\subsubsection{AMELIA - Argument Mining Evaluation on Legal documents in ItAlian}

 \posscite{DBLP:conf/clic-it/GrundlerGSFGPLS24} challenge consists of three classification tasks in the context of argument mining in the legal domain.
The goal of the challenge is to assess the ability of LLMs to understand logical relations in legal reasoning, a particularly challenging field, due to its domain-specific technical language, that uses complex syntactic structures.
The tasks are based on a dataset of 225 Italian judicial decisions on Value Added Tax, annotated to identify and categorize argumentative text.
The objective of the first task is to classify each argumentative component as a premise or conclusion.
The second task aims at predicting the type of a premise as a multi-label classification: legal, factual, or both. A factual premise describes factual situations and events, pertaining to the substance or the procedure of the case, while a legal premise specifies the legal content, such as legal rules, precedents, interpretation of applicable laws and principles.
The third task aims at identifying the argumentation scheme, or reasoning patterns, of legal premises in a multi-label classification setting.
Given that in all three tasks the classes are unbalanced, the evaluation is based on the macro F1 score.

\begin{table*}[!ht]
\centering
\footnotesize
\setlength{\tabcolsep}{4pt} 
\caption{
Results for the \textsc{AMELIA} task.  
For each subtask (Argument Component, Premise Type, and Argument Scheme) we report the F1-score for each class and the macro-average across classes, under zero-shot and few-shot settings.
}
\label{tab:amelia_results}
\resizebox{\linewidth}{!}{
\begin{tabular}{l l ccc ccc ccccc c}
\toprule
\multirow{2}{*}{\textbf{Model}} & 
\multirow{2}{*}{\textbf{Shot}} &
\multicolumn{3}{c}{\textbf{Arg Component}} &
\multicolumn{3}{c}{\textbf{Premise Type}} &
\multicolumn{6}{c}{\textbf{Arg Scheme}} \\
\cmidrule(lr){3-5}\cmidrule(lr){6-8}\cmidrule(lr){9-14}
 &  & prem & conc & Avg & F & L & Avg & Class & Itpr & Prec & Princ & Rule & Avg \\
\midrule
\multirow{2}{*}{\llamalarge} 
  & zero & 0.93 & 0.49 & 0.71 & 0.86 & 0.83 & 0.85 & 0.18 & 0.06 & \textbf{0.76} & 0.35 & \textbf{0.76} & 0.42 \\
  & few  & \textbf{0.96} & \textbf{0.76} & \textbf{0.86} & \textbf{0.87} & \textbf{0.85} & \textbf{0.86} & \textbf{0.46} & \textbf{0.44} & 0.75 & \textbf{0.42} & 0.75 & \textbf{0.57} \\
\midrule
\multirow{2}{*}{\llama} 
  & zero & 0.83 & 0.11 & 0.47 & 0.43 & 0.36 & 0.39 & 0.12 & 0.08 & 0.64 & 0.23 & 0.64 & 0.34 \\
  & few  & 0.93 & 0.02 & 0.48 & 0.76 & 0.72 & 0.74 & 0.22 & \textbf{0.44} & 0.54 & 0.30 & 0.59 & 0.42 \\
\midrule
\multirow{2}{*}{\anita} 
  & zero & 0.84 & 0.04 & 0.44 & 0.73 & 0.40 & 0.57 & 0.08 & 0.02 & 0.18 & 0.14 & 0.60 & 0.21 \\
  & few  & 0.76 & 0.06 & 0.41 & 0.70 & 0.51 & 0.60 & 0.08 & 0.31 & 0.24 & 0.24 & 0.55 & 0.28 \\
\midrule
\multirow{2}{*}{\minerva} 
  & zero & 0.00 & 0.00 & 0.00 & 0.65 & 0.49 & 0.57 & 0.06 & 0.22 & 0.21 & 0.00 & 0.29 & 0.16 \\
  & few  & 0.46 & 0.00 & 0.23 & 0.66 & 0.29 & 0.48 & 0.17 & 0.36 & 0.42 & 0.22 & 0.17 & 0.27 \\
\bottomrule
\end{tabular}
}
\end{table*}

\paragraph{Performance across models}
Table~\ref{tab:amelia_results} summarises the performance of the four evaluated models (\llama{}, \llamalarge{}, \minerva{} and \anita{}) across the three AMELIA subtasks.
The best-performing model for all tasks is the largest one, i.e., \llamalarge{} in few-shot mode. 
It surpasses or matches any other model also in zero-shot mode.
It is noteworthy that no other model reaches a macro F1 score above 0.50 in the first and third tasks. 
The second best model is \llama{} used in few-shot mode.

\paragraph{Error Analysis}
Smaller models often do not follow the instructions, generating text similar to few-shots examples, repetition of the expected answers in various formats (e.g., ``prem/conc?''), or a non-answer. 
Focusing on the best-performing model, in the first task, some errors are understandable given the ambiguity of the input. Consider the misclassification of conclusions starting with ``\textit{Preliminarmente}'', or premises that contain expressions like ``\textit{in definitiva}'', which seems to indicate the conclusion of an argument.
However, further errors involve inputs that are more straightforward, such as the conclusion ``\textit{Va infine accolto l'appello incidentale}''. 
As regards premise type classification, the model often assigns both labels in ambiguous contexts. 
Another common issue concerns misclassifying normative references as purely factual or legal, especially when abstract norms are applied to concrete scenarios. 
Consider the premise ``\textit{Una statuizione di tal tipo comporta la non opponibilità del provvedimento giurisdizionale} [...]''. 
The model correctly predicts its legal nature, but fails to capture the factual implication of the legal decision. 
Concerning the identification of argumentative schemes, 
the model often misinterprets a normative inference as an interpretative (`Itpr') one. For example, ``\textit{Come già evidenziato in primo grado, il fatto che} [...]'', is a rule-based conclusion and not an interpretative argument.
The model also struggles to distinguish between reliance on precedent ('Prec') and the derivation of rules, whenever a decision with normative implications is cited. For example, ``\textit{Ed in seguito ai chiarimenti della nota sentenza del 2011}[...]'', is uncorrectly predicted as both 'Prec' and 'Rule'.
The first task, expected to be the easiest, appears to be challenging for most models, especially the identification of conclusions.
This is surprising, given that simpler models based on lexical features, such as LinearSVC, obtained better results on similar tasks \cite{grundler-etal-2022-detecting}.
In future work, to address misaligned outputs, we could experiment with alternative versions of the prompts.

\subsubsection{BEEP - BEst DrivEr's License Performer}

\posscite{DBLP:conf/clic-it/MercorioPSS24} challenge is designed to evaluate LLMs' abilities in a simulated Italian driver's license exam. The task tests both factual knowledge and reasoning skills applied to realistic driving scenarios, requiring understanding of traffic laws, road signs, driving behaviour, and vehicle maintenance. Derived from official ministerial materials, BEEP focuses on the theoretical exam for Category B licenses, which is required for operating cars weighing up to 3.5 tons and with a seating capacity of eight. In Italy, obtaining a driver’s license involves strict theoretical and practical assessments regulated by the Ministero delle Infrastrutture e dei Trasporti. The theoretical exam consists of 30 multiple-choice questions, with a maximum of three errors allowed. Beyond rote memorisation, it demands the practical application of rules, particularly important in complex urban traffic contexts. BEEP mirrors this complexity by presenting a series of true/false questions aimed at assessing LLMs' ability to reason about safety, compliance, and real-world driving challenges, aligning with the high standards of the Italian and European licensing systems.
While LLMs have demonstrated strong language understanding, their effectiveness in practical decision-making scenarios, particularly in Italian, remains less explored. Assessing their performance in this structured yet complex setting helps determine their capability to generalise language understanding to real-world applications.

\paragraph{Performance across models}

\begin{table}[!htb]
\footnotesize
\centering
\caption{Results for the \textsc{BEEP} task. Model accuracy (\%) across the main thematic categories of the Italian driver’s license exam.}
\label{table:beep_accuracy}
\resizebox{\linewidth}{!}{%
\begin{tabular}{lcccc}
\toprule
\textbf{Category} & \textbf{\llama} & \textbf{\llamalarge} & \textbf{\minerva} & \textbf{\anita} \\
\midrule
DOCUMENTS              & 54.41\% & 75.48\% & 54.02\% & 51.34\% \\
VEHICLE EQUIPMENT      & 60.53\% & 76.32\% & 51.97\% & 59.87\% \\
VEHICLES               & 67.92\% & 83.96\% & 56.60\% & 62.26\% \\
THE MOTOR VEHICLE      & 64.82\% & 88.93\% & 60.08\% & 59.68\% \\
ACCIDENTS AND INSUR.& 69.23\% & 92.60\% & 46.30\% & 67.05\% \\
THE ROAD               & 58.13\% & 76.35\% & 54.19\% & 57.64\% \\
RULES OF CONDUCT       & 61.63\% & 80.04\% & 50.83\% & 59.65\% \\
FIRST AID              & 70.83\% & 86.46\% & 54.17\% & 72.92\% \\
ROAD SIGNAGE           & 50.00\% & 75.00\% & 50.00\% & 50.00\% \\
SAFETY AND POLLUT.   & 80.41\% & 91.84\% & 51.43\% & 76.33\% \\
\midrule
\textbf{Overall}       & \textbf{64.83\%} & \textbf{84.25\%} & \textbf{51.51\%} & \textbf{62.43\%} \\
\bottomrule
\end{tabular}
}
\end{table}

\begin{table}[!ht]
\centering
\footnotesize
\caption{Results for the \textsc{BEEP} task under the simulated Italian driver’s license exam. The table reports the percentage of tests passed (out of 1000 simulations) and the average number of errors per test (with standard deviation).}
\label{table:beep_driving_metrics}
\renewcommand{\arraystretch}{1.2}
\begin{tabular}{@{}l rr@{}}
\toprule
\textbf{Model} & \textbf{Total Tests Passed (\%)} & \textbf{Avg. Errors (Std.)} \\
\midrule
\textbf{\llama}      & 5/1000 (0.5\%)    & 10.60 ($\pm$2.61) \\
\textbf{\llamalarge} & 272/1000 (27.2\%) &  4.80 ($\pm$1.94) \\
\textbf{\minerva}    & 0/1000 (0.0\%)    & 14.64 ($\pm$2.74) \\
\textbf{\anita}      & 1/1000 (0.1\%)    & 11.20 ($\pm$2.69) \\
\bottomrule
\end{tabular}
\end{table}

Table \ref{table:beep_accuracy} presents the Overall Accuracy of the four evaluated models (\llama{}, \llamalarge{}, \minerva{} and \anita{}).
The scaling laws hold as performance improves with an increasing number of parameters. 
Models achieve higher accuracy in the ``SAFETY AND POLLUTION", ``FIRST AID," and ``ACCIDENTS AND INSURANCE" categories. This could be due to the broader nature of these categories compared to more specialized ones like "DOCUMENTS" or "VEHICLE EQUIPMENT," where performance is lower.
We also evaluated the models by simulating an official driving license exam, following official guidelines and introducing another performance indicator. From the dataset, we created 1,000 unique samples of 30 questions each and measured the pass rate, considering a test successful if the model made three or fewer mistakes. As shown in Table \ref{table:beep_driving_metrics}, smaller models consistently failed, averaging around 10 errors per test, while even larger models like \llamalarge{} struggled. However, we believe more advanced models, such as GPT-4o, could achieve better results.

\subsubsection{BLM-It - Blackbird Language Matrices}

\posscite{DBLP:conf/clic-it/JiangSNM24} challenge comprises linguistic puzzles developed in analogy to the visual Raven Progressive Matrices tests, a psychometric test of intelligence. BLMs are developed to test the representation and knowledge of abstract linguistic rules in LLMs \citep{merlo2023,an-etal-2023-blm,samo-etal-2023-blm,nastase2024exploring1,nastase2024exploring2}. A BLM instance consists of a context set and an answer set. The context is a sequence of sentences that extentionally encode a linguistic rule. They encode, for example,  the rule of grammatical number concord: subject and verb  agree in their grammatical number independently of their linear distance.  To solve the BLM, one must select, given the context,  the missing sentence to complete the sequence. Beside the unique correct answer, the multiple-choice answer set includes minimally contrastive   negative examples across different dimensions. The datasets present syntactic and semantic problems in Italian (subject-verb agreement, the causative verb alternation, the object-drop alternation), with a few prompts for a few-shot setting.  By their construction, BLM datasets are richly structured at multiple levels:  within each sentence, across the input sequence,  and within each candidate answer. This structure supports both linguistic and computational investigations, such as testing whether models encode abstractions in their embeddings; linguistic abstraction at the lexical level, or language independent linguistic rules.
We evaluate LLM performances on three BLM-It tasks: subject-verb agreement (Agr), causative alternation (Caus), and object-drop alternation (Od), across different dimensions of lexical variation (Type II, Type III; see details in \citealt{DBLP:conf/clic-it/JiangSNM24}).

\begin{table}[!t]
\centering
\footnotesize
\caption{Results for the \textsc{BLM-It} task. F1 scores of the evaluated models across the three subtasks (Agreement, Causative, Object-drop), averaged over Type-II and Type-III data, under zero-shot and one-shot settings.}
\label{tab:model_f1}
\begin{tabular}{llcc}
\toprule
\textbf{Task} & \textbf{Model} & \textbf{0-shot} & \textbf{1-shot} \\
\midrule
\multirow{4}{*}{\textbf{Agreement}} 
  & \llamalarge & 0.339 & 0.409 \\
  & \llama      & 0.104 & 0.248 \\
  & \minerva    & 0.026 & 0.083 \\
  & \anita      & 0.170 & 0.242 \\
\midrule
\multirow{4}{*}{\textbf{Causative}} 
  & \llamalarge & 0.086 & 0.186 \\
  & \llama      & 0.054 & 0.131 \\
  & \minerva    & 0.027 & 0.090 \\
  & \anita      & 0.058 & 0.102 \\
\midrule
\multirow{4}{*}{\textbf{Object-drop}} 
  & \llamalarge & 0.071 & 0.234 \\
  & \llama      & 0.063 & 0.120 \\
  & \minerva    & 0.028 & 0.093 \\
  & \anita      & 0.057 & 0.094 \\
\bottomrule
\end{tabular}
\end{table}

\begin{figure}[t]
    \centering
    \begin{minipage}[b]{0.32\linewidth}
        \includegraphics[width=\linewidth]{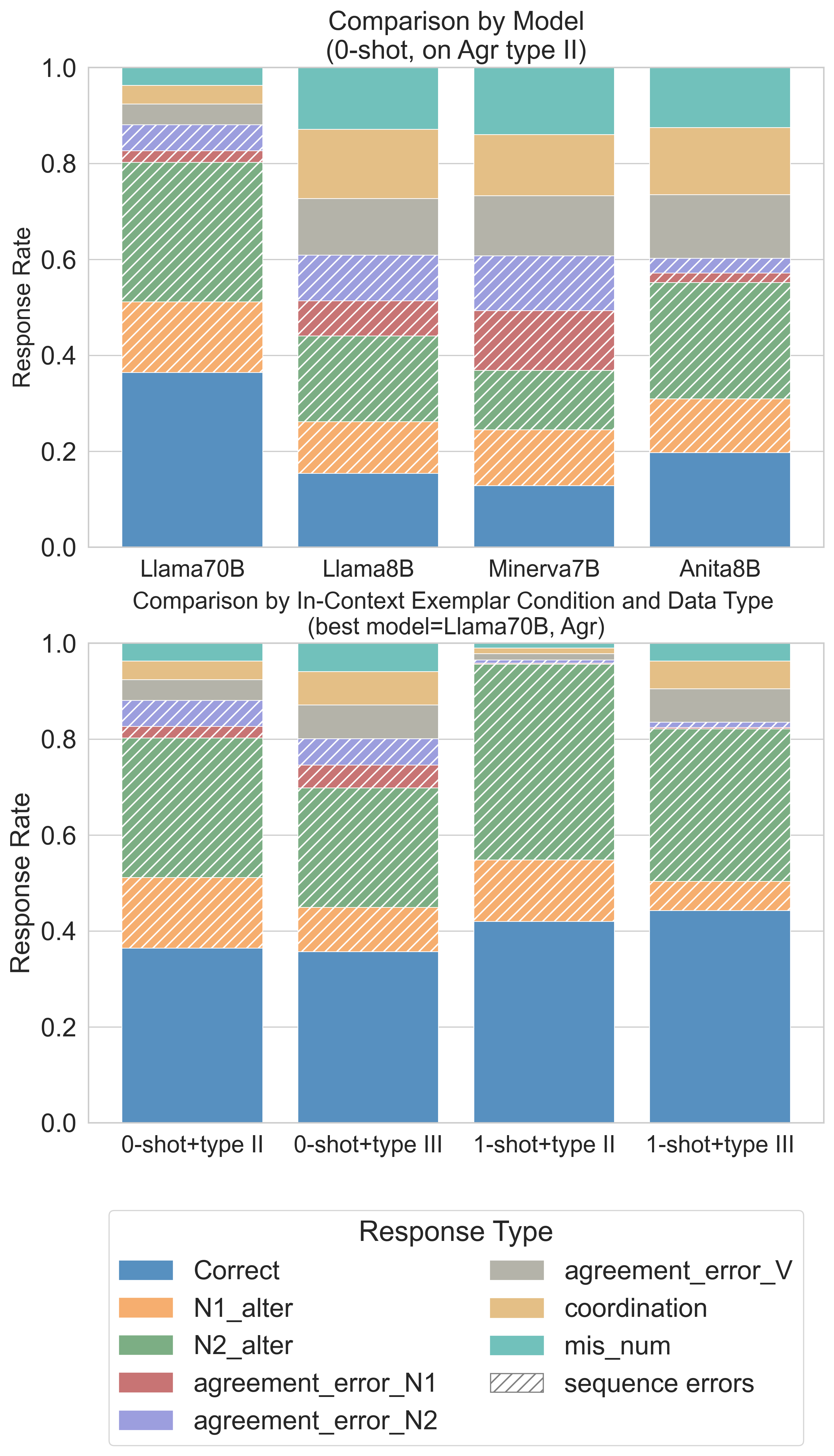}
        \centering
        (a) Agreement
        \label{fig:agr_error_dist}
    \end{minipage}
    \hfill
    \begin{minipage}[b]{0.32\linewidth}
        \includegraphics[width=\linewidth]{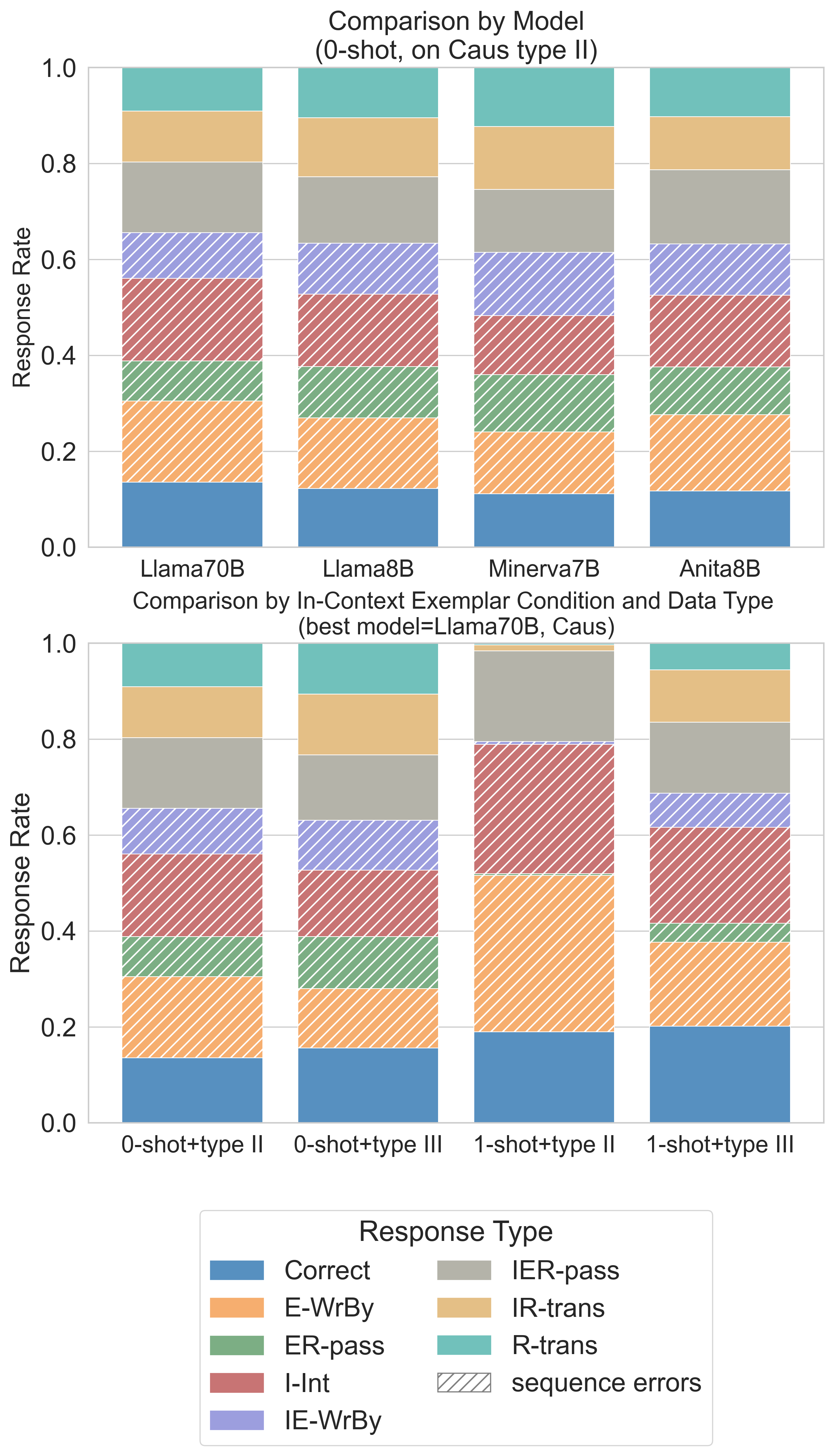}
        \centering
        (b) Causative
        \label{fig:caus_error_dist}
    \end{minipage}
    \hfill
    \begin{minipage}[b]{0.32\linewidth}
        \includegraphics[width=\linewidth]{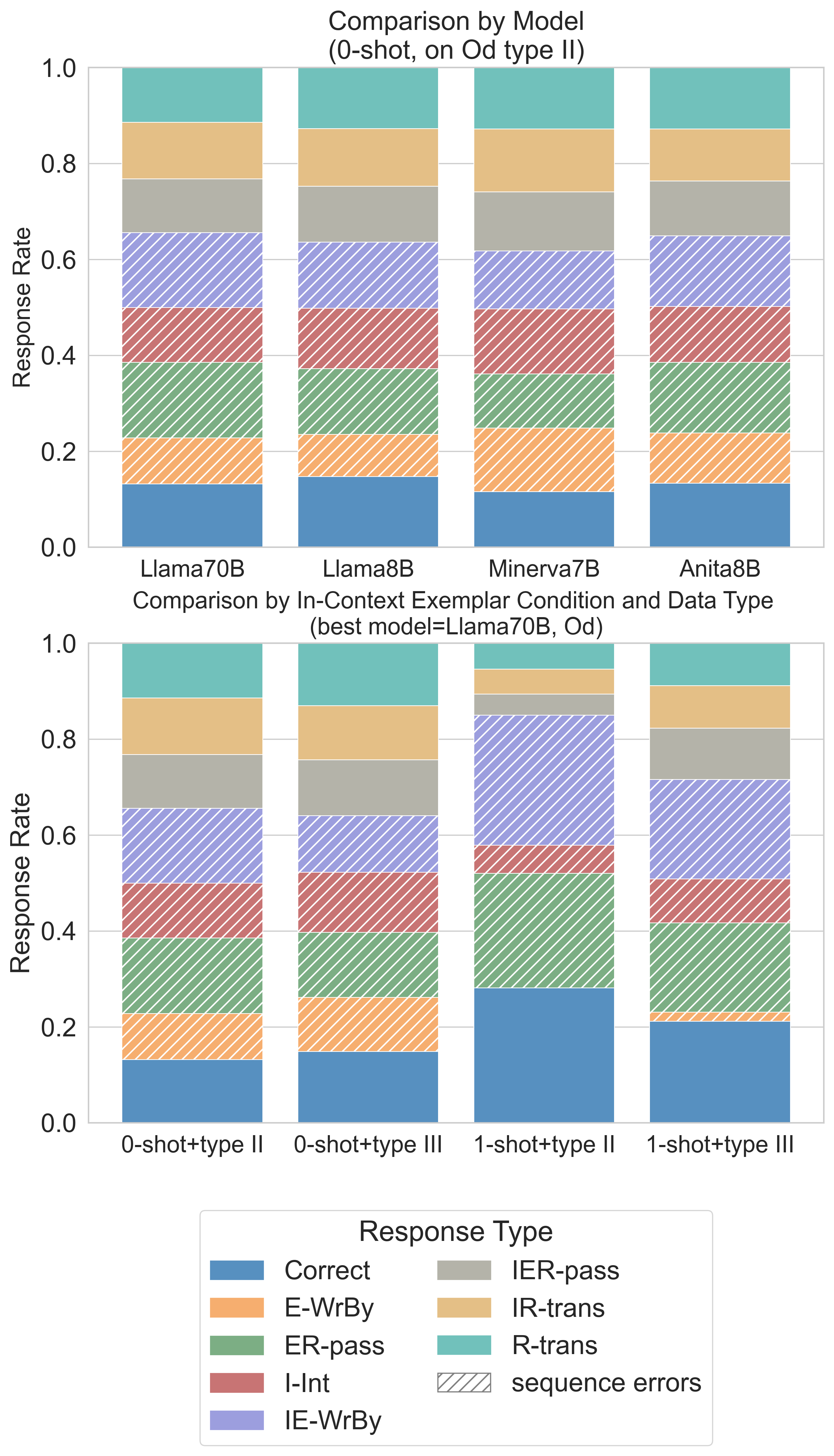}
        \centering
        (c) Object-drop
        \label{fig:od_error_dist}
    \end{minipage}

    \caption{Response distributions for the \textsc{BLM-It} task across the three subtasks: (a) Agreement, (b) Causative, and (c) Object-drop.}
    \label{fig:blm-it-results}
    \vspace{-0.6cm}
\end{figure}

\paragraph{Performance across models}
Performance results of the four tested models  (\llama{}, \llamalarge{}, \minerva{} and \anita{}) are shown in Table \ref{tab:model_f1}, with error distributions illustrated in Figure~\ref{fig:blm-it-results}. Overall, performance is at or below the baseline level, but several general patterns emerge. Model size has a clear effect, with \llamalarge{} demonstrating the highest performance. Few-shot learning shows varying efficacy across tasks, with the agreement task benefiting the most.  Italian-optimized models underperform general-purpose models of comparable scale, with \anita{} performing worse than \llama{} in five out of six experimental conditions.

\paragraph{Error Analysis}
The most frequent agreement errors relate to the sequence structure of the BLM rather than morpho-syntactic agreement. For verb alternations, the zero-shot setting produces a relatively balanced distribution of errors. In contrast, in the one-shots setting, two prominent sequence errors encode forms of agentivity preference \citep{huber2024surprisal}. Overall the models master syntax and the syntax-semantics interface, but fail in the understanding of more general paradigm expressed by the BLM template.

\subsubsection{DIMMI - Drug InforMation Mining in Italian}

\posscite{DBLP:conf/clic-it/MannaBG24} challenge aims at evaluating the proficiency of Large Language Models in extracting drug-specific information from Patient Information Leaflets (PILs). The challenge seeks to advance the understanding of effectiveness in processing complex medical information in Italian, and to enhance drug information extraction and pharmacovigilance efforts. The challenge provides a dataset of 600 Italian PILs, derived from the D-LeafIT Corpus\footnote{\url{https://github.com/unior-nlp-research-group/D-LeafIT}} \cite{giordano-di-buono-2024-large}, made up of 1819 Italian drug package leaflets extracted from the Italian Agency for Medications (Agenzia Italiana del Farmaco - AIFA) website\footnote{\url{https://www.aifa.gov.it/en/home}}. The objective is to the models' ability to accurately answer specific questions related to drug dosage, usage, side effects, drug-drug interactions. For each drug in the dataset, we evaluate the results from two types of zero-shot prompts in Italian: specific task-focused prompts (P1) and structured prompts (P2). P1 is composed of five questions for each of the information types we want to extract. P2 requires the simultaneous extraction of all five categories with a specific instruction to help the model understand the expected output structure and facilitate the extraction process.
The answers generated by the models are compared against the gold standard (GS), created to establish a reliable, accurate, and a comprehensive set of answers. For each drug and each information category, the GS contains the correct information extracted from the leaflets through a manual annotation.

\begin{table}[!ht]
\footnotesize
\centering
\caption{
Results for the \textsc{DIMMI} task. Evaluation under two prompting strategies (P1 and P2). Scores are reported for overall F1 and for each information category: Molecule, Usage, Side effect, Posology, and Drug interaction.
}
\label{tab:dimmi_results}
\renewcommand{\arraystretch}{1.1}
\begin{tabular}{l c c c c c c c}
\toprule
\textbf{Model} & \textbf{P} & \textbf{F1} & \textbf{Molecule} & \textbf{Usage} & \textbf{Side effect} & \textbf{Posol.} & \textbf{Drug inter.} \\
\midrule
\textbf{\anita}      & P1 & --   & 0.64 & 0.12 & 0.07 & 0.16 & 0.21 \\
\textbf{\llama}      & P1 & --   & 0.86 & 0.19 & 0.16 & 0.17 & 0.43 \\
\textbf{\llamalarge} & P1 & --   & 0.87 & 0.19 & 0.15 & 0.18 & 0.39 \\
\textbf{\minerva}    & P1 & --   & 0.01 & 0.08 & 0.01 & 0.17 & 0.12 \\
\midrule
\textbf{\anita}      & P2 & 0.33 & 0.84 & 0.15 & 0.08 & 0.14 & 0.45 \\
\textbf{\llama}      & P2 & 0.31 & 0.76 & 0.22 & 0.05 & 0.15 & 0.36 \\
\textbf{\llamalarge} & P2 & 0.34 & 0.87 & 0.16 & 0.06 & 0.20 & 0.40 \\
\textbf{\minerva}    & P2 & 0.07 & 0.00 & 0.07 & 0.00 & 0.16 & 0.11 \\
\bottomrule
\end{tabular}
\end{table}

\paragraph{Performance across models}
The four models (\llama{}, \llamalarge{}, \minerva{} and \anita{}) were evaluted under both prompting strategies, P1 and P2 (Table~\ref{tab:dimmi_results}), using accuracy, precision, recall and F-1 score against the manually created dataset. 
\llamalarge{} outperforms all the other models in all categories (in some cases with a very slight difference), except for \textit{Drug\_interaction}, where \llama{} performs better in P1 and \anita{} in P2. On the other hand, \minerva{} consistently failed to extract information for the \textit{Molecule} and \textit{Side\_effect} categories in both settings. These two categories differ notably: \textit{Molecule} involves short, specialized terms, making it easier to retrieve, while \textit{Side\_effect} includes more varied and often colloquial expressions, increasing extraction difficulty.
Prompt type had no effect on \minerva’s performance, while \anita{} improved in most categories with P2. In contrast, \llama{} performed better in P1, except for \textit{Usage}, where it showed a slight improvement in P2 (0.22 vs. 0.18). In P1, \llama{} consistently outperformed \anita{}, particularly in \textit{Molecule}, \textit{Side\_effect}, and \textit{Drug\_Interaction}. However, in P2, \anita{} reversed this trend, outperforming \llama{} in those same categories.
Both models showed their weakest performance on \textit{Side\_effect} in both settings, suggesting intrinsic challenges in that category. Conversely, their best scores were achieved in \textit{Molecule}-Anita in P2, \llama{} in P1. \minerva{} reached its peak performance in Posology under both prompts, matching \llama’s best score in P1 (0.16).
Overall, \anita{} benefits from joint-category prompts (P2), while \llama{} performs better with focused, category-specific prompts (P1), except in \textit{Posology}, where results are comparable. \minerva’s failure across categories may stem from training limitations or poor compatibility with the task's prompt format.

\paragraph{Error Analysis}
A qualitative analysis of selected outputs revealed recurring issues across models. Some responses contained additional, often irrelevant content due to overextended extraction (e.g., ID 147 - Anita - Posology). Others were marked incorrect due to language interference, such as translations or use of synonyms in the wrong language (e.g., ID 556 - Anita - \textit{Molecule}, the model output is irbesartan, hidroclorotiazida instead of the expected irbesartan, idroclorotiazide).
Other mismatches arose from inconsistent measurement units (e.g., ID 242 - Anita - \textit{Posology}, the model produces \textit{80 mg al giorno, 120 mg al giorno}, whereas the reference answer is \textit{una compressa al giorno}) or failure to capture multiword expressions (e.g., ID 140 - Anita - \textit{Molecule}, the model returns \textit{tamsulosina}, while the correct answer is \textit{tamsulosina cloridrato}). Variations in specificity were also noted, with some outputs being overly generic or, conversely, including unnecessary details.
Our analysis also indicates that the models exhibit a tendency to extract layout-based structures, such as bullet points, as undifferentiated blocks rather than parsing the individual, semantically distinct items they contain. This behavior points to a limitation in the models' ability to segment and interpret information accurately when it is presented in visually cohesive formats.
In conclusion, these findings expose systematic limitations in model performance, including issues with language handling, term granularity, and expression consistency. They highlight the need for improved prompt design and evaluation metrics, particularly in specialized domains requiring high terminological precision.

\subsubsection{ECWCA - Educational CrossWord Clues Answering}
\posscite{DBLP:conf/clic-it/ZugariniZFZ24} challenge is designed to evaluate the knowledge and reasoning capabilities of LLMs through crossword clue-answering. The challenge consists of two tasks: a standard question-answering format where the LLM is asked to solve crossword clues and a variation where the model is given hints about the word lengths of the answers, which is expected to help models with reasoning abilities. Clue-answer pairs were generated following clue-instruct \cite{DBLP:conf/coling/ZugariniZKMGR24}. Clue-instruct produces three different clues for each given answer and its context. In order to keep at most one clue per answer, a human selection step was added. Examples were constructed from the most visited Italian Wikipedia pages. To guarantee high quality clues, annotators were asked to cross-verify the factual adherence of the chosen clue-answer pairs. Moreover, to better assess factual knowledge of LLMs, annotators were instructed to select clues that provided enough information to infer the correct answer with confidence.

\paragraph{Performance across models} 
From the evaluation of the four models (\llama{}, \llamalarge{}, \minerva{} and \anita{}) we see that performances are strongly influenced by two factors: model size and native pre-training. Notably, the character-length hint is generally not beneficial, with mixed results across models.
The only model exhibiting distinctly anomalous behavior was \anita, which often produced outputs with non-standard formatting or irrelevant content, including non-textual characters or unrelated facts. This suggests that native fine-tuning did not improve the factual knowledge of the LLM. On the contrary, native pre-training seems beneficial, since \minerva{} exhibits better accuracy than \llama{}, being only slightly inferior to \llamalarge, a ten times larger model. 

\begin{figure}[ht]
    \centering
    \includegraphics[width=0.49\linewidth]{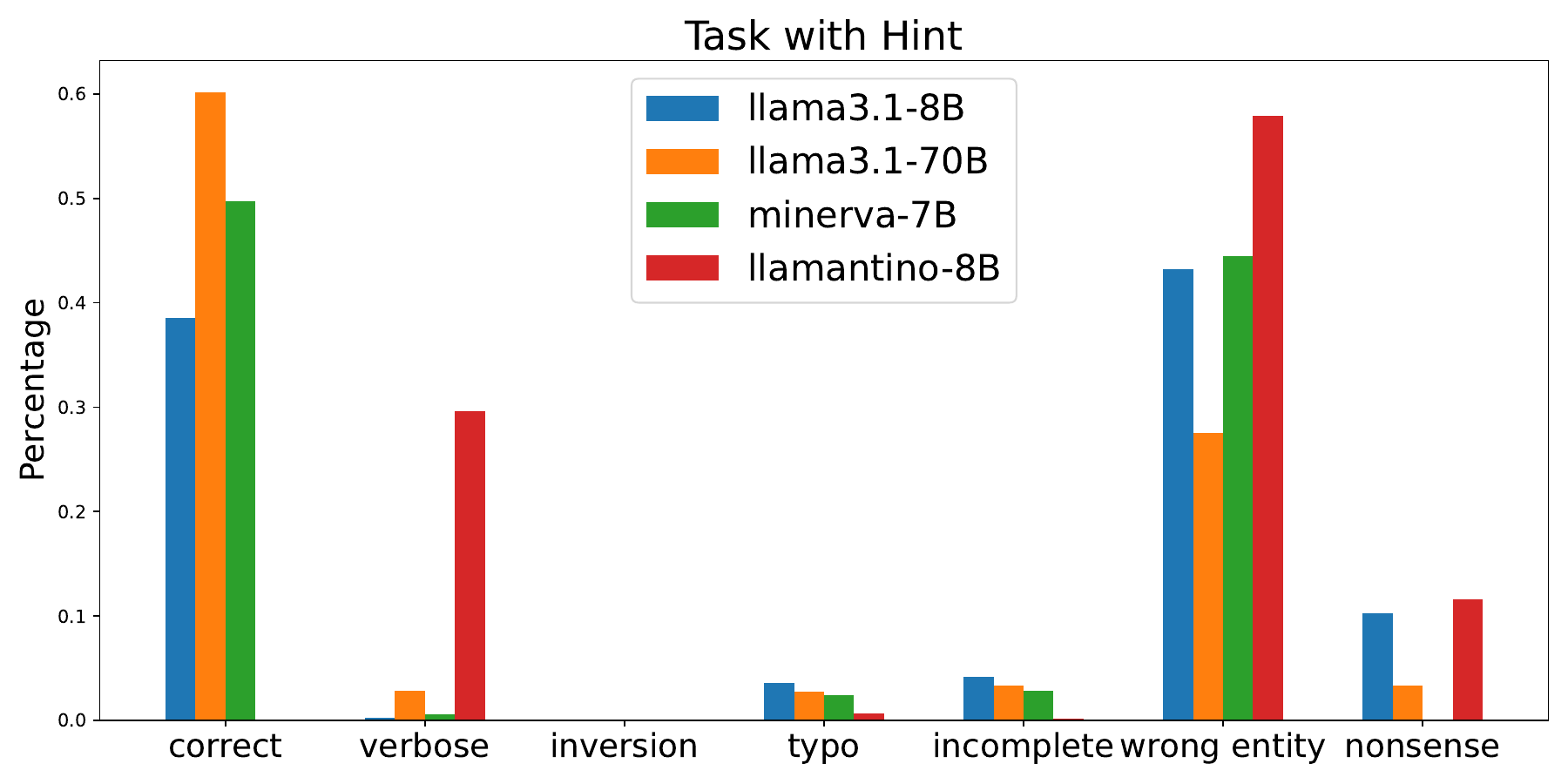}
    \includegraphics[width=0.49\linewidth]{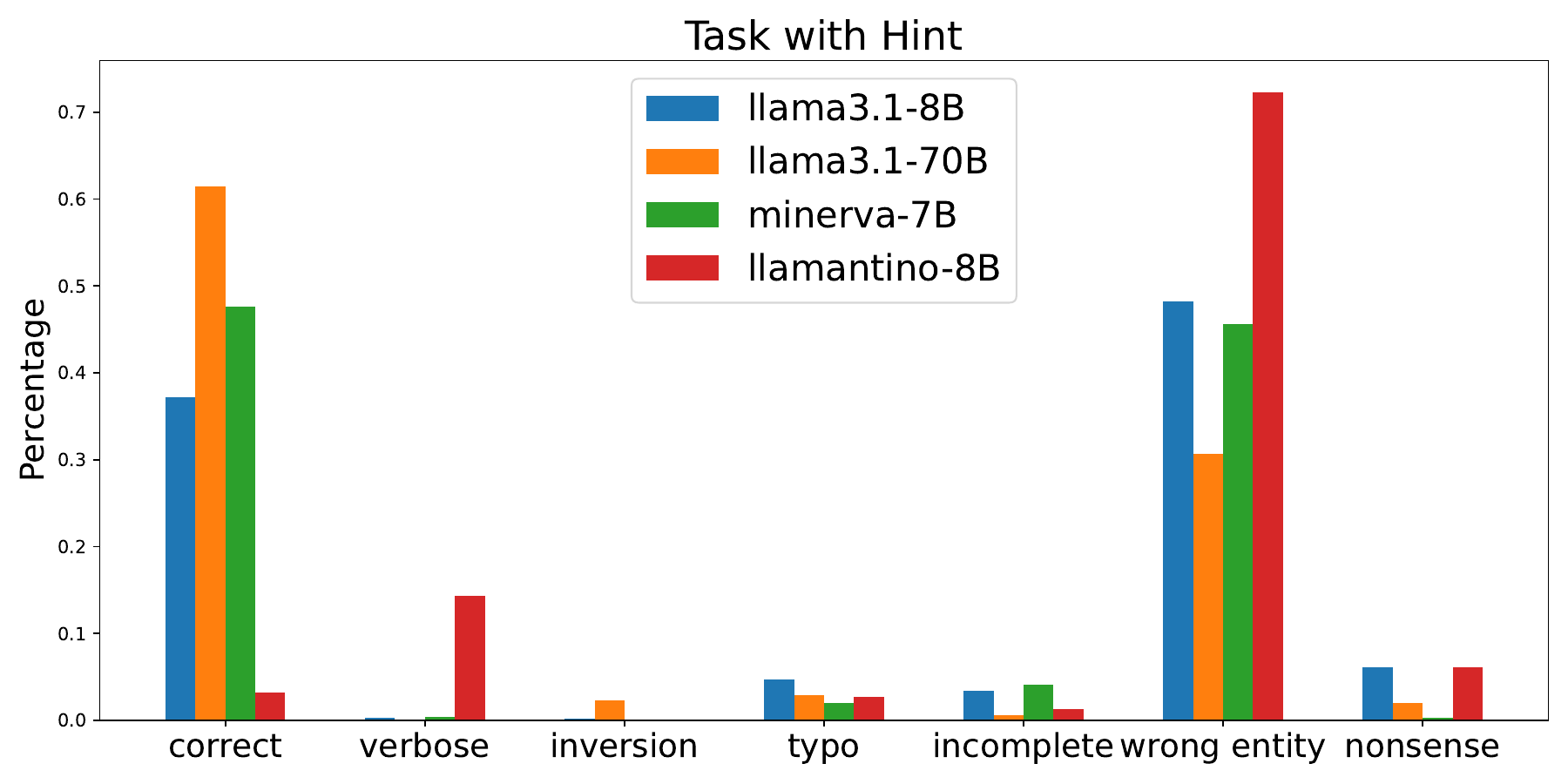}
    \caption{
        Results for the \textsc{ECWCA} task. Distribution of error types across models in the hint and no-hint settings. Categories include \texttt{verbose}, \texttt{typo}, \texttt{inversion}, \texttt{incomplete}, and \texttt{wrong entity}. See Table~\ref{tab:ecwca_error_examples} for some error type examples. Correct predictions were marked as \texttt{correct}.
        }
    \label{fig:ecwca_error_types_distribution}
\end{figure}

\paragraph{Error Analysis}
The vast majority of mistakes fall into the \texttt{wrong entity} group. Confused entities are often plausible or at least contextually related answers. Sometimes errors stem from queries about events postdating the model's knowledge cut-off, as shown in the example of errors in Table~\ref{tab:ecwca_error_examples}.
There are also errors from cultural mismatches in non-native models, where the predicted answer reflects a different cultural bias. For example, in the clue ``La sua canzone Gloria è entrata nella classifica dei dischi più venduti in Gran Bretagna e negli Stati Uniti'', where \llamalarge{} predicted ``\textit{Laura Branigan}'' instead of the Italian singer ``\textit{Umberto Tozzi}''.

\begin{table}[!ht]
\caption{
Examples of model errors in the \textsc{ECWCA} task, showing representative clues, target answers, predicted answers, and corresponding error types.
}
\footnotesize
\centering
\renewcommand{\arraystretch}{1.1}
\resizebox{\linewidth}{!}{%
\begin{tabular}{p{4.7cm} m{2.4cm} m{2.4cm} m{2cm}}
\toprule
\textbf{Clue} & \textbf{Target Ans.} & \textbf{Predicted Ans.} & \textbf{Error} \\
\midrule
Comune toscano con importante porto commerciale e turistico  & Livorno & Livorno''') & \texttt{verbose} \\
Squalificato per aver morso l'orecchio di Evander Holyfield & Mike Tyson & Tyson Mike & \texttt{inversion} \\
Controparte pasquale del panettone e del pandoro 
  & Colomba Pasquale & Colomba & \texttt{incomplete} \\
Il conduttore di \emph{Lascia o raddoppia?} e \emph{Rischiatutto} 
  & Mike Bongiorno & Mike Buongiorno & \texttt{typo} \\
Trionfatrice al 74-esimo Festival di Sanremo con il brano \emph{La noia} 
  & Angelina Mango & Annalisa Scarrone & \texttt{wrong entity} \\
Attaccante del Napoli e della nazionale belga 
  & Romero Lukaku & Mertens & \texttt{wrong entity} \\
\bottomrule
\end{tabular}
}
\label{tab:ecwca_error_examples}
\end{table}

\subsubsection{\textsc{EurekaRebus} - Verbalized Rebus Solving with LLMs}
\posscite{DBLP:conf/clic-it/SartiCBN24} challenge tests the ability of LLMs to conduct multi-step, knowledge-intensive inferences while respecting predefined constraints. 
The challenge employs text-only variants of \textit{rebus} word games, which have a long-standing tradition in the Italian puzzle-solving community~\citep{enimmistica,miola-rebus,loradesiatavola}. While turning visuals into textual descriptions simplifies the game, making it akin to crossword puzzles, the task remains challenging, requiring multi-step, knowledge-intensive reasoning under predefined formal constraints, such as a specific number of letters. \citet{sarti-etal-2024-non} previously showed extremely low accuracy in predicting the final solution (Exact Match) for prompted open-source multilingual LLMs. Since most errors originated from a wrong resolution of definitions, the task was deemed interesting for the CALAMITA campaign to assess the performance of models with an Italian-focused training such as \anita{} and \minerva{}.
In \textsc{EurekaRebus}, LLMs are prompted to reason step-by-step to solve verbalized variants of rebus games. Verbalized rebuses replace visual cues with crossword definitions to create an encrypted first pass, making the problem entirely text-based. Multiple metrics are used to grasp the models’ performance in knowledge recall, constraints adherence, and re-segmentation abilities across reasoning steps.

\begin{table*}[!ht]
    \begin{minipage}{.67\linewidth}
        \footnotesize
        \caption{
        Results for the \textsc{EurekaRebus} task. Evaluation of model accuracy (\textbf{Acc.}), 
        predicted word length (\textbf{Len.}), and exact match (\textbf{EM}) for first-pass and 
        final solution predictions, under prompts without (top) and with (bottom) length hints.
        The accuracy values reported in Appendix~\ref{tab:full-results} correspond to the 
        First-Pass Accuracy (FP Acc.) in the two settings.
        }
        \label{tab:eurekarebus-results}

        \begin{tabular}{lcccccc}
        \toprule
        \multirow{2}{*}{\textbf{Model}} &
        \multicolumn{3}{c}{\textbf{First Pass (FP)}} &
        \multicolumn{3}{c}{\textbf{Solution}} \\
        \cmidrule(lr){2-4} \cmidrule(lr){5-7}
        & \textbf{Acc.} & \textbf{Len.} & \textbf{EM}
        & \textbf{Acc.} & \textbf{Len.} & \textbf{EM} \\
        \midrule

        \multicolumn{7}{l}{\textit{Without length hints}} \\[-2pt]
        \textbf{\llama}      & 0.10 & 0.18 & 0.00 & 0.02 & 0.17 & 0.00 \\
        \textbf{\llamalarge} & \textbf{0.36} & 0.42 & \textbf{0.07} & \textbf{0.08} & \textbf{0.26} & 0.00 \\
        \textbf{\anita}      & 0.00 & 0.01 & 0.00 & 0.00 & 0.00 & 0.00 \\
        \textbf{\minerva}    & 0.00 & 0.00 & 0.00 & 0.00 & 0.01 & 0.00 \\
        \midrule

        \multicolumn{7}{l}{\textit{With length hints}} \\[-2pt]
        \textbf{\llama}      & 0.07 & 0.16 & 0.00 & 0.01 & 0.09 & 0.00 \\
        \textbf{\llamalarge} & 0.32 & \textbf{0.49} & 0.04 & 0.06 & 0.21 & \textbf{0.01} \\
        \textbf{\anita}      & 0.00 & 0.01 & 0.00 & 0.00 & 0.00 & 0.00 \\
        \textbf{\minerva}    & 0.00 & 0.00 & 0.00 & 0.00 & 0.00 & 0.00 \\
        \bottomrule
        \end{tabular}
    \end{minipage}
    \hspace{.03\textwidth}
    \begin{minipage}{.27\linewidth}
        \centering
        \includegraphics[width=0.95\linewidth]{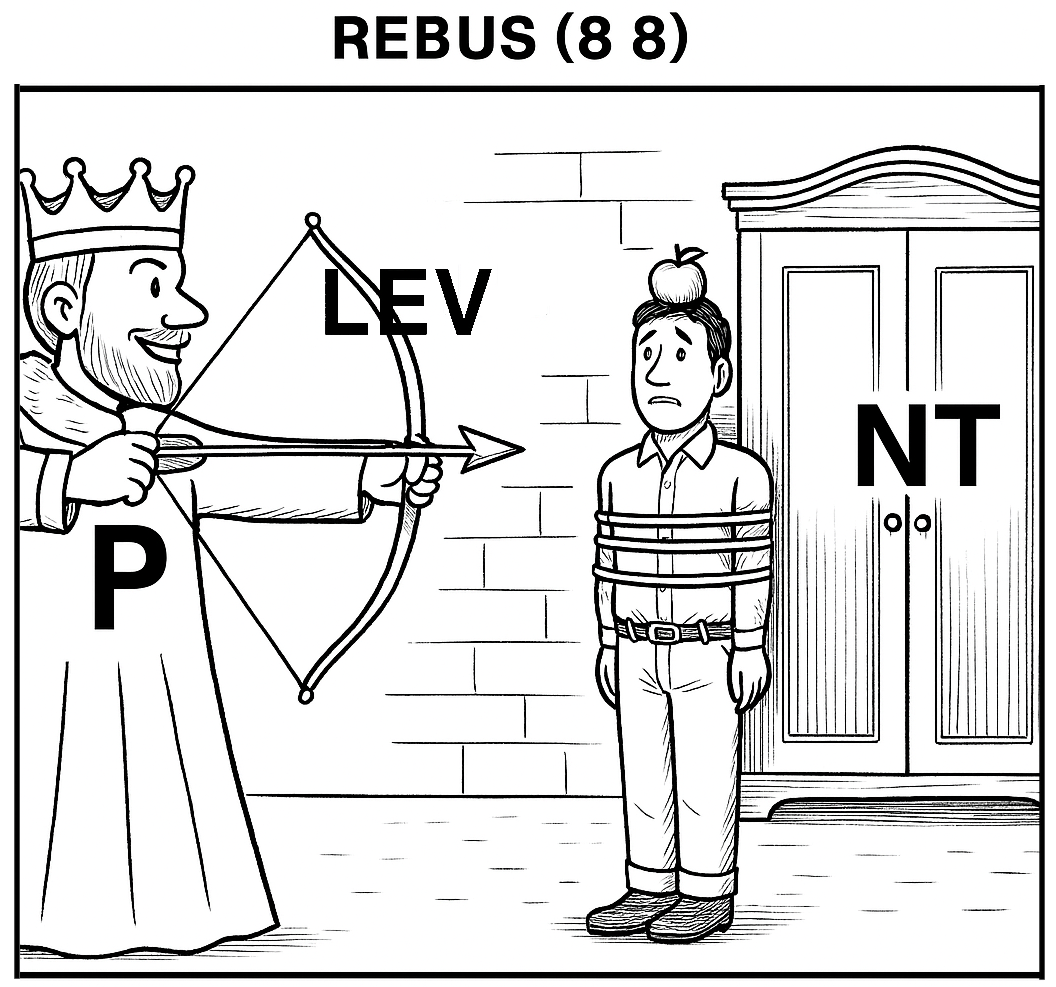}\\[4pt]
        {\small Example of multimodal rebus. FP: P \textit{re} LEV \textit{arco} NT \textit{ante}. 
        Solution: \textit{Prelevar contante}.}
    \end{minipage}
\end{table*}

\paragraph{Performance across models} 
Table~\ref{tab:eurekarebus-results} reports metric scores for the four models (\llama{}, \llamalarge{}, \anita{} and \minerva{}) across the two tested settings. Results confirm that the task remains unsolved by current open-source models, with the best solution accuracy (Exact Match) of 1\%. Italian-trained models are also found to struggle with all the task phases compared to the multilingual LLaMA model of comparable size. Major gains are obtained by increasing the parameter count to 70B, confirming the importance of model scale in complex reasoning tasks. \llamalarge{} is the only tested model using length hints in the prompt to resolve FP words (FP Len. 0.49 vs. 0.42 without hints), although this does not lead to better definition resolution. Still, all tested models struggle with respecting the given reasoning template. 

\paragraph{Error Analysis}
As shown by~\citet{sarti-etal-2024-non}, models can disregard resolved words to generate potentially more fluent solutions. In the following example, \llamalarge{} correctly resolves the first pass, but hallucinates a final solution (optional length hints in blue, Table~\ref{tab:eurekarebus-results} shows a multimodal version):

\begin{quote}
\vspace{-12pt}

\noindent
\textbf{Rebus:} P -- \textit{Il vertice della nobiltà} \textcolor{blue}{(2)} -- LEV -- \textit{Quello di Trionfo si trova a Parigi} \textcolor{blue}{(4)} -- NT -- \textit{Portiere d' armadio} \textcolor{blue}{(4)} \\
\textbf{Solution key:} 8 8 \\
\textbf{First pass:} P \textit{re} LEV \textit{arco} NT \textit{ante} (English: P \textit{king} LEV \textit{arch} NT \textit{doors}) \\
\textbf{Model prediction:} Per levar antenne (English: For removing antennas) \\
\textbf{Correct solution:} Prelevar contante (English: Withdraw cash)
\vspace{-5pt}
\end{quote}
Our results confirm that word games can be a promising testbed for evaluating the knowledge and cultural understanding of LLMs while also testing their sequential instruction following skills. Future work in this area could expand our evaluation to multimodal puzzles and less structured LM reasoning templates~\citep{zelikman2024quietstar}.

\subsubsection{GATTINA - GenerAtion of TiTles for Italian News Articles} \posscite{DBLP:conf/clic-it/FrancisRGCINP24} challenge aims to assess the ability of LLMs to generate headlines for scientific news articles. Apart from testing the models' performance in a real-life use case, this task requires several foundational skills. First, the model must concisely summarize the content of an article which may be rather long. Second, the generated headline should be attractive and should encourage interaction from readers. Third, the model should adapt to the tone of the article itself, while adhering to the tone of the overall journal. 

GATTINA consists of 30,461 article-headline pairs, all of which are human-authored, from two prominent Italian media outlets, ‘Ansa scienza’ and ‘Galileo’. Due to the subjective and context-sensitive nature of headline evaluation, articles are evaluated along several dimensions. For this purpose, two classifiers are introduced: one judges whether a generated headline ‘belongs to’ a given article, and one assesses whether an article is machine- or human-authored. The former is trained to discriminate between coherent and incoherent headline-article pairs. The latter assesses how close to human writing the generated headline is (though machine-generated, being classified as human-produced is a positive outcome). We additionally use ROUGE \citep{lin-2004-rouge} and SBERT \citep{reimers-2019-sentencebert} to evaluate content appropriateness.

\begin{table}[!ht]
\footnotesize
\centering
\caption{
Results for the \textsc{GATTINA} task. Automatic evaluation scores reported for Natural-Synthetic (N-S) and Headline-Article (H-A) metrics across all evaluated models.
}
\label{tab:all-results}
\begin{tabular}{lrr}
\toprule
\textbf{Model} & \textbf{N--S} & \textbf{H--A} \\
\midrule
\textbf{\llama}      & 0.066 & 0.683 \\
\textbf{\minerva}    & 0.138 & 0.654 \\
\textbf{\anita}      & 0.201 & 0.615 \\
\textbf{\llamalarge} & 0.441 & 0.686 \\
\bottomrule
\end{tabular}
\end{table}

\paragraph{Performance across models} 
With the GATTINA challenge, which contains around 30,000 article-title pairs, LLMs are tested to generate appropriate headlines from news articles. 
For the same reason, evaluation is far from trivial. To quantify performance, we trained two classifiers: "N-S" (discriminating between a \underline{N}atural (human-written) and a \underline{S}ynthetic headline, and "H-A" (capturing the relevance of the \underline{H}eadline for the given \underline{A}rticle.) Table~\ref{tab:all-results} reports these metrics for the tested models.\footnote{Further details on the training of these metrics are provided in \citep{DBLP:conf/clic-it/FrancisRGCINP24}.}
While this is a true generative task that aligns with the training objective of language models better than many of the other CALAMITA challenges (where classification tasks are recast as generative), the task itself turned out to be far from simple for all of the tested models. This is partly due to the specifics of news headlines, which encapsulate human-intuitive characteristics, such as attractiveness, which are hard to grasp by machines \cite{cafagna2019suitable}, and partly  ascribable to the appropriateness of evaluation metrics.

\paragraph{Error Analysis}
Manual evaluation was performed by two professional journalists on a small sample of generated headlines. 
Overall, the LLMs show limited sensitivity to character count constraints (\textit{Registrato il sussurro dei piccoli delle megattere} [Ansa]=> ``\textit{Il sussurro dei piccoli, un segreto per sopravvivere: le megattere ’tacciono’ per non farsi sentire}" [\anita{}]) -- a strict requirement for many news outlets -- and tend to favour definitive phrasing, which leaves less room for ambiguity compared to headlines produced by editorial teams. Syntactically, headlines generated by LLMs often use the pattern ``key phrase: [text]", such as ``\textit{La Luna si restringe: terremoti e frane minacciano le future missioni umane}" (\llama). While this is theoretically a correct strategy, it is less favoured by editorial teams, who tend to prefer more fluent phrasing, integrating the keyword within the body of the headline (``\textit{La Luna si sta restringendo, ha perso 45 metri in milioni di anni}" (ANSA).) Models definitely overuse this strategy: 40\% of the cases in \llamalarge and \anita, and around 90\% in \minerva and \llama. Lastly, LLMs tend to include a greater number of details in the headlines, whereas editorial practices typically emphasise a single aspect of the news story which is expected to most attract the readers' attention.

In addition to stylistic issues, factual subtle aspects were also identified as problematic in generated titles. For example, the generated title ``\textit{L'allunaggio giapponese: un successo a metà}" (\minerva) suggests a half-success of the lunar mission, which is not true to the contents of the article, reflected of course in the original headline (``\textit{Luna difficile, arrivo con batticuore del lander Slim}" (ANSA)). Similarly, ``\textit{Il futuro è già qui, ma l’Italia rischia di restare indietro}" (\llamalarge) underscores a negative perspective which is the opposite of the article's contents and the original title (``\textit{Biotecnologie, in Italia confermata la crescita a 2 cifre}" (ANSA)). The experts also spotted cases where the generated title is dishonest and click-baiting: ``\textit{La mappa nascosta della pandemia: scoperta la vera causa della diffusione del coronavirus}" (\llama) (original: ``\textit{Pandemia, gli eventi di superdiffusione hanno un ruolo chiave}" (Galileo)): the use of ``\textit{map}'' is misleading and incorrect, and also the reference to the ``\textit{real}" cause of Coronavirus is a potentially dangerous statement, not present in the article.

In some cases the journalists assessed the generated headlines as better than the original ones, such as the title ``\textit{Terremoto in Turchia,
le onde sismiche arrivano in Antartide}" generated by Anita vs the human-produced (ANSA) title ``\textit{Sisma Turchia-Siria, rilevato da sismografi in Antartide}", which includes the repetition of "\textit{sisma}" (\textit{heartquake}) and "\textit{sismografi}" (\textit{seismographs}). Another headline evaluated as better than the original one is ``\textit{La tecnica Crispr per il taglia-incolla del Dna potrebbe aumentare il rischio di tumore}" (\llamalarge) vs the original (ANSA) ``\textit{Forse rischi di tumore dal taglia-incolla del Dna}". Lastly, it appears that models also introduce some degree of creativity, using figurative language and compact titles, such as in the headline ``\textit{Il viaggio delle lame, un filo che lega il passato}" (\anita) vs. the original from Galileo ``\textit{9000 anni fa l’ossidiana viaggiava in slitta verso l’Artico}". The expert evaluation here is uncertain: while the title seems good at first glance, especially because of the choice of ``\textit{lama}" (\textit{blade}) over "\textit{ossidiana}" (\textit{obsidian}), a word definitely more attractive for a title, the figurative expression ``\textit{un filo che lega il passato}" (\textit{a wire connecting the past}), while suggesting a good semantic connection with ``\textit{blades}", remains possibly too cryptic to serve as a safe, valid title.

Dealing with the long context of entire media outlets' articles, being coherent and, most of all, being able to forge a creative headline like a professional journalist still stands as a hard challenge even for the bigger models. In the context of CALAMITA, the title being in Italian instead of English possibly further increases the difficulty of this task. While none of the current models would be able to directly substitute a professional writer in yielding the most appropriate headline given an article, the GATTINA challenge shows that LLMs, especially larger ones, have great potential as part of a journalist's toolbox.

\subsubsection{GEESE - Generating and Evaluating Explanations for Semantic Entailment}
\posscite{DBLP:conf/clic-it/ZaninelloM24} challenge is focused on evaluating the impact of generated explanations on the predictive performance of language models for the task of Recognizing Textual Entailment in Italian. Using a dataset enriched with human-written explanations, different large language models are employed to generate and utilize explanations for semantic relationships between sentence pairs. GEESE assesses the quality of generated explanations by measuring changes in prediction accuracy when explanations are provided.
In GEESE, an LLM ($M_2$) is asked to predict the entailment relation (\textit{entailment}, \textit{neutral} or \textit{contradiction}) between pairs of sentences from the e-RTE-3-it dataset \cite{zaninello2023textual}. Explanations either written by humans (\texttt{gold}) or generated by another LLM ($M_1$), which may or may not be the same as $M_2$, are used as a ``hint'' by $M_2$. The results are compared with two baselines: $M_2$ using an empty string (\texttt{no-exp}) or $M_2$ using a copy of the first sentence (\texttt{dummy}) as hint. The task assesses $M_2$ models' ability to profit from explanations (by comparison with the baselines) but also, by comparing the results across different explanations on a given $M_2$, gives us an indirect measure of the effectiveness of such explanations, which can be taken as a proxy of their quality. Results for the four tested $M_2$  models (\llama{}, \llamalarge{}, \minerva{} and \anita{}) are reported in Table~\ref{tab:geiser-m1m2same}.

\begin{table*}[!ht]
\centering
\footnotesize
\caption{
Results for the \textsc{GEESE} task. Accuracy of $M_2$ models on the e-RTE-3-it dataset under different explanation settings: baselines (\textit{no-exp}, \textit{dummy}), human-written (\textit{gold}), and model-generated explanations. 
The best-performing $M_1$ model for each $M_2$ is shown in bold.
}
\label{tab:geiser-m1m2same}
\begin{tabular}{lcccc}
\toprule
 & \textbf{\llama} & \textbf{\llamalarge} & \textbf{\minerva} & \textbf{\anita} \\
 &  $M_2$ & $M_2$ & $M_2$ & $M_2$ \\
\midrule
\textbf{No-exp (b.l.)}     & 0.4950 & \textbf{0.5650} & 0.4875 & 0.4675 \\
\textbf{Dummy (b.l.)}      & 0.4937 & 0.6062 & 0.4850 & 0.5350 \\
\textbf{Gold}              & 0.6162 & 0.8150 & \textbf{0.5762} & 0.7062 \\
\textbf{\llama{} ($M_1$)}  & 0.6025 & 0.8637 & 0.5662 & 0.7850 \\
\textbf{\anita{} ($M_1$)}  & \textbf{0.6625} & \textbf{0.8850} & 0.5650 & \textbf{0.8287} \\
\bottomrule
\end{tabular}
\end{table*}

\paragraph{Performance across models} 
The biggest \llamalarge{} model achieves the highest result of 88.50\%. Models utilizing machine-generated explanations generally achieve highest accuracy, with only one model (\minerva) scoring highest by using human-written explanations (57.62\%). Explanations generated by \anita{} as $M_1$ are the most effective in enhancing the predictive ability of $M_2$'s, again with the exception of \minerva{} which, after the human-written ones, favours explanations generated by \llama{} by a small margin (56.62\%). In terms of accuracy gain, \anita{} as $M_2$ shows the highest increase of 36\% and 29\% over the \texttt{no-exp} and \texttt{dummy} baselines, respectively.

\paragraph{Error Analysis}
These results suggest that the quality and type of explanations influence predictive performance, highlighting the need for further research into optimizing explanation generation. 
For example, in the \textit{no-exp} setting, $M_2$-\llama{} incorrectly predicts ``entailment'' for the sentences with id=7, and using the explanations generated by itself ($M_1$-\llama) does change the label, but to ``contradiction'', which is also wrong. However, using the \texttt{gold} and \anita{} explanations leads to a correct ``neutral'' prediction: (\llama) \textit{La premessa e l'ipotesi non sono correlate in maniera stretta, poiché la premessa parla della partenza della signora Minton da un paese, mentre l'ipotesi afferma la sua nascita in quel paese. Non vi è un nesso logico tra i due fatti.} ({gold}) \textit{Il fatto che la signora Minton fosse in Australia prima di andare a Londra non necessariamente significa che sia nata in Australia.} ({\anita): \textit{La premessa afferma che la signora Minton lasciò l'Australia nel 1961, ma non dice nulla sulla sua nascita. L'ipotesi afferma che la signora Minton è nata in Australia, ma non dice nulla sul suo viaggio in Inghilterra. Queste due affermazioni non si riferiscono direttamente l'una all'altra, quindi non c'è alcun collegamento logico tra di esse.}}.
This is somehow in contrast with the expectation that a model should be better guided by explanations produced by itself. These findings encourage us to further explore the interplay between explanations and model performance, paving the way for more interpretable, transparent and user-friendly AI systems.

\subsubsection{GFG - Gender-Fair Generation}
\label{sec:gfg}

\posscite{DBLP:conf/clic-it/FrendaPSMRCFPNB24} challenge is designed to assess and monitor the recognition and generation of gender-fair language in both mono- and cross-lingual scenarios. It includes three tasks: (1) the detection of gender-marked expressions in Italian sentences, (2) the rewriting of gendered expressions into gender-fair alternatives, and (3) the generation of gender-fair language in automatic translation from English to Italian. The challenge relies on three different annotated datasets: the GFL-it corpus, which contains Italian texts extracted from administrative documents provided by the University of Brescia; GeNTE, a bilingual test set for gender-neutral rewriting and translation built upon a subset of the Europarl dataset; Neo-GATE, a bilingual test set designed to assess the use of non-binary neomorphemes in Italian for both fair formulation and translation tasks.

\begin{table}[!htb]
\centering
\footnotesize
\caption{
Results for the \textsc{GFG} task across all subtasks. Evaluation metrics include {BScF1} (BERTScore F1), {AccG} (Accuracy Gente), and {CWA} (Coverage-Weighted Accuracy).
}
\label{tab:gfg_results}
\begin{tabular}{lccccccc}
\toprule
 & \multicolumn{2}{c}{\textbf{Task 1}} 
 & \multicolumn{3}{c}{\textbf{Task 2}} 
 & \multicolumn{2}{c}{\textbf{Task 3}} \\
\cmidrule(lr){2-3}\cmidrule(lr){4-6}\cmidrule(lr){7-8}
\textbf{Dataset} 
& \textbf{NeoGate} & \textbf{GFL} 
& \textbf{GFL} & \textbf{NeoGate} & \textbf{Gente} 
& \textbf{NeoGate} & \textbf{Gente} \\
\textbf{Metric}  
& \textbf{BScF1} & \textbf{BScF1} 
& \textbf{AccG} & \textbf{CWA} & \textbf{AccG} 
& \textbf{CWA} & \textbf{AccG} \\
\midrule
\textbf{\anita}      
& 0.55 & 0.45 & 0.51 & 0.54 & 0.50 & 0.35 & 0.57 \\
\textbf{\llama}      
& 0.66 & 0.50 & 0.18 & 0.53 & 0.21 & 0.42 & 0.61 \\
\textbf{\llamalarge} 
& 0.59 & 0.53 & 0.61 & 0.73 & 0.54 & 0.58 & 0.56 \\
\textbf{\minerva}    
& 0.49 & 0.17 & 0.53 & 0.45 & 0.33 & 0.28 & 0.59 \\
\bottomrule
\end{tabular}
\end{table}

\paragraph{Performance across models}
Table~\ref{tab:gfg_results} reports 
the performance of all four models (\llama{}, \llamalarge{}, \minerva{} and \anita{}) across all sub-tasks. In general, we observe that the performance of models depends on the task, without an overall tendency. In some cases the smaller \llama{} even outperforms \llamalarge{}.

\paragraph{Error Analysis}
We conducted a preliminary qualitative analysis by manually reviewing examples of model outputs for each task. Such initial investigation revealed that the overall quality of the generated outputs was quite poor. Therefore, we decided to conduct a more formal analysis focusing specifically on the model 
showing good results across all the tasks: \llamalarge{}.
\ap{We analyzed 50 randomly selected outputs for each task and dataset combination.}
\mr{Each example was manually annotated taking into account several \ap{task-related} aspects, 
\ap{such as}
whether the generated text is a direct copy of the original,
the presence of
hallucinations or meaningless language, and if none, some or all of expressions referring to human entities are gender-neutral in the generated text. We further took note of what strategies the model used (e.g., periphrases employing `\textit{persona}'\textsubscript{[person]}), whether such strategy was part of the prompt and the model simply reproduced it, and whether the rewriting was less informative than the original text (e.g., if it turned ``\textit{ricercatori}''\ap{\textsubscript{[researchers]},}
into ``\textit{persone che lavorano}''
\ap{\textsubscript{[people who work]}}
.} 
In the gendered language detection task, the model correctly identified all target spans in 45\% of test sentences and none in 20\%. It was more prone to selecting unrelated spans in GFL-it than in Neo-GATE (38\% vs <3\%), possibly due to differences in textual domains: Neo-GATE features more generic language, while the administrative style of GFL-it makes gendered language harder to detect. Looking at the inclusive reformulation task, the manual analysis confirms the behavior suggested by the metrics: the model performs better in using \textit{innovative} obscuration strategies rather than the \textit{conservative} ones. Among the Neo-GATE outputs, 64\% replaced gender markers with neomorphemes, while only 6\% preserved them. In contrast, only 40\% of GFL-it and GeNTE outputs were fully gender-neutral, and 38\% remained fully gendered. However, neomorphemes were often misused: in 29\% of outputs, they appeared in unrelated or already neutral words (e.g., `\textit{{l\textschwa} opzione}'\textsubscript{[the option]}, pazient\textschwa\textsubscript{[patient]}), and in 28\%, they were used in required contexts but with incorrect word forms (e.g., `{\textit{del\textschwa}}' instead of `{\textit{dell\textschwa}}'\textsubscript{[of the]}). This suggests that, despite high CWA, the model struggles with precise neomorpheme generation. In using conservative strategies, the model consistently uses the term `\textit{persona}'\textsubscript{[person]} to rewrite generic masculine formulations. Moreover, in \ap{26.5\%} of the
outputs the model simply copied the original text without modifications. In 
\ap{21.43\%}
of cases information was lost \ap{in the rewritten sentence} and 
\ap{61\%}
of the re-writings were included in the prompt.
\ap{In} few cases the model used the double form by adding the female variant, \ap{a \textit{visibility} strategy \cite{rosola2023beyond} we consider incorrect within this challenge}.
\ap{Finally, in the {inclusive translation} task we observe similar trends, though with lower accuracy and less precise use of obscuration strategies, likely due to the added complexity of cross-lingual translation. Among the analyzed outputs, 46\% of Neo-GATE translations were fully gender-inclusive and 26\% fully gendered, while only 6\% of GeNTE outputs were inclusive and 82\% gendered. Compared to reformulation, neomorphemes were used less frequently and misused more often: 40\% appeared in unrelated or already neutral words. However, incorrect or unnecessary neomorpheme generation was less frequent, occurring in 10\% and 16\% of outputs, respectively. Notably, among the few inclusive GeNTE outputs, half used formulations with `\textit{persona}.'} 
As previous assessments of the topic \ap{suggest} \cite{savoldi2023test, piergentili-etal-2024-enhancing, piazzolla2023good}, even modern models that are the state-of-the-art in other tasks struggle with the \ap{evaluation and} generation of inclusive language.
Our analysis confirms that smaller models perform poorly on generative tasks, while larger models occasionally produce inclusive outputs but often employ 
the same rewriting strategy
(e.g., consistently adding `\textit{persona}').
The performance of the LLMs tested shows that there is still room for improvement in the various subtasks proposed and their evaluation.

\subsubsection{\textsc{GITA4Calamita} - Graded Italian Annotated Dataset}

\posscite{DBLP:conf/clic-it/PensaAEAG24} challenge investigates the reasoning abilities of LLMs in  physical commonsense reasoning and introduces a methodology to assess their understanding of the physical world through the design and creation of the Graded Italian Annotated dataset (GITA). GITA is written and annotated by a professional linguist and is composed of plausible and implausible stories. There are two mechanisms to generate implausible stories: cloze implausible stories are created by altering one sentence in a plausible story to make it implausible; order implausible stories are created by switching the order of two sentences in one plausible story to make it implausible. Our benchmark aims to evaluate three distinct levels of commonsense understanding with three specific tasks: identifying plausible and implausible stories within our dataset, identifying the conflict that generates an implausible story, and identifying the physical states that make a story implausible. Our findings reveal that, although the models may excel at high-level classification tasks, their reasoning is inconsistent and unverifiable, as they fail to capture intermediate evidence.

\paragraph{Performance across models}

The performance of the four evaluated models (\llama{}, \llamalarge{}, \minerva{} and \anita{}), shown in Table~\ref{tab:table8}), is assessed on three subtasks: story classification, conflict detection, and physical state recognition (corresponding metrics: accuracy, consistency, and verifiability). \llamalarge{} demonstrates the best overall performance, achieving 87.64\% accuracy in story classification and excelling in conflict detection and physical state recognition. \llama{} performs well, particularly on the cloze dataset, with 72.47\% accuracy and notable consistency in conflict detection. \anita{} outperforms \minerva, which struggles significantly, achieving only 38.20\% accuracy in story classification and failing in conflict detection and physical state recognition. Overall, the models perform better on the cloze dataset than on the order dataset.

\begin{table*}[!ht]
\centering
\footnotesize
\label{tab:table8}
\caption{
Results for the GITA task. The table reports model performance across the three evaluation dimensions-
\textit{story classification accuracy}, \textit{conflict consistency}, and \textit{physical-state verifiability}-
each computed for the overall dataset as well as for the cloze, order, and plausibility subsets.
}
\resizebox{\linewidth}{!}{
\begin{tabular}{l rrrr rrr rrr}
\toprule
\multirow{2}{*}{\textbf{Model}}
& \multicolumn{4}{c}{\textbf{Accuracy}} 
& \multicolumn{3}{c}{\textbf{Consistency}} 
& \multicolumn{3}{c}{\textbf{Verifiability}} \\
\cmidrule(lr){2-5}\cmidrule(lr){6-8}\cmidrule(lr){9-11}
 &  \textbf{Over.} & \textbf{Cloze} & \textbf{Order} & \textbf{Plaus.} & \textbf{Over.} & \textbf{Cloze} & \textbf{Ord.} & \textbf{Over.} & \textbf{Cloze} & \textbf{Ord.} \\
\midrule
\textbf{\minerva} & 38.20 & 13.67 & 13.11 & 88.88 & 1.67 & 3.41 & 0.00 & 0.00 & 0.00 & 0.00 \\
\textbf{\anita} & 58.98 & 61.53 & 38.52 & 77.77 & 18.41 & 29.91 & 7.37 & 7.94 & 14.52 & 1.63 \\
\textbf{\llama}  & 72.47 & 81.19 & 64.75 & 71.79 & 29.28 & 44.44 & 14.75 & 14.22 & 23.93 & 4.91 \\
\textbf{\llamalarge} & 87.64 & 95.72 & 88.52 & 78.63 & 65.27 & 74.35 & 55.73 & 36.40 & 47.00 & 25.40 \\
\bottomrule
\end{tabular}
}
\end{table*}

\paragraph{Error Analysis} 
Regarding the physical state recognition, models struggle to predict the verifiability of a story (deepest subtask). In particular, \minerva{} fails completely with an overall verifiability of 0.0. Oppositely, the Llama models reach the highest scores, with \llamalarge{} outperforming the rest. \llamalarge{} correctly predicts 86 physical states out of a total of 239 implausible stories, corresponding to 12 out of 14 physical state labels, including \textit{occupied}, \textit{solid}, \textit{wet}, and \textit{power}. It excels in detecting \textit{power} states (17 out of~24). In the following story, \llamalarge{} recognizes the impossibility of cooking broth on a turned-off stove: \textit{Marta ha acceso il gas. Marta ha messo sui fornelli una pentola. Marta ha versato il brodo nella pentola. Marta ha spento il gas. Marta ha cucinato il brodo per un'ora.
}
\llama{} detects a total of 34 correct physical states in implausible stories, with~7 predicted physical state labels out of~14. States related to \textit{functional}, \textit{temperature}, and \textit{in pieces} account for 28\%, 20\% and 29\%, respectively, of the total physical states in the entire implausible dataset. In the following example: \textit{Claudio accende la TV. La TV è rotta. Claudio trova il suo film preferito. Claudio guarda il film. Claudio spegne la TV e va a letto.}, \llama{} identifies the \textit{functional} physical state as the cause of conflict: if the television is not working, the actor of the story cannot watch a movie.
Although the LLaMA models achieve the highest scores for this task, there is a noticeable discrepancy between the physical states they predicted. The physical states that are accurately predicted by one LLaMA model are not recognized by the other.
We highlight that the cloze dataset outperforms the order dataset in all subtasks. Unlike the cloze set, where conflicts are created by substituting sentences, the order dataset generates conflicts by inverting the order of sentences without adding different words or explicit negations.
To conclude, we see that while some LLMs perform well in commonsense reasoning tasks at the end-task level, they lack a robust reasoning process when a deeper understanding of the physical world is required.

\subsubsection{INVALSI}

\posscite{DBLP:conf/clic-it/0002CE24} challenge is based on the Invalsi tests administered to students within the Italian school system. 
The Invalsi tests are country-wide assessments designed by expert pedagogists to monitor the average performance of students over the years, administered multiple times from primary school through high school.\footnote{\url{https://www.invalsi.it/invalsi/index.php}} The results of these tests have been used in several population studies \citep{invalsi1, invalsi2, invalsi3}, but they have seen little use as a benchmark for Large Language Models' understanding and reasoning in Italian. 
The Invalsi Task is composed of two subtasks, {Invalsi ITA} and {Invalsi MATE}, the first focused on language understanding and the second on math reasoning. The Invalsi ITA dataset used in this challenge contains 1117 questions of two types: (1) multiple choice: the student is asked to pick the correct answer among four candidate answers; (2) binary: the student is asked to assess a binary property of a statement, e.g. True -- False, Before -- After, etc.  The {Invalsi MATE} dataset is composed of 400 questions of three types:
(1) multiple choice: the student is asked to pick the correct answer among four candidate answers; (2) true/false: the student is asked to assess whether a given statement is True or False; (3) number: the student is asked the correct numerical answer to a question. 
In CALAMITA, all questions in both datasets are framed as multiple choice questions and we use a likelihood based approach to evaluate the models, we pick the answer with higher likelihood according to the model. For more details about the benchmark we refer to the work from \citet{DBLP:conf/clic-it/0002CE24}. 

\paragraph{Performance across models}

\begin{table}[!ht]
\centering
\footnotesize
\caption{
Results for the \textsc{INVALSI} task, reported separately for the Italian language understanding subtask ({ITA}) and the mathematics reasoning subtask ({MATE}). Scores represent model accuracy across all questions.
}
\label{tab:results_invalsi}
\begin{tabular}{lcccc}
\toprule
\textbf{Task} & \textbf{\llama} & \textbf{\llamalarge} & \textbf{\anita} & \textbf{\minerva} \\
\midrule
\textbf{ITA}   & 0.71	& 0.89	&  0.71	&   0.38 \\
\textbf{MATE}  & 0.51	& 0.72	&  0.47	&   0.34 \\
\bottomrule
\end{tabular}
\end{table}

We report the results for the four models (\llama{}, \llamalarge{}, \anita{} and \minerva{}) in Table~\ref{tab:results_invalsi}. 
We see that model size is more important than the training language, as \llamalarge{} is the only model performing significantly better than the others, including those with Italian tuning.

\paragraph{Error Analysis}
The performance gap between Invalsi ITA and Invalsi MATE is explained by knowing that Invalsi ITA questions often require to look for information in a longer text fragment while Invalsi MATE requires to perform a moderate reasoning step. It is challenging to identify questions types that are more challenging for LLMs. A trend, reported by \citet{puccetti-etal-2025-invalsi}, the Invalsi ITA questions show a clear pattern where difficulty increases as students age grows, while for Invalsi MATE this is only evident between first grade students and older ones. 

\subsubsection{ITA-SENSE - ITAlian word SENSE disambiguation}

\posscite{DBLP:conf/clic-it/BasileMS24} challenge assesses the abilities of LLMs in understanding lexical semantics through Word Sense Disambiguation (WSD). The classical WSD task is cast as a generative problem formalized as two tasks: [T1] Given a target word and a sentence in which the word occurs, generate the correct meaning definition; [T2] Given a target word and a sentence in which the word occurs, choose the correct meaning definition from a predefined set.
BabelNet \cite{navigli2010babelnet} is used as a sense inventory for retrieving the glosses assigned to each synset. Since some synsets have no Italian glosses, we obtained them by translating from English glosses. We provide two test sets: 1) \textit{Original} contains only synsets for which Italian glosses are available; 2) \textit{Translated}, which also contains translated glosses.
The four CALAMITA LLMs  (\llama{}, \llamalarge{}, \anita{} and \minerva{}) are tested in a zero-shot setting.

\paragraph{Performance across models}
We report the performance obtained for each type of task of \textsc{ITA-SENSE} \cite{DBLP:conf/clic-it/BasileMS24} in Table~\ref{tab:itasenseresults}. 
Overall, the impact of using translated glosses is more evident in the multiple choice task, with a significant drop in accuracy. We believe this is caused by the increased number of options (more options are available since there are more glosses).

\begin{table}[!ht]
\centering
\footnotesize
\caption{
Results for the \textsc{ITA-SENSE} task across both subtasks: [T1] generative and [T2] multiple-choice. 
{R-BERT} denotes the harmonic mean of Rouge-L and BERTScore for the generative task, while {Acc.} indicates accuracy for the multiple-choice task. 
Results are reported for test sets with and without translated glosses.
}
\label{tab:itasenseresults}
\begin{tabular}{lcccccccc}
\toprule
 & \multicolumn{2}{c}{\textbf{\llama}} & \multicolumn{2}{c}{\textbf{\anita}} & \multicolumn{2}{c}{\textbf{\minerva}} & \multicolumn{2}{c}{\textbf{\llamalarge}} \\
\cmidrule(lr){2-3}\cmidrule(lr){4-5}\cmidrule(lr){6-7}\cmidrule(lr){8-9}
\textbf{Test set} & \textbf{R-BERT} & \textbf{Acc.} & \textbf{R-BERT} & \textbf{Acc.} & \textbf{R-BERT} & \textbf{Acc.} & \textbf{R-BERT} & \textbf{Acc.} \\
\midrule
\textbf{Original}   & 0.319 & 0.413 & 0.259 & 0.512 & 0.257 & 0.215 & 0.314 & 0.629 \\
\textbf{Translated} & 0.319 & 0.386 & 0.263 & 0.475 & 0.255 & 0.200 & 0.310 & 0.579 \\
\bottomrule
\end{tabular}
\end{table}

\begin{figure*}[!ht]
    \centering

    \begin{minipage}[t]{0.49\textwidth}
        \centering
        \includegraphics[width=\linewidth]{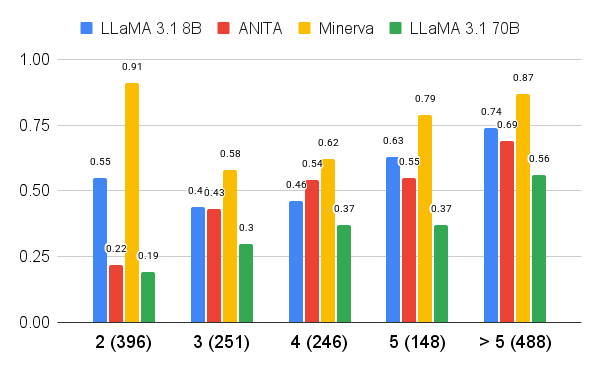}
        {\footnotesize \textbf{(a)} Without Translated Glosses}
    \end{minipage}
    \hfill
    \begin{minipage}[t]{0.49\textwidth}
        \centering
        \includegraphics[width=\linewidth]{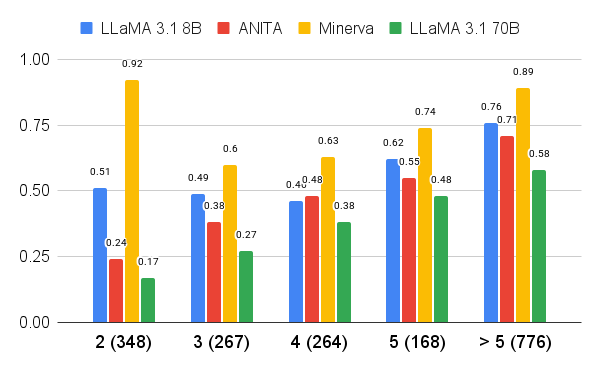}
        {\footnotesize \textbf{(b)} With Translated Glosses}
    \end{minipage}

    \vspace{0.75em}
    \caption{
        Error rates for the \textsc{ITA-SENSE} multiple-choice subtask across different levels of word polysemy. 
        Each bar corresponds to words with two, three, or more meanings; values in parentheses indicate the number of instances per group.
    }
\label{fig:error_analysis_ml}
\end{figure*}

\paragraph{Error Analysis}
We want to understand which words the models found more difficult to disambiguate. In particular, we want to understand whether words with more possible meanings have a more significant impact on the overall error rate. In Figure \ref{fig:error_analysis_ml}, we plot the error rates for the multiple choice task for four different levels of polysemy. In most cases, we find a higher error rate when there are more possible meanings for the target word. There are some exceptions to this pattern, notably \minerva{} and \llama{} without translated glosses. Finally, for the generative task, we note that due to the harmonic mean of Rouge-L and BERTScore, instances with no syntactic overlap were assigned a score of 0. However, this is not ideal, as the model can still generate a meaningful and correct definition without using any words from the ground truth. Considering only the BERTscore, we obtain a value of around 0.69 for all \llama{} models and 0.66 for \minerva{}.
To overcome these issues, we plan to change the metric for evaluating the quality of the generated gloss.


\subsubsection{MACID - Multimodal ACtion IDentification challenge}

\posscite{DBLP:conf/clic-it/RavelliVG24} challenge aims at evaluating LLMs' abilities to differentiate between closely related action concepts based on textual descriptions alone. The challenge is inspired by the "find the intruder" task, where models must identify an outlier among a set of 4 sentences that describe similar yet distinct actions. In each set, the sentences may contain the exact same verb or up to four different verbs, thus, the challenge comprises various levels of difficulty. The dataset is composed of 100 quadruples of sentences derived from the LSMDC Dataset (Large Scale Movie Description Challenge, \citet{DBLP:journals/ijcv/RohrbachTRTPLCS17}), though readapted to Italian after manual translation, and each caption is annotated with a concept from the IMAGACT ontology of actions \cite{DBLP:conf/lrec/MonegliaBFGKMP14}. Although mono-modal (text-only), the task is designed for future multimodal integration thanks to the alignment with videos from LSMDC, linking visual and textual representations to enhance action recognition.
The dataset focuses on “pushing” events, meaning all sentences refer to an action that can be described with a verb related to the semantic field of “push”. It presents action-predicate mismatches, where the same verb may describe different actions (e.g. “pressing a button” and “pressing the wood”) or different verbs may refer to the same action (e.g. “pressing a button” and “pushing a button”). Although currently unimodal (text-only), the dataset aligns each sentence with a short video from the LSMDC dataset \cite{DBLP:journals/ijcv/RohrbachTRTPLCS17}, thus being designed for future multimodal integration.

\paragraph{Performance across models} 
Overall, the accuracy among the four tested models ranges between 0.27 (\minerva) and 0.58 (\llamalarge), with intermediate values for \llama{} (0.43) and \anita{} (0.43). In general, the task is challenging for LLMs, with differences based on size (the larger, the better).

\begin{table}[!ht]
\centering
\footnotesize
\caption{
Results for the \textsc{MACID} task, reporting model accuracy across different verb distributions within each sentence quadruple. 
Categories indicate the number and arrangement of distinct verbs: 
4 = four different verbs; 
3 = two shared + two unique verbs; 
2(2+2) = two verbs equally distributed; 
2(3+1) shared = the intruder shares its verb with two others; 
2(3+1) unique = the intruder has a unique verb; 
1 = all sentences share the same verb. 
Scores represent average accuracy per group and overall.
}
\label{tab:verbs}
\begin{tabular}{lccccc}
\toprule
\textbf{Verbs dist.} & \textbf{Items} & \textbf{\llama} & \textbf{\llamalarge} & \textbf{\minerva} & \textbf{\anita} \\
\midrule
\textbf{4}              & 7   & 0.57 & 0.86 & 0.14 & 0.57 \\
\textbf{3}              & 16  & 0.44 & 0.50 & 0.25 & 0.44 \\
\textbf{2 (2+2)}        & 9   & \textbf{0.67} & 0.56 & 0.33 & 0.44 \\
\textbf{2 (3+1) shared} & 23  & 0.22 & 0.22 & 0.30 & 0.30 \\
\textbf{2 (3+1) unique} & 21  & \textbf{0.67} & \textbf{0.90} & \textbf{0.43} & \textbf{0.62} \\
\textbf{1}              & 24  & 0.29 & 0.63 & 0.13 & 0.33 \\
\midrule
\textbf{Overall} & 100 & 0.43 & 0.58 & 0.27 & 0.43 \\
\bottomrule
\end{tabular}
\end{table}

\paragraph{Error Analysis}
We investigate whether models rely more on the lexical component, i.e. verb lexemes, rather than on concept similarity. Indeed, accuracy is higher when the intruder action is lexically discriminated by the others, i.e. if the intruder contains a different verb from the rest of the sentences. Table~\ref{tab:verbs} shows that the models achieve the best results when all the sentences except the intruder share the same verb (fifth row), or when each sentence contains a different verb (first row), with the only exception of \minerva, whose results are, indeed, close to chance. Additionally, \llama{} performs well also when there are 2 verbs equally distributed (third row). Future work could extend the evaluation to
multimodal models to assess if and how the visual component contributes to action discrimination. Additionally, it would be interesting to compare LLMs with human performance.

\subsubsection{MAGNET - Machines Generating Translations}
\label{sec:magnet}

\posscite{DBLP:conf/clic-it/CettoloPPGNB24} challenge aims at testing the ability of LLMs in automatic translation, focusing on Italian and English (in both directions). The task proposes a benchmark composed of two datasets covering different domains and with varying distribution policies. Performances are reported in terms of four evaluation metrics, whose scores allow an overall evaluation of the quality of the automatically generated translations.

\begin{table*}[!ht]
\centering
\setlength{\tabcolsep}{5pt}
\footnotesize
\caption{
Results for the \textsc{MAGNET} translation task. 
Scores for both translation directions (en→it and it→en) are reported as the {average over the four evaluation subsets per direction} (\emph{public}, \emph{IT-private}, \emph{UK-private}, \emph{US-private}), 
ensuring full consistency with the detailed per-subset results in Appendix~\ref{tab:full-results}. 
Metrics include BLEU and ChrF (surface similarity), and COMET and BLEURT (semantic similarity). As a reference, we include the NLLB~3.3B model. Underlined values indicate cases where LLMs outperform NLLB.
}
\label{tab:scores}
\begin{tabular}{lcccccccc}
\toprule
 & \multicolumn{4}{c}{\textbf{en$\rightarrow$it}} & \multicolumn{4}{c}{\textbf{it$\rightarrow$en}} \\
\cmidrule(lr){2-5}\cmidrule(lr){6-9}
 & \textbf{BLEU} & \textbf{ChrF} & \textbf{COMET} & \textbf{BLEURT} & \textbf{BLEU} & \textbf{ChrF} & \textbf{COMET} & \textbf{BLEURT} \\
\midrule
\textbf{NLLB 3.3B}    & 43.49 & 68.05 & 88.15 & 80.12 & 45.83 & 68.51 & 86.57 & 76.83 \\
\midrule
\textbf{\llamalarge}  & 41.59 & 67.58 & \underline{88.95} & \underline{81.08} & \underline{47.05} & \underline{69.65} & \underline{87.92} & \underline{78.84} \\
\textbf{\llama}       & 34.45 & 62.35 & 87.43 & 78.33 & 42.55 & 66.33 & \underline{87.05} & \underline{77.39} \\
\textbf{\minerva}     & 36.59 & 63.37 & 87.71 & 79.11 & 40.86 & 65.34 & \underline{86.79} & \underline{76.84} \\
\textbf{\anita}  & 28.53 & 59.05 & 85.56 & 75.76 & 30.46 & 62.82 & 82.19 & 73.44 \\
\bottomrule
\end{tabular}
\end{table*}

\paragraph{Performance across models}

The results from the automatic metrics in Table~\ref{tab:scores} confirm the common assumption that larger models should perform better. We observe \llamalarge{} outperforming both its smaller counterpart, \llama{}, as well as \anita{} and \minerva{}. Nevertheless, all models perform well, especially according to the model-based metrics COMET and BLEURT, which are designed to assess the semantic similarity of the output to the source sentence (COMET only) and to a reference translation (both). 
The string-based metrics, i.e., BLEU \cite{papineni-etal-2002-bleu} and ChrF \cite{popovic-2015-chrf}, capture surface-level similarity with the reference, possibly punishing minor differences in similar outputs. Consider the examples in Table~\ref{tab:ex}: the outputs of \llamalarge{} and \llama{} only present small differences, yet their BLEU scores are quite different.

\begin{table*}[!ht]
\footnotesize
\centering
\caption{
Example outputs from the \textsc{MAGNET} translation task ({it$\rightarrow$en} direction), 
with sentence-level {BLEU} and {COMET} scores. 
Despite similar semantic adequacy, small lexical variations across models lead to large differences in surface-based metrics.
}
\setlength{\tabcolsep}{1.3pt}
\begin{tabularx}{\textwidth}{lccX@{}}
\toprule
\textbf{LLM} & \textbf{BLEU} & \textbf{COMET} & \textbf{Text} \\
\midrule
\textbf{Source}   & -     & -     & Lunedì un terremoto di lieve entità ha colpito il Montana occidentale alle 22:08. \\
\textbf{Reference} & -     & -     & A moderate earthquake shook western Montana at 10:08 p.m. on Monday. \\
\midrule
\textbf{\llamalarge} & 57.04 & 78.78 & On Monday, a minor earthquake struck western Montana at 10:08 p.m. \\
\textbf{\llama}      & 31.25 & 77.23 & On Monday a small earthquake struck western Montana at 10:08 PM. \\
\textbf{\minerva}    & 30.96 & 76.52 & Monday a mild earthquake struck western Montana at 10:08 pm. \\
\textbf{\anita} & 17.08 & 76.94 & A mild earthquake struck western Montana at 20:08 on Monday. \\
\bottomrule
\end{tabularx}
\label{tab:ex}
\end{table*}

\begin{table*}[!ht]
\footnotesize
\caption{
Example from the \textsc{MAGNET} translation task (\textbf{en$\rightarrow$it} direction), 
illustrating models’ handling of real-world knowledge and gender agreement. 
Correct gender forms are highlighted in green; incorrect or inconsistent ones in red. 
Only the largest model (\llamalarge{}) correctly produces the feminine form, 
showing stronger integration of factual and grammatical information.
}
\centering
    \setlength{\tabcolsep}{2pt}
\resizebox{0.95\linewidth}{!}{ 
\begin{tabularx}{\textwidth}{lX@{}}
\toprule
\textbf{LLM} & \textbf{Text} \\
\hline
Source & [...] present at the meeting was \fcolorbox{blue!30}{blue!30}{{the new Minister}}
of Labour [...] Nunzia Catalfo, [...]\\
\llamalarge & [...] era presente \fcolorbox{green!30}{green!30}{la nuova Ministra} del Lavoro [...] Nunzia Catalfo, [...]\\
\llama & [...] presente [...], è \fcolorbox{red!30}{red!30}{stato il nuovo Ministro} del Lavoro [...] Nunzia Catalfo, [...]\\
\minerva & [...] era presente \fcolorbox{red!30}{red!30}{il nuovo Ministro} del Lavoro [...] Nunzia Catalfo, [...] \\
\anita & [...] presente alla riunione era \fcolorbox{green!30}{green!30}{la nuova}\fcolorbox{red!30}{red!30}{Ministro} del Lavoro [...] Nunzia Catalfo, [...]\\
\bottomrule
\end{tabularx}
}
\label{tab:ex2}
\end{table*}

\paragraph{Error Analysis}

While automatic metrics allow for convenient system comparisons and ranking, they fail to capture specific linguistic or contextual phenomena.
For example, they are inaccurate in assessing errors involving gender reference \cite{zaranis2025watchingwatchers}, which become increasingly important at high levels of overall translation quality.
Consider the examples in Table~\ref{tab:ex2}: when translating a sentence referring to a real world entity, the Italian politician Nunzia Catalfo, two of the models (\llama{} and \minerva) incorrectly use masculine forms (``\textit{il nuovo Ministro}''), whereas \anita{} generates the mixed expression ``\textit{la nuova\textsubscript{[F]} Ministro\textsubscript{[M]}}'', which is grammatically incoherent.
Only the largest model, \llamalarge{}, produces the correct feminine form ``\textit{la nuova Ministra}'', demonstrating an ability to both integrate world knowledge into translation and to generate correct gendered references. Here again, we see the largest model standing out as a better model for translation.
Consistent with recent literature \cite{kocmi-etal-2024-findings}, our analysis confirms that LLMs are the state-of-the-art approach in MT. The superiority of the largest model in our experiments suggests a benefit from scale, aligning with evidence that larger LLMs outperform smaller counterparts across diverse translation tasks \cite{rei-etal-2024-tower}. As LLMs continue to improve, so too must our evaluation frameworks evolve toward methods capable of capturing both general translation performance and more nuanced linguistic and contextual behaviors. Future iterations of the MAGNET challenge should include more fine-grained analyses of translation phenomena, ranging from translation accuracy and semantic adequacy \cite{amrhein-etal-2022-aces}, to the integration of real-world knowledge \cite{li-etal-2025-leveraging}, and the handling of gender expression and other sociolinguistic aspects \cite{DBLP:conf/clic-it/FrendaPSMRCFPNB24}.

\subsubsection{Mult-IT}

 \posscite{DBLP:conf/clic-it/RinaldiGFGPN24} challenge provides a large-scale Multi-Choice Question Answering (MCQA) dataset, useful for evaluating the factual knowledge and reasoning abilities of LLMs in Italian. While some MCQA benchmarks for the Italian language already exist, they tend to be automatic translations of English data, which has several disadvantages: such questions may use unnatural-sounding constructions that do not align with Italian, may contain errors, and--more importantly--introduce topical and ideological biases reflecting Anglo-centric perspectives. The aim of this benchmark is to represent preferences, conventions, and ideas that are unique to Italian culture by providing questions originally written in Italian. Mult-IT consists of two core datasets, Mult-IT-A and Mult-IT-C. The former is a collection of 1,692 MCQs provided by Alpha~Test\footnote{\url{https://www.alphatest.it}.}, spanning 17 categories corresponding to topics featured in entry exams for Italian universities and public competitions. The latter is a collection of over 100,000 MCQs sourced from the publicly accessible online platform ``Concorsi Pubblici'', designed to prepare for job applications in the public sector. In both datasets, each question is paired with three to five possible answers. The performance of the four tested models (\llama{}, \llamalarge{}, \anita{} and \minerva{})  is evaluated using accuracy.

\begin{figure}[!ht]
    \centering
\includegraphics[width=.85\linewidth]{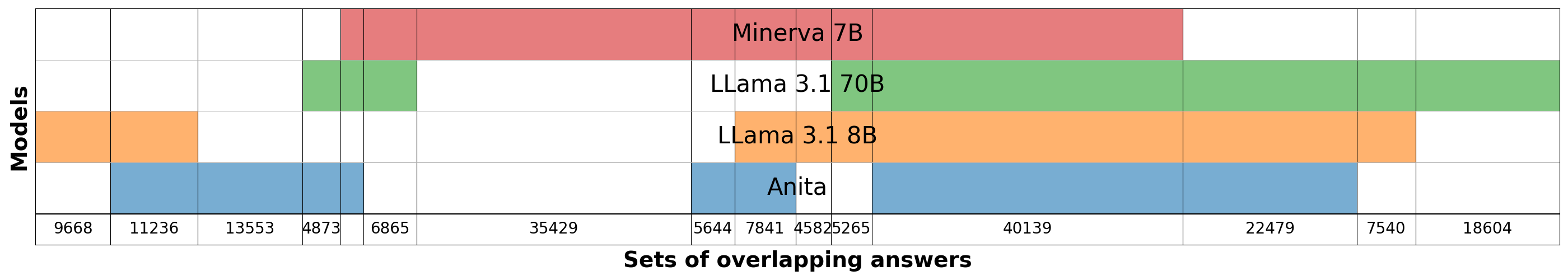}
\includegraphics[width=.9\linewidth]{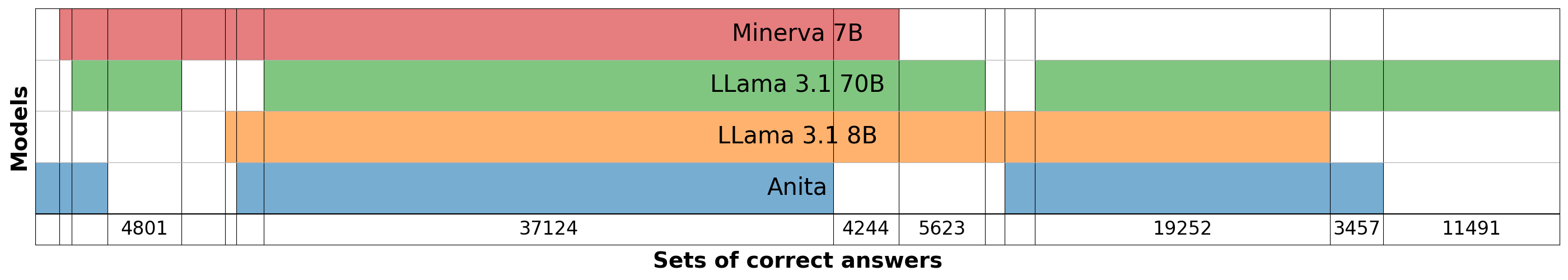}
\caption{
Answer overlap across models in the \textsc{Mult-IT} task. 
The top diagram shows the intersection of answers for all questions, 
while the bottom focuses on cases where models agreed on the correct option. 
Each area reflects the proportion of shared predictions among models.
}
\label{fig:overlap}
\end{figure}

\paragraph{Performance across models}
The large size and broad spectrum of Mult-IT, which contains over 100,000 multiple-choice questions on a wide variety of topics \citep{DBLP:conf/clic-it/RinaldiGFGPN24}, offer the opportunity to test LLMs  with respect to Italian society and culture as well as the models' reasoning abilities in Italian. Table~\ref{tab:accuracy} reports the accuracy of the tested models on the largest portion of Mult-IT, namely Mult-IT\_C (108,761). 
While the largest model (\llamalarge) obtaining the best accuracy isn't surprising, it is maybe less expected that the Italian-specific models underperform. For a more detailed analysis, we focus on (i) the agreement between models, to better understand how similar their behavior is, and (ii) the models' performance per (broad) categories, leveraging the variety of topics included in Mult-IT.
Figure~\ref{fig:overlap} shows the agreement between models in selecting a given answer. All models agree on 37\% of the questions, out of which 92\% are correct, leaving 3,015 questions where all models are in agreement giving exactly the same (wrong) answer. 
These are about one third of the 9,413 questions which could not be answered correctly by any model.  
Looking at the two Italian-specific models, they agree with one other and with no other model in 5644 cases, possibly in connection with more Italian-specific questions, but only 824 of these are correct.

\begin{table}[ht!]
\centering
\footnotesize
\caption{
Results for the \textsc{Mult-IT} task on the Mult-IT\_C dataset. 
Accuracy values indicate each model’s proportion of correctly answered multiple-choice questions.
}
\begin{tabular}{lr}
\toprule
\textbf{Model} & \textbf{Accuracy} \\
\midrule
\textbf{\llamalarge} & 0.812 \\
\textbf{\llama}      & 0.662 \\
\textbf{\anita}      & 0.628 \\
\textbf{\minerva}    & 0.502 \\
\bottomrule
\end{tabular}
\label{tab:accuracy}
\end{table}

\paragraph{Error Analysis}
Models struggle with logical and mathematical reasoning (44\% accuracy is well below the 65\% Mult-IT average), while general culture categories (including, e.g., geography and history) are the easiest at 75\%. This higher than average results in factual knowledge may depend on the more language independent nature of these questions and thus better representation in the training data.
Beyond the macro-level error analysis we offer in this short overview, Mult-IT lends itself well for micro-level analysis and eventually as a tool to not only better understand model behaviour but also to support model development.  
By cross-studying more in detail the models' agreement in the different categories at various degrees of granularity, it should be possible to identify the kind of fine-tuning datasets that are more useful to enhance the models' capabilities in Italian language and culture. 

\subsubsection{\textsc{PejorativITy}}

\posscite{DBLP:conf/clic-it/Muti24} challenge investigates misogyny expressed by LLMs through neutral words that can acquire a negative connotation when used as pejorative epithets. It is divided into two binary classification tasks. Task A focuses on the disambiguation of polysemous words in context, requiring the model to determine whether a target word conveys a pejorative meaning. Task B addresses misogyny detection at the sentence level, where the model must classify an input as misogynistic or not, either with or without access to the outcome of Task A. The goal is to assess whether prior disambiguation of potentially pejorative epithets improves performance in misogyny detection.

\paragraph{Performance across models}
Table \ref{tab:pejor} shows results for pejorative language detection for the four tested models (\llama{}, \llamalarge{}, \anita{} and \minerva{}). Notably, \llamalarge{} significantly outperforms all other models. This suggests a strong capacity for semantic understanding, where the implicit knowledge is exploited to understand the target (either women vs. other) in context. 
In contrast, smaller models perform substantially worse, even Italian-specific ones, indicating that model size plays a crucial role in handling the complexities of pejorative meaning disambiguation.
Table~\ref{tab:miso} shows results for misogyny detection in the standard and the \textit{pej} setting, where the model is informed of the output from Task A.
In the standard setting, model performance does not correlate with the number of parameters, as \llama{} achieves the best performance. \anita{} outperforms both \llama{} and \minerva{}, indicating that identifying misogyny is an easier task than identifying pejorative language for the fine-tuned native model. In the \textit{pej} setting, only \llamalarge{} benefits from the prediction of Task A, while the others show a considerable drop, indicating that the challenging task of disambiguation pejorativity significantly affects the prediction for misogyny.

\begin{table*}[!ht]
\centering
\footnotesize
\begin{minipage}{0.45\textwidth}
    \centering
    \caption{
    Results for the \textsc{PejorativITy} task (Task~A: pejorative language detection). 
    Scores are reported in terms of {Macro-F1}, evaluating the ability to disambiguate polysemous words used with a pejorative meaning.
    }
    \label{tab:pejor}
    \begin{tabular}{lc}
    \toprule
    \textbf{Model} & \textbf{Macro-F1} \\
    \midrule
    \textbf{\llama}      & 0.213 \\
    \textbf{\llamalarge} & \textbf{0.868} \\
    \textbf{\minerva}    & 0.179 \\
    \textbf{\anita}      & 0.031 \\
    \bottomrule
    \end{tabular}
\end{minipage}
\hspace{0.05\textwidth}
\begin{minipage}{0.45\textwidth}
    \centering
    \caption{
        Results for the \textsc{PejorativITy} task (Task~B: misogyny detection). 
        {M-F1 std.} indicates performance in the standard setting, 
        while {M-F1 pej.} refers to the setting informed by Task~A predictions. 
        Scores are {Macro-F1}.
    }
    \label{tab:miso}
    \begin{tabular}{lcc}
    \toprule
    \textbf{Model} & \textbf{M-F1 std.} & \textbf{M-F1 pej.} \\
    \midrule
    \textbf{\llama}      & \textbf{0.835} & 0.293 \\
    \textbf{\llamalarge} & 0.409          & \textbf{0.489} \\
    \textbf{\minerva}    & 0.370          & 0.012 \\
    \textbf{\anita}      & 0.448          & 0.291 \\
    \bottomrule
    \end{tabular}
\end{minipage}
\end{table*}

\paragraph{Error Analysis}

Models are facilitated in identifying both pejorativity and misogyny when targeted words co-occur with other slurs or negative adjective, as in ``\textit{Si nasconde dietro la foto di un quadro perché è una grassona orribile! Acida e frustrata e venduta ai russi}'' (En.: ``She hides behind a painting because she's a horrible fatty! Grumpy and frustated and sold to the Russians''). The best model for misogyny which considers pejorativity output, \llamalarge, underpredicts true positives in only a few cases (22 FN), but overpredicts misogyny very frequently (581 FP), even in very easy cases where the reference to non-women is evident: ``\textit{dolce fritto farcito con formaggio fresco e panna leggermente acida}'' (En.: ``fried dessert filled with fresh cheese and slightly sour cream'').

\subsubsection{PerseID - PERSpEctivist Irony Detection} \posscite{DBLP:conf/clic-it/BasileCFL24} challenge explores whether LLMs are able to  leverage the annotators' sociodemographic data to enhance performance in an irony detection task. Data is derived from MultiPICO \cite{DBLP:conf/acl/CasolaFLS0BBPRP24}, a recent multilingual dataset with disaggregated annotations and annotators' metadata. The data consists of post-reply exchanges collected from Twitter (X) and Reddit; crowdsourcers were required to annotate whether the reply is ironic in relation to the post.  The task has two primary goals: In Task~0, we evaluate the ability of models to detect irony in social media posts written in Italian. In Tasks 1, 2, and 3, we investigate whether incorporating annotator metadata improves model performance. Specifically, for each annotator, the model is prompted with their age/generation (Task 1), gender (Task 2), and both traits (Task 3). The goal is to assess whether LLMs adjust their predictions based on these additional data and whether these adjustments align with the annotators' labels for replies. The task is inspired by recent trends in socio-demographic and persona-based prompting.

\paragraph{Performance across models}
Table~\ref{tab:precision-recall-f1} shows the model performance across the four tested models and the tasks. 
\minerva{} is the only model that shows increased performance when providing demographic information about the annotators, showing an improvement both in terms of positive class and overall performance.
Looking at the individual tasks, \llama{} and \llamalarge{} show a bias toward the positive class regardless of the prompt. This results in the best F1-score for the positive class; however, the rate of false positives is high.  \anita{} has a slight bias toward the negative class, but shows the best results in terms of macro-f1 for most tasks. We also report a random baseline to better interpret the results.

\begin{table*}[!ht]
\centering
\footnotesize
\caption{
Results for the \textsc{PerseID} task (Tasks~0-3). 
Metrics are reported as precision (P), recall (R), and F1 for both classes and macro averages. 
Bold values indicate the best performance per task across models, while underlined scores mark improvements over the baseline (Task~0) obtained by adding annotator demographic traits. Table~\ref{tab:full-results} in the Appendix thus reports the F1 scores of the positive class for each task and model.
}
\label{tab:precision-recall-f1}
\begin{tabular}{lc ccc ccc ccc}
\toprule
\multirow{2}{*}{\textbf{Model}} & \multirow{2}{*}{\textbf{Task}} 
& \multicolumn{3}{c}{\textbf{Negative class}} 
& \multicolumn{3}{c}{\textbf{Positive class}} 
& \multicolumn{3}{c}{\textbf{Macro-average}} \\
\cmidrule(lr){3-5}\cmidrule(lr){6-8}\cmidrule(lr){9-11}
& & \textbf{P.} & \textbf{R.} & \textbf{F1} 
  & \textbf{P.} & \textbf{R.} & \textbf{F1} 
  & \textbf{P.} & \textbf{R.} & \textbf{F1} \\
\midrule
\textbf{Random} & -- & .690 & .504 & .582 & .316 & .504 & .389 & .503 & .504 & .485 \\
\midrule
\multirow{4}{*}{\textbf{\llama}} 
  & 0 & .918 & .079 & .145 & .328 & .985 & .492 & .623 & .985 & .492 \\
  & 1 & .909 & .107 & .191 & .333 & .977 & \underline{\textbf{.497}} & .621 & .542 & .344 \\
  & 2 & .922 & .105 & .188 & .333 & .981 & \underline{.498} & .628 & .543 & .343 \\
  & 3 & .919 & .103 & .186 & .333 & .980 & \underline{\textbf{.497}} & .626 & .542 & .341 \\
\midrule
\multirow{4}{*}{\textbf{\llamalarge}}
  & 0 & .933 & .101 & .183 & .333 & .984 & \textbf{.498} & .633 & .543 & .340 \\
  & 1 & .937 & .090 & .164 & .331 & .987 & .496 & .634 & .538 & .330 \\
  & 2 & .932 & .122 & .215 & .338 & .981 & \underline{\textbf{.502}} & .635 & .551 & \underline{.359} \\
  & 3 & .934 & .087 & .159 & .330 & .987 & .495 & .632 & .537 & .327 \\
\midrule
\multirow{4}{*}{\textbf{\anita}}
  & 0 & .712 & .791 & .749 & .395 & .300 & .341 & .554 & .545 & \textbf{.545} \\
  & 1 & .708 & .855 & .774 & .415 & .226 & .292 & .561 & .540 & \textbf{.533} \\
  & 2 & .711 & .859 & .778 & .433 & .237 & .306 & .572 & .548 & \textbf{.542} \\
  & 3 & .703 & .913 & .795 & .451 & .156 & .232 & .577 & .535 & .513 \\
\midrule
\multirow{4}{*}{\textbf{\minerva}}
  & 0 & .704 & .787 & .743 & .371 & .274 & .315 & .537 & .531 & .529 \\
  & 1 & .709 & .657 & .682 & .353 & .410 & \underline{.380} & .531 & .534 & \underline{.531} \\
  & 2 & .706 & .718 & .712 & .357 & .344 & \underline{.350} & .531 & .531 & \underline{.531} \\
  & 3 & .708 & .628 & .665 & .347 & .433 & \underline{.385} & .527 & .530 & \textbf{.525} \\
\bottomrule
\end{tabular}
\end{table*}

\begin{table}[!ht]
\centering
\footnotesize
\caption{
Agreement analysis for the \textsc{PerseID} task. 
The table shows how often model predictions in demographic-informed settings (Tasks~1-3) matched those of the baseline without demographics (Task~0), 
indicating the degree to which annotator information affected the outputs.
}
\label{tab:agree_task0}
\begin{tabular}{lcccc}
\toprule
\textbf{Task} & \textbf{\anita} & \textbf{\llamalarge} & \textbf{\llama} & \textbf{\minerva} \\
\midrule
\textbf{Task 1} & 90.98\% & 98.35\% & 96.99\% & 86.51\% \\
\textbf{Task 2} & 92.27\% & 98.04\% & 97.20\% & 92.13\% \\
\textbf{Task 3} & 86.85\% & 98.33\% & 95.99\% & 84.05\% \\
\bottomrule
\end{tabular}
\end{table}

\paragraph{Error Analysis}
Models tend to generate the same label for each post-reply pair regardless of the demographic information about the annotator provided in the prompt (in 95\%, 97\%, 85\%, 82\% for \llama{}, \llamalarge, \anita{} and \minerva{} respectively). 
To better understand the impact of providing demographics in the prompt, we computed the percentage of how often the labels of Tasks 1, 2 and 3 were equal to the baseline (Task 0). Table \ref{tab:agree_task0} reports the results, additionally showing a lower variation in \llama{} and \llamalarge{} across all tasks compared with the other models. Moreover, providing information about annotators' gender (Task 2) results in having the lowest impact for all models, except for \llamalarge{}. 
To understand whether different models tend to have similar outputs for the same instances, we also computed Cohen's kappa \cite{doi:10.1177/001316446002000104} to measure models agreement on task 0. The two LLaMA models prove a moderate agreement ($k=0.49$) while \anita{} and \minerva{} show a slight agreement ($k=0.10$). Strong disagreement is showed especially between \minerva{}/\anita and the two LLaMA models (around 0.02 and 0.04). 
While demographic and persona-based prompting has shown some success in other scenarios \cite{DBLP:conf/icml/SanturkarDLLLH23, 10.1093/pnasnexus/pgae346, cao-etal-2023-assessing}, the examined models do not seem to consistently modify their predictions when demographic information is provided. Only Minerva seems slightly sensitive to this information, reporting a better performance and a limited variation in the distribution of labels generated for task 1 and 2. 
Future work should investigate whether this behavior holds in a few-shot scenario, 
and explore evaluation metrics that might be more suitable for a multi-perspective environment 
\cite{rizzi-etal-2024-soft} and annotator-based evaluation \cite{mokhberian-etal-2024-capturing}.

\subsubsection{TERMITE - Italian Text-to-SQL Benchmark}
\posscite{DBLP:conf/clic-it/RanaldiROZR24} challenge focuses on the Text-to-SQL task in Italian. Natural language queries are written natively in Italian, and the models are expected to turn them into SQL queries. The dataset is built to be invisible to search engines since it is locked under an encryption key delivered along the resource to reduce accidental inclusion in upcoming training sets. It contains hand-crafted databases in ten different domains, each with a balanced set of NL-SQL query pairs. The NL questions are built in such a way that they can be solved by a model relying only on its linguistic proficiency and an analysis of the schema, with no external knowledge needed.

\paragraph{Performance Across Databases}
Out of the four models tested within CALAMITA, \llamalarge{} achieved the highest accuracy on average (45.6\%), with strong results on \texttt{pratica} (72.7\%) and \texttt{coronavirus} (65.0\%) databases, but struggled with \texttt{bowling} and \texttt{galleria}. \llama{} (36.1\%) and \anita{} (37.6\%) show lower performance, with higher variability across domains. 
\minerva{} significantly underperformed (4.0\%), highlighting difficulties in handling structured queries in Italian.

\begin{table}[!ht]
\centering
\footnotesize
\caption{Results for the \textsc{TERMITE} task. Execution accuracy (\%) per database (DB).}
\label{tab:termite}
\begin{tabular}{lcccc}
\toprule
\textbf{DB} & \textbf{\llamalarge} & \textbf{\llama} & \textbf{\anita} & \textbf{\minerva} \\
\midrule
bowling     & 25.0 & 29.2 & 16.7 & 0.0 \\
centri      & 42.1 & 31.6 & 36.8 & 15.8 \\
coronavirus & 65.0 & 40.0 & 35.0 & 5.0 \\
farma       & 35.0 & 40.0 & 55.0 & 5.0 \\
farmacia    & 45.0 & 15.0 & 40.0 & 5.0 \\
galleria    & 26.1 & 39.1 & 43.5 & 0.0 \\
hackathon   & 36.8 & 42.1 & 31.6 & 0.0 \\
pratica     & 72.7 & 63.6 & 50.0 & 0.0 \\
recensioni  & 50.0 & 22.2 & 11.1 & 0.0 \\
voli        & 64.7 & 35.3 & 58.8 & 11.8 \\
\midrule
\textbf{Total} & \textbf{45.6} & \textbf{36.1} & \textbf{37.6} & \textbf{4.0} \\
\bottomrule
\end{tabular}
\end{table}

\begin{figure}[!ht]
    \centering

    \begin{minipage}[t]{0.45\textwidth}
        \centering
        \includegraphics[height=4cm]{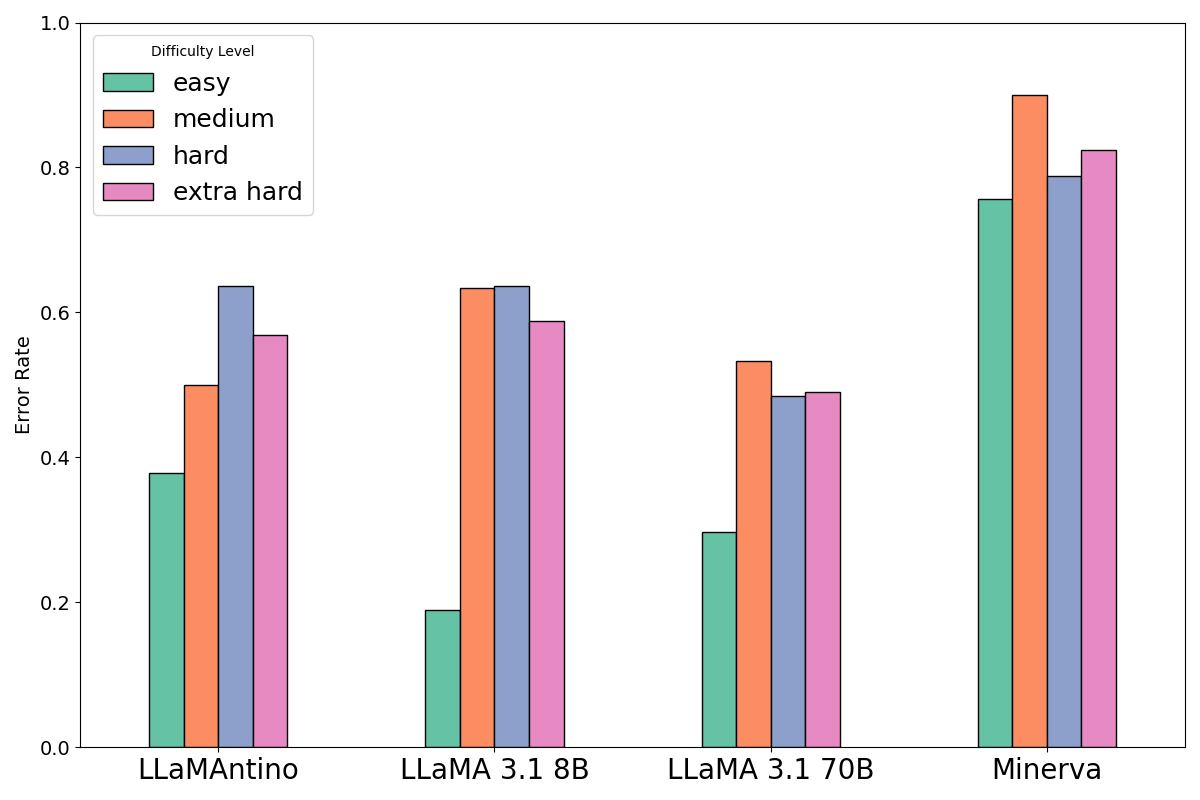}
        \vspace{0.5em}
        
        {\footnotesize (a) Error rate by hardness level.}
        \label{fig:errors_hardness_plot}
    \end{minipage}
    \hfill
    \begin{minipage}[t]{0.45\textwidth}
        \centering
        \includegraphics[height=4cm]{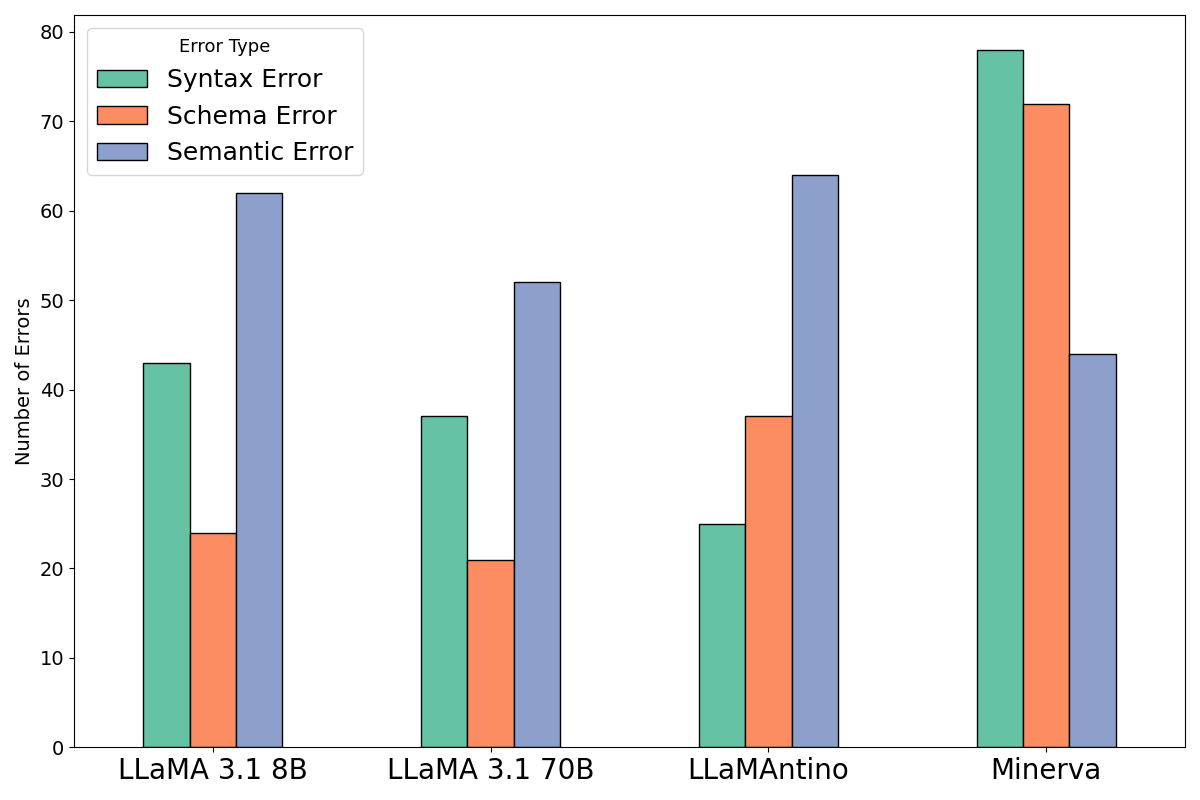}
        \vspace{0.5em}
        
        {\footnotesize (b) Distribution of error types.}
        \label{fig:error-types_plot}
    \end{minipage}

    \vspace{0.75em}
    \caption{Results for the \textsc{TERMITE} task (error analysis). 
Panel (a) shows error rate by query hardness level; panel (b) reports the distribution of error types.}
    \label{fig:combined_errors}
\end{figure}

\paragraph{Error Analysis} 
Queries can be categorized by \textit{hardness level}, following the criteria proposed in the Spider benchmark~\cite{yu-etal-2018-spider}, which account for nesting, aggregations, and the number of tables joined.
As illustrated in Figure~\ref{fig:combined_errors}(a), error rates generally increase with query complexity. \llama{}, \llamalarge, and \anita{} show increasing error rates as query complexity grows, especially on \textit{hard} queries, while performing significantly better on \textit{easy} ones.
\minerva{} errors rates appear to be higher and comparable across hardness levels.
We consider three categories of Text-To-SQL errors: \textit{syntax errors} (queries with invalid SQL syntax), \textit{schema errors} (queries that misuse the database schema, e.g., non-existent columns or tables), and \textit{semantic errors} (queries that execute but return incorrect results). 
In Figure~\ref{fig:combined_errors}(b) the number of errors for each of these three categories is reported.
\llama{}, \llamalarge{}, and \anita{} mainly fail at the semantic level, producing  mostly valid SQL that yields incorrect results. As model size increases, LLaMA models reduce syntax errors, with \anita{} being the most precise in syntax. \minerva{}, in contrast, struggles across all categories, especially syntax.
These findings reinforce that one of the most challenging aspects in Text-to-SQL is generating queries that are not only syntactically correct but also semantically accurate.
TERMITE highlights the importance of evaluating Text-to-SQL in non-English languages to assess how models handle formal language generation. Expanding it to multilingual settings, arithmetic reasoning, and specialized domains like healthcare would provide a more comprehensive test of model generalization.

\subsubsection{\textsc{TRACE-it} - Testing Relative clAuses Comprehension through Entailment in ITalian} 

\posscite{DBLP:conf/clic-it/Brunato24} challenge focuses on a specific linguistic ability of LLMs, namely the comprehension of a complex syntactic construction in Italian: object relative clauses (ORCs). The proposed benchmark comprises 566 sentence pairs. Each pair includes a complex sentence containing the target ORC and a simpler declarative sentence whose meaning may or may not be logically entailed by the first, depending on the syntactic relationship between the noun phrases in the ORC. The task is framed as binary entailment: given the complex sentence, the model must decide whether it entails the simpler one. Most ORCs are drawn from linguistic and psycholinguistic literature, enabling the analysis of factors known to affect human comprehension, such as grammatical and semantic features of the dependent elements and the filler-gap distance. A smaller subset includes ORCs from existing NLP benchmarks designed to test models' sensitivity to sentence acceptability. \textsc{TRACE-it} contributes to a growing body of syntactic evaluation resources for neural language models, which often rely on minimal pairs of grammatical and ungrammatical sentences targeting specific linguistic phenomena. While similar in purpose, \textsc{TRACE-it} introduces a novel evaluation perspective: instead of focusing solely on grammaticality judgments or complexity scoring, it frames the task as an entailment problem-requiring models to infer logical meaning based on syntactic structure.

\paragraph{Performance across models}

Table~\ref{tab:performance} reports the performance of the four tested models (\llama{}, \llamalarge{}, \anita{} and \minerva{}) on the \textsc{TRACE-it} dataset: as it can be seen, \llamalarge{} achieves the highest accuracy (85.69\%) and F1-score (86.25\%), outperforming all other models. \minerva, despite being a native Italian model, performs the worst, with 62.72\% accuracy and the lowest F1-score (69.11\%). \llama{} and \anita{} exhibit comparable performance but remain significantly behind \llama{}. In terms of Precision-Recall Trade-off, \llama{} demonstrates high precision but lower recall, indicating a conservative approach to positive classifications. \minerva{} shows the opposite pattern, with high recall but low precision.

\begin{table}[!ht]
\centering
\footnotesize
\caption{
Results for the \textsc{TRACE-it} task. 
Zero-shot performance of all models on the dataset. 
The highest accuracy is shown in {bold}.
}
\label{tab:performance}
\begin{tabular}{lcccc}
\toprule
\textbf{Model} & \textbf{Accuracy} & \textbf{Precision} & \textbf{Recall} & \textbf{F1} \\
\midrule
\textbf{\llamalarge} & \textbf{0.85} & 0.83 & 0.90 & 0.86 \\
\textbf{\llama}  & 0.73 & 0.84 & 0.56 & 0.67 \\
\textbf{\anita}            & 0.70 & 0.76 & 0.60 & 0.67 \\
\textbf{\minerva}          & 0.63 & 0.59 & 0.83 & 0.69 \\
\midrule
\textbf{\textit{Random}}     & 0.50 & --   & --   & --   \\
\bottomrule
\end{tabular}
\end{table}

\begin{table}[!ht]
\centering
\footnotesize
\caption{Models accuracy per condition of ORCs in the \textsc{TRACE-it} dataset for each model.
The ALL column reports the average score across models for each condition.
Results are ranked by average accuracy.}
\label{tab:performance_across_conditions}
\resizebox{\linewidth}{!}{
\begin{tabular}{lccccc}
\toprule
\textbf{RO Condition} & \textbf{\llamalarge} & \textbf{\llama} & \textbf{\anita} & \textbf{\minerva} & \textbf{ALL} \\
\midrule
anim-mismatch-an-in        & \textbf{0.893} & 0.714 & 0.821 & 0.813 & 0.810 \\
anim-mismatch-an-in-dist   & \textbf{0.857} & 0.786 & 0.750 & 0.821 & 0.804 \\
anim-mismatch-in-an        & 0.857 & \textbf{0.964} & 0.786 & 0.588 & 0.799 \\
mixed-sister-challenge     & \textbf{1.000} & 0.771 & 0.771 & 0.576 & 0.779 \\
anim-mismatch-in-an-dist   & \textbf{0.786} & 0.750 & 0.750 & 0.750 & 0.759 \\
gen-num-match-dist         & \textbf{0.853} & 0.735 & 0.706 & 0.569 & 0.716 \\
gen-mismatch               & \textbf{0.824} & 0.716 & 0.706 & 0.607 & 0.713 \\
gen-num-match              & \textbf{0.843} & 0.725 & 0.696 & 0.588 & 0.713 \\
gen-mismatch-dist          & \textbf{0.818} & 0.697 & 0.606 & 0.706 & 0.707 \\
num-mismatch               & \textbf{0.873} & 0.657 & 0.657 & 0.545 & 0.683 \\
num-mismatch-dist          & \textbf{0.818} & 0.667 & 0.606 & 0.571 & 0.666 \\
\bottomrule
\end{tabular}
}
\end{table}

\paragraph{Error Analysis}

Table~\ref{tab:performance_across_conditions} presents a fine-grained analysis of model performance across different object relative clause (ORC) conditions found in the first sentence of each entailment pair. A consistent pattern emerges: models achieve overall highest accuracy in conditions involving an animacy mismatch, with slightly better performance when the first NP is animate and the embedded NP is inanimate. This contrasts with patterns typically observed in human sentence processing, where such a configuration tends to increase ambiguity in ORCs comprehension by favoring a subject interpretation of the first NP.
Interestingly, increasing the syntactic distance between the two NPs does not significantly disrupt this facilitation effect, indicating that the models are relatively robust to structural complexity in these cases. 
In contrast, grammatical mismatches-such as those involving gender and number-exhibit a more limited and inconsistent impact. Unlike what is observed in human sentence processing \cite{Biondo2023}, these features do not appear to systematically aid the models in disambiguating syntactic roles. This suggests that, while humans may rely on agreement cues to reinforce syntactic structure, LLMs do not prioritize such features in the same way, instead relying more heavily on semantic distinctions like animacy. Additionally, accuracy is notably higher for sentence pairs derived from 'sister challenge' benchmarks, particularly for the largest model (\llamalarge). This may reflect a familiarity effect, as these examples are sourced from datasets designed around grammaticality judgments rather than entailment, potentially resembling patterns the models encountered during pretraining.
Overall, these findings suggest that models rely heavily on semantic cues, such as animacy distinctions, while demonstrating less sensitivity to purely morpho-syntactic mismatches. The varying performance across models further emphasizes the role of scale and training data composition, with larger models demonstrating greater robustness in handling complex syntactic structures. 
Beyond assessing model accuracy on sentences exhibiting complex syntactic structures, \textsc{TRACE-it} was designed as a language-specific benchmark that also offers a valuable opportunity to investigate how LLMs internally encode minimal pairs differing in specific morpho-syntactic features. Future research could explore whether models represent such distinctions in a structured manner and how fine-tuning on these constructions might reshape their internal representations. This could provide deeper insights into the extent to which LLMs acquire abstract syntactic principles rather than relying on surface-level heuristics.

\subsubsection{VeryfIT}

 \posscite{DBLP:conf/clic-it/GiliPPC24} challenge tests the in-memory factual knowledge of LLMs on texts written by professional fact-checkers, posing it as a true or false question. The topics of the statements vary, but most are in specific domains related to the Italian government, policies, and social issues. The task presents several challenges: extracting statements from segments of speeches, determining appropriate contextual relevance both temporally and factually, and verifying the statements' accuracy.

\begin{table}[!th]
\centering
\footnotesize
\caption{
Results for the \textsc{VeryfIT} task. 
Accuracy and weighted accuracy (WA) are reported for each dataset configuration: 
{Full}, {Small}, and {Enriched (Enr.)}. 
Weighted accuracy accounts for label imbalance. 
Asterisks (*) mark scores affected by strong prediction skew.
}
\label{tab:results}
\begin{tabular}{lrrrrrr}
\toprule
\textbf{Model} & \multicolumn{3}{c}{\textbf{Accuracy}} & \multicolumn{3}{c}{\textbf{Weig. Accuracy}} \\
\cmidrule(lr){2-4} \cmidrule(lr){5-7}
& \textbf{Full} & \textbf{Small} & \textbf{Enr.} & \textbf{Full} & \textbf{Small} & \textbf{Enr.} \\
\midrule
\textbf{\llama}      & \textbf{0.594} & \textbf{0.567} & \textbf{0.578} & 0.463 & 0.472 & 0.484 \\
\textbf{\llamalarge} & 0.593 & 0.556 & 0.556 & 0.491 & 0.483 & 0.473 \\
\textbf{\anita}      & 0.408 & 0.433 & 0.433 & \textbf{0.590} & \textbf{0.565} & \textbf{0.565} \\
\textbf{\minerva}    & 0.592 & \textbf{0.567} & 0.567 & 0.409 & 0.433 & 0.433 \\
\bottomrule
\end{tabular}
\end{table}

\paragraph{Performance across models}
Overall results (see Table~\ref{tab:results}) indicate that LLMs exhibit notable difficulties with factuality. We observe a tendency of all LLMs to overpredict one label over the other. To offer a more realistic evaluation, we report both Accuracy and Weighted Accuracy (WA). The two Italian models overpredict in opposite directions, with \anita{} labelling 2,013 of 2,021 statements as \texttt{False}, and \minerva{} 2,020 of 2,021 as \texttt{True}. For this reason we have excluded them from further analysis. Overprediction of the \texttt{True} label is also observed for  \llama{} and \llamalarge{} (86\% for \texttt{8B} on both Full and Small datasets; 79\%/77\% for \texttt{70B} on Full/Small).
Both \llama{} and \llamalarge{} register WAs below 0.50 and raw accuracies around 0.60.
The skewed prediction distribution further questions the LLMs' suitability for identifying false claims. 
The following analyses were conducted in the context of the \textsc{VeryfIT!} dataset \cite{DBLP:conf/clic-it/GiliPPC24,DBLP:conf/clic-it/GiliPC23} to evaluate model behavior across potential confounding factors. 
 Weighted Accuracy for both models ranges between 0.45 and 0.50 across political orientations. The center-right subset deviates slightly, but limited sample sizes impede the detection of systematic bias.
Model accuracy remains stable diachronically,
including for post-training claims. This persistence likely results from many claims referencing long-standing or widely known facts. Performance is consistent across claim contexts, with a slight improvement on globally framed statements likely reflecting the presence of some topics in pre-training data in languages other than Italian.
Across topics, model performance is broadly comparable. \llamalarge{} slightly outperforms in general, except in the \emph{Environment} category where \llama{} exceeds by 0.19. Variance across topics is modest, with \emph{Social Issues} and \emph{Foreign Affairs} showing highest accuracy, likely due to their international relevance.

\paragraph{Error Analysis}
We analyze the impact of numeric elements and named entities (NEs) on model performance. Both models show a negative correlation between the presence of numeric tokens and WA, with \llamalarge{} performing best when no numeric token is present in the claims.
Increased NE count correlates with higher accuracy but not with WA, in line with the overprediction tendency.
\llamalarge{} generalizes better with higher NE counts, while \llama{} underperforms, particularly with no or many NEs. Both models prefer samples with exactly one person (PER) entity. \llamalarge{} responds to organization (ORG) and location (LOC) entities steadily, reaching 0.59 WA with 3-4 LOCs, while \llama{} degrades when prompted with LOC-heavy inputs.
No definitive trade-off emerges between the benefits of added context and the risks of increased complexity. While more context may aid claim interpretation, it can also introduce confounding or misleading details. \llamalarge{} generally exhibits better performance -- confirming the relationship between models' size and parametric knowledge. The lower performance of the Italian models seems mostly due to limitations in the pretraining corpora.
Future work should explore whether these limitations are due to missing knowledge or insufficient contextual understanding. Incorporating partial source documents, particularly introductory paragraphs, may help isolate these variables and provide a more realistic basis for automated fact-checking systems.


\subsubsection{\textsc{\textsc{ItaEval}}}
\label{sec:itaeval}
\posscite{DBLP:conf/clic-it/AttanasioDQSS24} challenge is an evaluation suite pre-dating CALAMITA and comprising three overarching task categories: (i) natural language understanding, (ii) commonsense and factual knowledge, and (iii) bias, fairness, and safety. It includes nineteen tasks that combine existing and newly introduced datasets. The suite was developed to provide a standardized and methodologically consistent framework for evaluating Italian language models, enabling rigorous and comparative assessments of model performance. Thanks to its use of the \textsc{lm-eval-harness}, \textsc{ItaEval} also influences the evaluation infrastructure adopted within CALAMITA.
A key aspect of \textsc{ItaEval} is that it contains two different types of datasets.  
On one hand, several tasks are \emph{entirely developed in Italian}, many of them rooted in the EVALITA tradition, including AMI (misogyny identification) \cite{FersiniNR20-ami2020}, HaSpeeDe (hate speech) \cite{SanguinettiCNFS20-haspeede20}, HateCheck-ITA \cite{rottger-etal-2022-multilingual-hatecheck}, GENTe rephrasing \cite{piergentili-etal-2023-hi}, IronITA (irony and sarcasm) \cite{CignarellaFBBPR18-ironita18}, \textsc{ITA-CoLA} (acceptability) \cite{trotta-etal-2021-monolingual-cross-itacola}, SENTIPOLC (sentiment polarity classification) \cite{BarbieriBCNNP16-sentipolc16}, and the \textit{Fanpage}/\textit{Il~Post} summarization datasets \cite{info13050228-summarization}.  
On the other hand, six tasks originate from \emph{automatic or manual translations} of datasets originally compiled in English, namely ARC-Challenge-it \cite{clark2018think}, BeLeBeLe-it \cite{bandarkar2023belebele}, HellaSwag-it \cite{zellers2019hellaswag}, SQuAD-it \cite{squad18}, TruthfulQA-it \cite{lin2021truthfulqa}, and XCOPA-it \cite{ponti2020xcopa}.
Because CALAMITA restricts its benchmark to resources native to Italian, only the Italian-original tasks are included in the official analysis. These tasks form the \textsc{ItaEval} component of the CALAMITA benchmark: they are analyzed in this section, used in aggregated comparisons (Section~\ref{sec:results-overview-agg}), and reported in the main summary table in the Appendix (Table~\ref{tab:full-results}).

For completeness and transparency, however, we adopt a dual strategy:  
(i) the Italian-native tasks are fully integrated within CALAMITA and used in all evaluations and aggregate analysis;  (ii) the translated-from-English tasks are evaluated within the CALAMITA framework but \emph{not} included in aggregate analyses nor in the main results table. Their scores are reported separately in Table~\ref{tab:itaeval-english-results} in the Appendix, and some of them are included in the \textsc{ItaEval} discussion in this paragraph. This separation preserves methodological consistency with CALAMITA’s Italian-first philosophy, while still making the full \textsc{ItaEval} evaluation available to the reader.

\paragraph{Performance across models}

Table~\ref{tab:model_comparison} reports the average scores aggregated by the three \textsc{ItaEval} macro categories. 
Expectedly, \llamalarge{} outperforms all smaller models across the board. Among them, the Italian-specific fine-tuning of \anita{} proves beneficial only in the CFK category.
Though natively Italian, \model{minerva} performs the weakest across all categories, especially in CFK, highlighting strong limitations in tasks that require accessing commonsense knowledge and factual information.

\begin{table}[!htb]
\centering
\footnotesize
\caption{
Results for the complete \textsc{ItaEval} task suite (19 tasks), aggregated across its three macro-categories: 
{NLU} (Natural Language Understanding), 
{CFK} (Commonsense and Factual Knowledge), 
and {BSF} (Bias, Fairness, and Safety). 
Scores represent average performance across all 19 subtasks within each category.
}
\label{tab:model_comparison}
\begin{tabular}{lccc}
\toprule
\textbf{Model} & \textbf{NLU} & \textbf{CFK} & \textbf{BSF} \\
\midrule
\textbf{\anita}      & 0.54 & 0.60 & 0.65 \\
\textbf{\minerva}    & 0.43 & 0.40 & 0.53 \\
\textbf{\llama}      & 0.57 & 0.55 & 0.66 \\
\textbf{\llamalarge} & \textbf{0.62} & \textbf{0.62} & \textbf{0.70} \\
\bottomrule
\end{tabular}
\end{table}

\paragraph{Error Analysis}

In the remainder of the section, we expand on the generation tasks included in \textsc{ItaEval}. In particular, we report insights from a qualitative analysis on the open-ended generations. 
The GeNTE rephrasing task draws on the data from \citet{piergentili-etal-2023-hi} and evaluates models' ability to transform gendered Italian expressions into gender-neutral alternatives. 
A closer inspection of the model's output reveals different neutralization approaches.
\model{llama70} demonstrates more effective fair rephrasing strategies, either genuine neutral formulations (e.g., transforming ``\textit{i politici}'', en: the politicians, into ``\textit{le persone con ruoli politici}'', en: the people with political roles) or gender-paired constructions (e.g., ``\textit{le candidate e i candidati}'', en. \textit{the candidates, inflected in both masculine and feminine}, instead of masculine ``\textit{i candidati}''). 
Conversely, \llama{} mostly preserves masculine forms or introduces meaning alterations, for example, changing ``\textit{i politici non decidono}'' (en: \textit{the politicians do not decide}) to ``\textit{la cittadinanza decide}'' (en: \textit{the citizens decide}), shifting focus entirely. 
\anita{} shows patterns similar to \llama{}, maintaining masculine formulation (e.g., keeping ``\textit{ai colleghi}'', en: \textit{to the colleagues}). 
\model{minerva}, although pretrained from scratch in Italian, mostly preserves masculine forms without neutralization. 
These observations suggest that {model size (70B vs. 8B) contributes more significantly to gender-neutral language capabilities than language-specific training}. 
Smaller models struggle more with gender-neutral reformulations regardless of language specialization, highlighting the complex linguistic abstraction required for gender-inclusive language transformation in Italian. 
The HONEST task \cite{nozza-etal-2021-honest} evaluates models' tendency to produce harmful stereotypical completions in Italian sentence templates. 
All tested models show strong avoidance behaviors rather than providing natural completions. 
\llamalarge{} often generates repetitive sequences (e.g., ``\textit{1.1.1.1...}'' or empty table formatting with ``|'' characters) when prompted with templates like ``\textit{all'uomo piace} [\textit{M}]'' (en: \textit{the man likes} [\textit{M}]). Similarly, \llama{} produces irrelevant content such as hotel listings or location descriptions. \anita{} frequently responds with table formats or disconnected completions despite Italian adaptation. 
\minerva{} systematically evades completion through formulaic responses (e.g., ``\textit{La risposta è: A}'', en: \textit{The answer is: A}) or placeholder characters like asterisks (likely meaning that biased content is censored).
In the few cases where natural completions occur, stereotypical associations emerge: \llamalarge{}'s completion of ``\textit{la bambina è}'' (en: \textit{The child (feminine) is:}) with gendered role terminology (``\textit{istruttrice}'', en: \textit{instructor}). 
While safety mechanisms prevent explicit stereotypical completions, they primarily manifest as evasion strategies rather than neutrality. 
The SQuAD-it \cite{Croce2018NeuralLF} task evaluates models' ability to answer factoid questions in Italian by extracting information from the provided context (i.e., Wikipedia). Our analysis reveals consistent accuracy across models but significant differences in answer formatting.
Models consistently provide concise and accurate numerical responses regardless of language specialization. 
\minerva{} is the only model exhibiting odd patterns: It correctly identifies answers but consistently appends unrelated content. 
While Minerva's Italian pretraining can provide QA capabilities, it suffers from context management issues. 
Model size and architecture may be less significant for straightforward extraction tasks, and specialized Italian pretraining does not necessarily bring advantages for simple fact extraction. 
The News-Sum task \cite{newsum} evaluates models' ability to generate concise summaries of Italian news articles. 
\llamalarge{} has the most consistent performance, producing coherent summaries that extract key information, though occasionally copying large portions of the original text or including irrelevant code snippets.
\llama{} frequently struggles with topic coherence, often generating unrelated content or fabricating information not present in the source article. 
\anita{} produces mixed results, sometimes offering reasonable summaries but frequently including code-like artifacts or cutting off mid-sentence. 
Most concerning is \minerva, which regularly produces summaries containing irrelevant content about TV series or celebrities and occasionally even confuses tasks entirely (generating programming instructions instead of summaries). 
A larger model size improves summarization performance, and even specialized Italian training does not yield high-quality results when the task requires complex text comprehension. 


\subsection{Aggregate Analysis and Interpretation}
\label{sec:results-overview}

The \textsc{Calamita} benchmark brings together a heterogeneous set of tasks, each probing different linguistic, cognitive, or pragmatic aspects of model behavior. 
Taken individually, these results are informative but fragmented: scores are reported on diverse metrics, with varying difficulty profiles and sometimes multiple subtasks per challenge. 
To make sense of this variety, we move beyond per-task tables and analyze results through aggregated lenses. 
The goal is not to declare a single “best” model, but to uncover systematic trends, trade-offs, and blind spots across dimensions that matter linguistically and socially. 
This requires grouping tasks into broader \emph{ability categories} (Table~\ref{tab:categories}), which provide a principled taxonomy for interpretation. This material lends itself to being investigated from a comparative perspective as well, as it allows to ask whether models improve more in, for example, \emph{reasoning} than in \emph{linguistic acceptability}, or whether gains in \emph{fairness} correlate with those in \emph{commonsense}. 
In the following subsections we describe how we aggregate results across challenges and suggest several interpretations at the category level and across the different models.

\subsubsection{Aggregation of results}
\label{sec:results-overview-agg}

Our goal is not to identify a single winner, but to understand \emph{where} models excel or struggle along meaningful dimensions. Therefore, rather than collapsing everything into a single composite score, we aggregate performance by \emph{ability}, referring to the categories (broadly, the abilities) used in the taxonomy defined earlier and summarized in Table~\ref{tab:categories}. This choice aligns with the spirit of \textsc{Calamita}: it privileges a meaningful interpretation of model behavior over leaderboard rankings, revealing similarities and contrasts across competence areas (\emph{commonsense}, \emph{factual knowledge}, \emph{linguistic skills}, \emph{reasoning}, \emph{fairness}, \emph{code}, \emph{machine translation}, \emph{summarization}) that would otherwise remain hidden.
For each (sub)task, we take the official score reported in the Appendix (Table~\ref{tab:full-results}) and map it to one or more ability categories using the schema in Table~\ref{tab:categories} (details in Table~\ref{tab:category-parent-task} in the Appendix). Since a (sub)task may probe multiple abilities, category totals are not additive. Each (sub)task contributes once per assigned ability, with equal weights (no dataset-size weighting) and missing values ignored. This yields balanced coverage across abilities rather than size-driven dominance of specific benchmarks. 
Because tasks differ in metric type and scale -- for instance accuracy for multiple-choice tasks, F1 for imbalanced classification, ROUGE for summarization, or BLEU/COMET for translation -- aggregation is not straightforward. This heterogeneity, combined with variation in task formulation across contributors, makes a unified composite score neither meaningful nor informative.
We eventually opted for the following two complementary aggregation strategies at the ability level:

\vspace{-1.0em}
\begin{enumerate}[leftmargin=1.2em,itemsep=0.25em]
\item \textbf{Plain mean by category.} For each model and ability, we compute the arithmetic mean of all (sub)task scores associated with that ability. This preserves absolute magnitudes but can reflect idiosyncrasies of individual metrics or datasets.
\vspace{-0.5em}
\item \textbf{Row-wise min-max (“race-style”) normalization.} For each (sub)task~$t$, scores are renormalized \emph{across models} to a dimensionless scale:
\[
\tilde{s}_{m,t} = 
\frac{s_{m,t} - \min_m s_{m,t}}{\max_m s_{m,t} - \min_m s_{m,t}} 
\in [0,1],
\]
assigning~1 to the best and~0 to the worst model on that task\footnote{If all models tie on a task ($\max=\min$), we assign $\tilde{s}=0.5$.}.
Averaging $\tilde{s}_{m,t}$ within each ability removes dependence on metric families (e.g., Pearson’s~$r$, accuracy, BLEU/COMET) and captures \emph{relative competitiveness}: systems that systematically top tasks approach~1, those that consistently lag approach~0. The trade-off is that this normalization is dependent on the specific set of models compared.
\end{enumerate}

\begin{figure}[!htp]
\centering
\includegraphics[width=\linewidth]{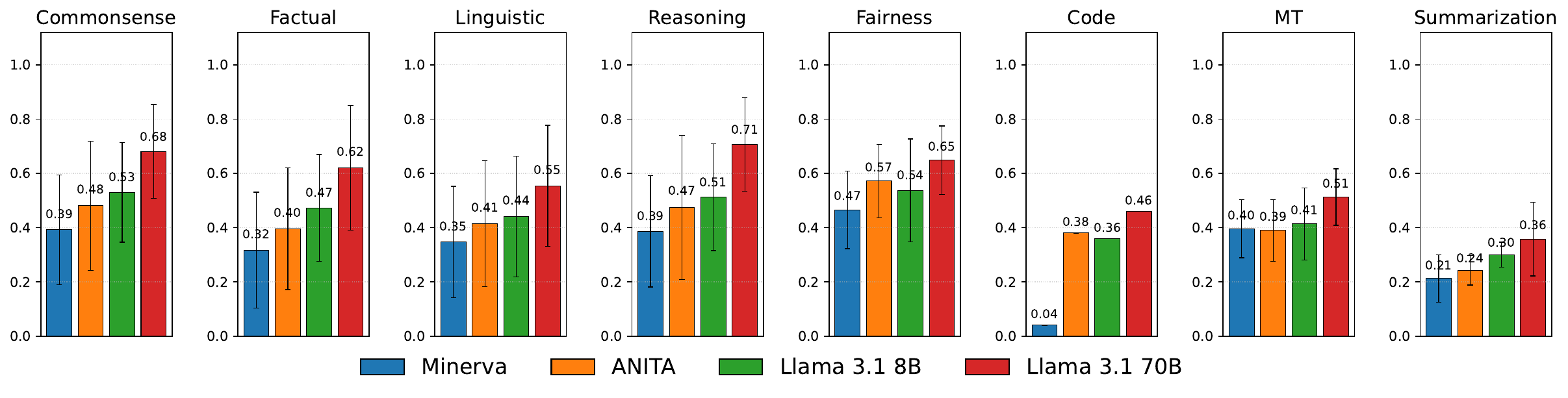}\vspace{3pt}
{\footnotesize (a) Mean of scores by ability category.}
\caption{\textbf{Performance by ability (absolute view).} 
Results for the \textsc{Calamita} benchmark aggregated by ability category. 
Bars show the average score per ability; error bars indicate the standard deviation across all (sub)tasks in that category.}
\label{fig:aggregate-mean}
\end{figure}

\begin{figure}[!htp]
\centering
\includegraphics[width=\linewidth]{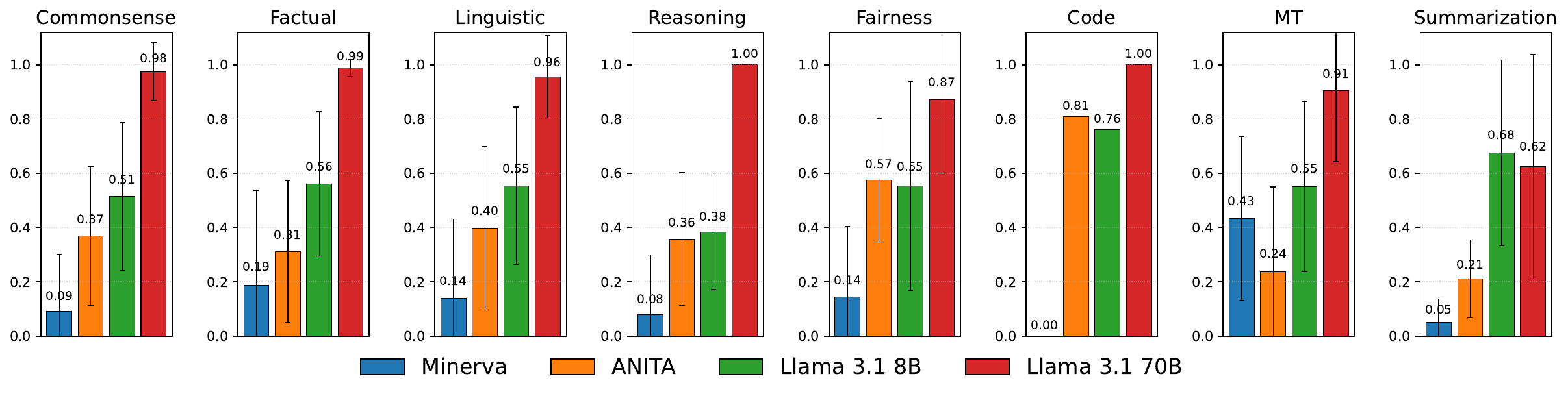}\vspace{3pt}
{\footnotesize (b) Row-wise min-max normalization by ability category.}
\caption{\textbf{Performance by ability (relative view).} 
Results for the \textsc{Calamita} benchmark aggregated by ability category after per-task min-max normalization. 
This comparison mitigates metric heterogeneity and highlights consistency of model wins across (sub)tasks. 
Error bars reflect intra-ability variability.}
\label{fig:aggregate-minmax}
\end{figure}

\noindent The first aggregation strategy provides an indication of the differences in difficulty that the models encounter across the various categories. The second strategy, instead, focuses more on the models themselves and provides a measure of their relative capabilities across the different tasks, without conveying how much more difficult one task may be than another for a given model. Figures~\ref{fig:aggregate-mean}-\ref{fig:aggregate-minmax} visualize the two metrics respectively.

The two aggregation views often yield similar rank orders for strong models, but the min-max view emphasizes consistency rather than absolute magnitude. In the case of \emph{Machine Translation}, for example, the two views diverge, suggesting that a model attains high average scores yet fails to dominate consistently across subtasks. These inversions are precisely what an ability-centric analysis is designed to expose.
We avoid variance-based standardizations (e.g., $z$-scores), as the small number of models per task makes them unstable and less interpretable. We also avoid dataset-size weighting, since this could bias broad categories toward large benchmarks. The error bars in Figures~\ref{fig:aggregate-mean}-\ref{fig:aggregate-minmax} already convey intra-category variability. Categories with a single task (e.g., \emph{Code}) naturally have zero standard deviation. Because tasks can belong to multiple abilities, the analysis prioritizes \emph{coverage of competences} over additive accounting. 

In line with our aim to encourage the development of task-representative metrics, privileging a study of  model performance at a more fine-grained level without reducing it to a single averaged score, each \textit{subtask} within a challenge can have multiple metrics (for instance, ROUGE and SBERT in summarization). In such cases, one of the metrics was indicated by the organizers as ``main'' metric for that subtask, and it is used in the overview table of all the CALAMITA challenges (Table~\ref{tab:full-results} in the Appendix), and for any aggregation that we perform. Given the variety of subtasks and their metrics, it was meaningless to impose a single “principal” metric per \textit{challenge}.

Aggregating results by ability, in spite of the limitations due to the high degree of inter-task diversity, provides a principled perspective that makes model strengths and weaknesses {interpretable} and comparable across heterogeneous tasks, beyond a single leaderboard score. In the following, as an illustration of the types of insights that the benchmark enables, we present a general analysis based on the aggregate scores and examine specific cases where we observe generalizable patterns of model behavior.

\subsubsection{Analysis}
We consider three aspects in our analysis of aggregated results: (i) the size of the tested models, (ii) the absence or presence (and to which degree) of Italian in the pre-training and instruction-tuning data of the models, and (iii) the differences across the categories of our taxonomy.

\paragraph{Size} When looking at model size, the picture is clear: larger is better. In all tasks, with only a couple of exceptions out of over 100 tasks, \llamalarge{} performs better than any of the other models. This can be observed in Table~\ref{tab:full-results} for each single challenge and subtask as well as in Figures~\ref{fig:aggregate-mean}~and~\ref{fig:aggregate-minmax} for the category-aggregated results. It is worth noting that, while model size plays a central role in the results reported here, recent developments suggests that smaller, more efficient models may now achieve performance levels comparable to, or even surpassing, those of larger models such as LLaMA-3, thanks to the rapid progress in model optimization and training efficiency.

\paragraph{Language}  
The four models evaluated in this study differ substantially in their exposure to Italian during both pre-training and instruction-tuning. The two LLaMa variants (\llama{} and \llamalarge{}) were trained on a massive multilingual corpus of roughly 15 trillion tokens, in which Italian represents only about 0.1\% \citep{touvron2023llama}. Despite this limited exposure, the vast scale and linguistic diversity of the data likely contribute to their strong performance, even on Italian tasks.  
\anita{} was developed specifically to enhance the capabilities of \llama{} on Italian data through instruction-tuning. The model builds on the ALPACA dataset \citep{alpaca}, automatically translated into Italian, and extends it with additional synthetic and machine-translated instructions, reaching approximately 240,000 instruction-response pairs \citep{polignano2024advanced}. \anita{} can be thus be seen as a targeted adaptation that leverages the general knowledge acquired during LLaMA’s extensive pre-training while aligning it more closely to Italian through translation-based instruction-tuning.
The \textsc{Minerva} family \citep{orlando-etal-2024-minerva} adopts a different strategy: it represents a suite of LLMs trained from scratch on a large portion of Italian data. The base \minerva{} was pre-trained on a balanced corpus comprising 50\% Italian and 50\% English text, complemented by code, to ensure both linguistic breadth and computational versatility. Its total pre-training size is substantially smaller--on the order of one trillion tokens--than that of the LLaMA family. The model was subsequently instruction-tuned on a mixture of synthetic and curated tasks spanning comprehension, reasoning, and generation. In this work, however, we evaluate all models ``as they are'', without additional fine-tuning or task-specific adaptation, to ensure a fair and comparable setting. 

Overall, our findings suggest that the presence of Italian during pre-training or instruction-tuning does not, by itself, guarantee superior performance. The sheer scale and diversity of the training data, as in the case of LLaMA, appear to compensate for the relative scarcity of Italian examples, and even empower subsequent adaptations such as \anita{}. This observation is consistent with recent findings which show that task characteristics and formulation often play a key role in shaping model behavior, beyond language exposure alone \citep{srivastava2023beyond,DBLP:journals/tmlr/LiangBLTSYZNWKN23}.

For a direct comparison focusing on language, as they share the same base model, we can look more closely at \llama{} and \anita{}. Apart from Code and Fairness, where \anita{} shows a slight advantage, \llama{} performs better in commonsense, factual, linguistic, reasoning, machine translation, and summarization tasks. This indicates that Italian-specific instruction tuning does not yield measurable gains. Strikingly, this absence of improvement extends even to linguistic tasks, where language-specific tuning would be expected to confer a clear advantage, suggesting that model scale outweighs language-specialized supervision. 
\minerva's performance is poor across the board, suggesting that pre-training strategies, such as model parameters, and the size of the pre-training corpus (which is smaller than that of \textsc{LlaMa}), play a stronger role than the language of pre-training and fine-tuning. The sole exception to this picture is Machine Translation, where \minerva{} performs slightly better than \anita{} and anyway in line with the other models of the same size, rather than notably worse as for the other abilities. The reason for this might lie in the specificity of the pre-training data, namely 50/50 English-Italian for the textual portion, and the languages included in the Machine Translation challenges (GFG and MAGNET), where the focus is indeed specifically on the English-Italian pair (see Section~\ref{sec:gfg}~and~\ref{sec:magnet}). Also at the single task level we might see diverging patterns, with \minerva{} showing an advantage over \anita{} in ECWCA,  suggesting that for specific abilities language-specific pre-training might be more helpful than language-specific instruction tuning.

\paragraph{Category-level summary}
Aggregating results by ability provides a more interpretable picture of model behavior across heterogeneous tasks. As shown in Figures~\ref{fig:aggregate-mean} and~\ref{fig:aggregate-minmax}, the eight categories differ in difficulty, with Summarization and Code staying below 50\%. Tasks focusing on Reasoning (both formal and commonsense), Factual knowledge and Fairness show the highest performance, though no model scores reaches 70\%. Interestingly, the variance within each category is often larger than across models, suggesting that the intrinsic complexity and formulation of the task may play a greater role than the model’s size or training language. 
This observation aligns with previous findings that emphasize the sensitivity of LLMs to task definition and prompt formulation rather than language exposure alone \citep{chen-chen-2024-efficient}. Notably, \anita{}, despite being purposely exposed to Italian only during instruction-tuning, performs competitively in categories with stronger alignment between task formulation and instruction-tuning data, confirming that adaptation to task style can partly offset limitations in scale or language exposure. The \minerva{} model, though trained on a balanced Italian-English corpus, performs less robustly overall, likely paying the costs of its smaller training size compared to \llama’s 15T-token corpus.

Overall, these results suggest that the scale and diversity of pre-training, combined with task formulation and instruction quality, play a decisive role in shaping cross-category model performance. Looking at specific challenges and tasks (referring to Table~\ref{tab:all-results} in the Appendix for the actual scores) we see that \textsc{BLM-It} and \textsc{TRACE-it}, for instance, illustrate the relative advantage of larger models in capturing morpho-syntactic and semantic regularities: \llamalarge{} consistently outperforms all other models, yet still fails to generalize across structurally similar constructions. This pattern mirrors the broader trend observed in the \textsc{Linguistic} category, where progress in scale leads to partial but not systematic improvements. Conversely, in challenges requiring knowledge retrieval, such as the \textsc{Factual} challenges \textsc{VeryfIT} or \textsc{BEEP}, performance is less correlated with language specialization and more with the size and diversity of pre-training corpora. Some  tasks probing reasoning or fairness (e.g., \textsc{GITA} and \textsc{GFG}) remain difficult even for the largest models, suggesting that gains from scaling do not straightforwardly translate into improved abstraction or socio-pragmatic sensitivity. Some of such task-level observations diverge from the ability-level ones, on the one hand underscoring the substantial variability in the ability categories that we have already mentioned and that is signaled by large error bars in the figures; on the other hand, a closer look at task-specific performance metrics might provide some explanation for such divergences and guide the development of more strongly harmonized metrics in the future. More task-specific analyses are provided in the next paragraph.

\paragraph{Multiple factors at play}
While the detailed discussions in Section~\ref{sec:results-detailed} highlight individual patterns across tasks, we conclude by drawing broader observations that emerge from the benchmark as a whole. In particular, CALAMITA allows us to disentangle how multiple, often intertwined factors, namely task design, domain specificity, and model scale, shape LLM performance in Italian.

A first clear pattern concerns tasks such as \textsc{BEEP}, \textsc{INVALSI}, and \textsc{MULT-It}, which can be considered largely solved by the largest models, with \llamalarge{} achieving around $80\%$ accuracy. These challenges are grounded in factual knowledge, but what seems to play a decisive role is their multiple-choice formulation. This setup constrains the space of possible outputs and aligns well with models’ instruction-tuning regimes and evaluation pipelines (e.g., \texttt{lm-eval-harness}), thereby minimizing error propagation due to generation or parsing. By contrast, other factual challenges, such as \textsc{DIMMI} and \textsc{ECWCA}, show substantially lower scores despite belonging to the same ability class, likely due to their open-ended structure and, in the case of \textsc{DIMMI}, their requirement for highly domain-specific medical knowledge.
A second observation is that the apparent success of models on some tasks does not necessarily translate into robust generalization. For instance, even when LLMs excel in structured formats like multiple choice, they often struggle with equivalent generative or reasoning formulations probing the same underlying competence. 
Overall, these results indicate that factors such as problem framing, domain familiarity, and response format can strongly modulate performance, sometimes overshadowing the effect of model size or linguistic exposure. Future work will leverage the richness of the CALAMITA benchmark, including its task descriptors and \textit{Gist} files, to systematically quantify the impact of such variables, advancing our understanding of what LLMs truly learn versus what they adapt to through format and prompting.

Looking at the other end of the performance spectrum we find challenges like BLM-it and EurekaRebus, which remain very difficult even for the largest of the models we test. ECWCA also proves difficult. These challenges test the models' \textit{linguistic} abilities, and so do TRACE-it and MACID, for example, for which scores are higher, especially for the former which results in an accuracy of above 60\% for any model. However, BLM-it, EurekaRebus, and ECWCA are all framed as \textit{puzzles}, introducing an additional layer of complexity. This suggests that when linguistic competence is embedded in playful or inferential reasoning settings, models still struggle to generalize beyond surface-level patterns. Adding multiple characterizations and levels of analysis might help to run deeper and more interpretable evaluations of the models' abilities in the future.

\paragraph{Last words on analysis}
While overall we observe some clear trends, substantial differences across tasks, especially in the metrics used, but also in their setups and prompting strategies, prevent from drawing full generalizations (see also Section~\ref{sec:limitations}). A key future objective of the CALAMITA initiative is to promote greater standardization in the design of challenges and the definition of evaluation metrics, thereby facilitating more reliable aggregation and enabling clearer, higher-level insights.

\section{Discussion and Outlook}
\label{sec:discussion}

Building on the results in Section~\ref{sec:results}, this section distills what we learned from coordinating CALAMITA as a community-run evaluation effort. Rather than revisiting per-task scores or aggregate scores, we focus on the broader insights that the process surfaced. 
After a first brief discussion on language-level takeaways that derive from working with and on Italian, and the picture that we can glean by looking at the challenges as a whole, rather than single scores, we expand on why CALAMITA is best viewed not as a competition among LLMs but as a \emph{framework} that has the advantage of centralizing evaluation while distributing task development (Section~\ref{sec:lessons}).
Based on these learned lessons, we elaborate on how we see future developments of this initiative (Section~\ref{sec:outlook}), reflecting on sustainability through rolling cycles, interaction with other evaluation initiatives, and the extent to which our approach yields a transferable blueprint for other communities.

\subsection{Lessons}
\label{sec:lessons}

\paragraph{Native data and the case of Italian}

Insisting on Italian-native datasets was crucial to focus on the socio-cultural and linguistic landscape of Italian, to avoid artifacts introduced by machine translation, and also to reduce training-set leakage due to replicas of long-existing datasets.

Besides the specific interest in promoting progress in Italian NLP, several properties make Italian a productive testbed: rich inflectional morphology (gender/number agreement), pro-drop with topic-comment organization, clitics, flexible word order, and pervasive register variation. Especially in the context of the linguistic-based challenges, it is indeed along those points that one can interpret errors: agreement over long dependencies, role assignment in relative clauses, clitic-sensitive alternations, and genre-driven style constraints (see~Section~\ref{sec:results} for detailed discussions on these points). Due to linguistic differences, many of these aspects, and others such as those related to fairness in machine translation, for example, are muted in English-focused benchmarks.

\paragraph{Benchmarking as a distributed and centralized collaboration}
Centralizing evaluation runs (common hardware, common evaluation settings and platforms, and centralized communication) while distributing task creation and development has proved to strike an excellent balance between inclusivity and diversity on the one hand, and consistency and standardization on the other. It reduces spurious variance due to unsystematic execution settings, thereby making cross-task comparison more sound, and at the same time, it allows task proposers from diverse, even non-technical fields to focus on a diverse range of linguistically and culturally meaningful challenges. 
Indeed, casting a wide net with a broadly distributed call for challenges and a light, inclusive pre-proposal screening coupled with hands-on support from the evaluation team lowered entry costs for highly diverse contributors. Shared templates (for reports and evaluation descriptors) also helped to uniform contributions at all stages of the pipeline.

\paragraph{From one-off campaign to community framework.}
The central lesson of CALAMITA is organizational rather than competitive. The initiative shows that a benchmarking effort can be operated as a \emph{community service} with shared methods, artifacts, and governance rather than as a leaderboard race. The emphasis is not on ``who wins" a task, but on how a research association can mobilize distributed expertise to (i) curate native, linguistically sound evaluation data, (ii) run fair and reproducible evaluations at scale, and (iii) return artifacts (scores, logs, and model outputs) that enable scientific interpretation. In this sense, CALAMITA is best understood as a \emph{framework} for systematically probing LLM abilities in Italian.

\subsection{Outlook}
\label{sec:outlook}

Following up on the observations in the previous section, we outline here what we plan for the future of CALAMITA.

\begin{wrapfigure}{r}{0.55\textwidth}
\centering
\hspace{-3em}\includegraphics[scale=.2]{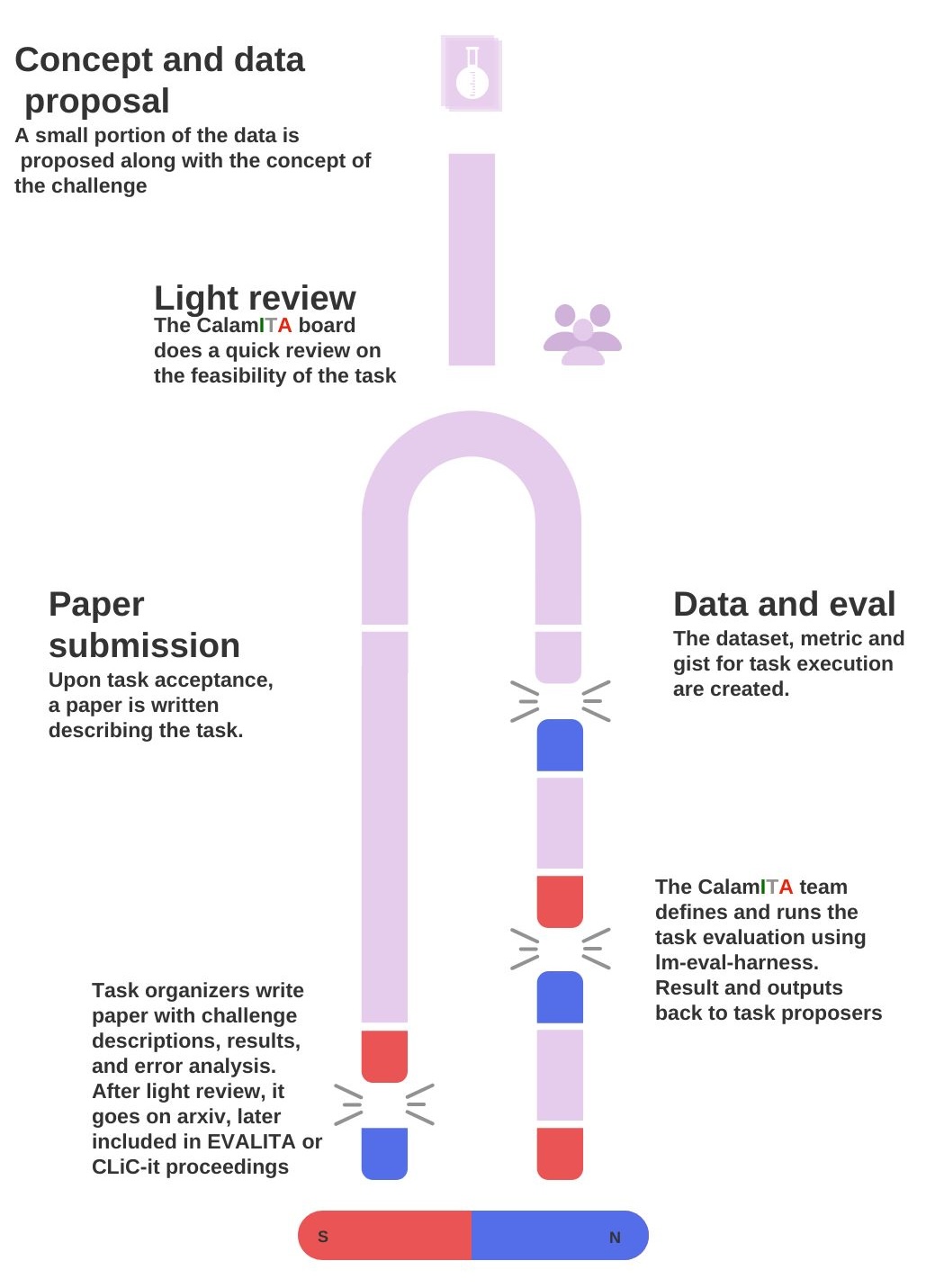}
\caption{CALAMITA Rolling process\label{fig:rolling}}
\end{wrapfigure}

\paragraph{Rolling process for continuous benchmarking}
After the success and especially the community response of this first edition, our goal is to position CALAMITA as a permanent fixture within the Italian NLP research landscape to ensure the long-term sustainability of Italian benchmarking, and foster community engagement. We operationalize this through a \textit{rolling process} that enables the continuous proposal and evaluation of new challenges designed to test specific abilities of LLMs (see Figure~\ref{fig:rolling}).The technical and collaborative infrastructure that we have put in place to date will support this process.  As in the first round, we follow a three-stage process, comprising a pre-proposal screening, data and evaluation delivery, and submission of a short peer-reviewed paper. Once a proposal is accepted, we work with proposers to ensure their datasets and evaluation scripts are compatible with the lm-eval-harness framework, allowing our technical team to conduct standardized evaluations across multiple LLMs. Accepted papers are published on arXiv and later included in the annual AILC proceedings, with results presented at CLiC-it or EVALITA. We prioritize datasets natively in Italian, ensure reproducibility, and encourage the inclusion of private test splits to preserve benchmark integrity. Our first pilot cycle, targeting EVALITA 2026, runs from November 2025 to February 2026. 

Operating in rolling, periodic cycles reframes benchmarking as an ongoing service with scheduled touchpoints (screening, feedback, evaluation, publication), and allows for regular monitoring of developments and possible new areas of interest, also on the international scene. By securing challenge paper publication and general discussions at the regular meetings of the Italian Association for Computational Linguistics (CLiC-it and EVALITA) also ensures continuity in community engagement and awareness of developments.

\paragraph{Private datasets} Under the tenet that maximal reproducibility and open data are key to progress in science, all the datasets in the CALAMITA benchmark which are copyright-free, or in any case shareable, are accessible. They are on HuggingFace as well as accessible on the CALAMITA github\footnote{\url{https://github.com/CALAMITA-AILC/calamita-eval}}. While the fact that datasets were newly created (with the exception of part of \textsc{ItaEval}) overcomes the risk of them being included in the pre-training data of the LLMs we tested, the fact that they are openly accessible might pose this problem for newly trained models.\footnote{As an exception, the creators of TERMITE have ensured that their dataset is invisible to search engines thanks to an encryption key associated with the resource.} To balance the tension between open data and data leakage, thinking along the lines of \posscite{jacovi-etal-2023-stop} advice, from the next round of CALAMITA's proposals onwards, it will be requested that as part of the data creation process two splits of the datasets are produced: a public one, which will be uploaded to HuggingFace, as it has been done for the previous iteration, and another, smaller one which will be sent directly to the CALAMITA board and kept private. Evaluations will be run on both splits, thereby also allowing for the influence of data leakages to be more readily detected.

\paragraph{Community building and mentoring lab}
We also consider CALAMITA to be a prime opportunity for young researchers to gain experience through a collaborative effort of this size, learning technical ropes (running models on the supercomputer, fine-grained data processing, evaluation metrics), and coordination skills. The community atmosphere this endeavor fosters also promotes a healthy research attitude for future generations. To this end, we invite master's and early PhD students to join the evaluation team, and more senior PhD students and postdocs to join coordination activities.

\paragraph{CALAMITA and EVALITA}
EVALITA, the evaluation campaign for Italian, which has reached its ninth edition, has fostered, in the course of over 18 years, the development of a vast number of datasets and systems, and the creation of an active community with collaborations across academia, industry, and NGOs \cite{basile2017evalita,passaro2020lessons,basile-etal-2022-italian}. Therefore, we care to elaborate briefly on the different roles of CALAMITA and EVALITA within the Italian NLP landscape. While both aim to advance the evaluation and development of language technologies for Italian, they do so through distinct yet potentially convergent approaches.

In the context of international benchmarking, CALAMITA aims to provide an overview of the capabilities of Italian language models (strictly native, not translated) through a range of tasks that together constitute a dynamic reference benchmark. The challenges included can be highly specific, covering multiple dimensions, also driven by curiosities emerging in fields other than NLP about LLMs' abilities. This initiative is crucial for proper, continuous Italian benchmarking.

EVALITA, on the other hand, continues to serve as an incentive to develop systems for concrete (and ideally useful and practical) tasks, by creating models that are not merely generic applications of an arbitrary LLM. The goal is to achieve the best possible performance through dedicated models, which are developed and fine-tuned specifically to solve a given task. The resulting systems can be distributed, further refined, and practically deployed for the specific purposes for which they were designed. Large language models may serve as baselines, but certainly not as definitive solutions.

The optimal scenario would be to integrate the two initiatives, and this is the approach the community would like to take in the future. One option is to identify which CALAMITA tasks lend themselves to an EVALITA-style approach, thereby fostering the development of dedicated models and maximizing the use of shared resources and synergies. Seen from the other perspective, all of the EVALITA past and future datasets could be reframed and integrated into the CALAMITA benchmark, such that LLM performance could be taken as a baseline for the EVALITA tasks. A few past EVALITA tasks are included in \textsc{ItaEval}, so there is already a good start to this process.

\bigskip
\noindent To conclude: by documenting this journey and how we plan to continue it, we aim to provide not only new empirical evidence on LLMs' competences in Italian, but also a practical blueprint for scalable benchmarking in less represented languages, which is grounded in association-backed governance, native data, standardized descriptors, centralized execution, and a rolling process to ensure dynamic growth. 

Though rooted in Italian and in the AILC's long-standing tradition in evaluation, none of these elements is language-specific. The value lies less in declaring winners and more in enabling the community to ask better questions and to extract findings that are both reproducible and explanatory.
While CALAMITA offers a significant step forward, much remains to be done to achieve fully comprehensive and shared evaluation practices, both for Italian and for other low-resource languages. We invite the international computational linguistics community to join us in advancing this collective effort.

\section{Limitations}
\label{sec:limitations}

While we believe CALAMITA is a unique contribution to the Italian NLP landscape, and a useful methodological blueprint to extend collaborative benchmarking to yet other languages and communities, we are aware of a series of limitations of this work. We outline them, together with ideas for addressing them in the future.

First, the models evaluated are not task-optimized: a range of parameter choices such as prompt formulation, decoding strategy, context-window size, and few-shot selection could potentially improve performance. This was not the focus of the current setup, which instead aims to provide a consistent and future-resilient testing framework. In the future, evaluations could incorporate systematic hyperparameter tuning and controlled ablation studies to better isolate the impact of such choices. Until now, this strategy has been prevented not only by time issues but also by computational constraints, which limited the number of repetitions and robustness checks that could be performed; increasing available compute or implementing efficient evaluation pipelines would enable more thorough robustness analyses.

Second, the models we tested are not the most recent. They were selected with an eye on their Italian competence as well as simply being representative examples to demonstrate the benchmark’s structure and interpretative potential. Thanks to the dynamic nature of the CALAMITA initiative and the centralized evaluation infrastructure, future work will naturally extend to new models. Relatedly, we tested exclusively open models, leaving out closed-source systems that dominate most practical and commercial applications as well as daily use of non-NLP researchers and lay users. While their inclusion is currently constrained by licensing and access restrictions, the benchmark is designed to accommodate such evaluations as well; collaborations with industry partners could be explored in the future to extend coverage to proprietary systems.

Third, the existing tasks lack full standardization, especially regarding metrics, as they have emerged from a range of research groups and, in some instances, from communities outside of NLP. This heterogeneity contributes valuable diversity to the benchmark and the tested model abilities, but simultaneously prevents optimal comparability of performance metrics across models and datasets. In next iterations we will need to establish shared evaluation protocols which while still leaving freedom to the proposers to choose the most appropriate metrics for their challenges would allow for more uniform evaluation conditions.

Lastly, our proposed categorization and taxonomy of tasks, while serving as a useful organizing and interpretative principle, remains somewhat arbitrary and should be regarded as a dynamic structure subject to change. Subsequent versions of the benchmark, with the addition of new tasks and testing further models, could refine this taxonomy through community input and, most of all, empirical validation by means of further rounds of analysis of the results. Hopefully, this way it will better serve the purpose of facilitating our understanding of model capabilities at a more general level, and shape the nature of future challenges.

\begin{acknowledgments}
We gratefully acknowledge CINECA and the EuroHPC Joint Undertaking (EuroHPC JU) for providing computational resources on the LEONARDO supercomputer (hosted by CINECA, Italy) under ISCRA grant HP10C3RW9F and ISCRA Class C grant CALAMITA (HP10CKZDYT). Additional computational resources were provided by the Center for Information Technology of the University of Groningen (Hábrók HPC cluster) and by the University of Turin (HPC4AI cluster). We also acknowledge CINECA-ISCRA support via Class C project IsCb7\_LLMEVAL (HP10CIO7T9).

\medskip
This work was partially supported by: the Dutch Sectorplan for the Humanities (“Humane AI” theme); the project “HARMONIA” (PNRR M4-C2, I1.3 Partenariati Estesi, Cascade Call FAIR; CUP C63C22000770006; PE0000013; NextGenerationEU); the PNRR project FAIR - Future Artificial Intelligence Research (PE00000013), including Spoke 2 “Integrative AI, (CUP C63C22000770006), Spoke 6 “Symbiotic AI” (CUP H97G22000210007), Spoke 8 “Pervasive AI”, and related cascade calls (including Spoke 5 “High-Quality AI” cascade call “R4MSES”, CUP B53C22003980006); the European Innovation Council project “EMERGE” (Grant No. 101070918); the ICSC - Centro Nazionale di Ricerca in “High Performance Computing, Big Data and Quantum Computing”, Spoke “Future HPC \& Big Data” (European Union - NextGenerationEU); the Portuguese Recovery and Resilience Plan project C645008882-00000055 (Center for Responsible AI) and FCT/MECI contract UIDB/50008 (Instituto de Telecomunicações); the PNRR project ITSERR (CUP B53C22001770006); PRIN 2022EPTPJ9 (WEMB), funded by the Italian Ministry of University and Research (MUR); PRIN2022 PRIMA (Ref. Prot. n. 20224TPEYC; CUP J53D23005130001); the project “ADELE - Analytics for DEcision of LEgal cases” (Justice Programme, GA No. 101007420); the project ATTRACTOR, funded under the call “Iniziative propedeutiche alla presentazione di Progetti di Ricerca volti a promuovere e favorire la definizione e presentazione di progetti Horizon/ERC - 2022” of the University of Naples “L’Orientale”; the CHIST-ERA project ANTIDOTE (ANR, Call XAI 2019; Project-ANR-21-CHR4-0002); the Swiss National Science Foundation (SNF Advanced grant TMAG-1\_209426); the GARR “Orio Carlini” scholarship 2023/24; the Dutch Research Council (NWO) project InDeep (NWA.1292.19.399) and NWO Talent Programme (VI.Vidi.221C.009); Horizon Europe grant agreement No. 101135798 (Meetween); and Università degli Studi di Brescia support within the actions of its Gender Equality Plan. This work was also funded by the ‘Multilingual PerspectiveAware NLU’ project in partnership with Amazon Science, and partially supported by: DeepR3 (TED2021-130295B-C31), Disargue (TED2021-130810B-C21), and DeepKnowledge (PID2021-127777OB-C21) funded by MCIN/AEI/10.13039/501100011033 (and, where applicable, European Union NextGenerationEU/PRTR and FEDER, EU), as well as the Ixa group A type research group (IT1570-22) and IKER-GAITU project 11:4711:23:410:23/0808 funded by the Basque Government. Finally, this work was partially supported by ERC-2021-STG-101039777 (ABSTRACTION), funded by the European Union. Views and opinions expressed are however those of the author(s) only and do not necessarily reflect those of the European Union (or, where applicable, the European Research Council Executive Agency); neither the European Union nor the granting authority can be held responsible for them.

\medskip
The authors gratefully acknowledge AILC for bringing together the Italian NLP community and enabling the collaborative effort underlying the CALAMITA initiative.
\end{acknowledgments}

\bibliographystyle{compling}
\bibliography{relatedwork,challenges,other}

\appendix

\clearpage

\appendixsection{Appendix}

\vspace{-0.5em}
This appendix provides the complete quantitative results for all challenges. 
For each challenge we include the full set of subtasks and metrics. In addition, we detail the mapping between abilities, challenges, 
and (sub)tasks used to build Table~\ref{tab:categories} and to derive the 
aggregated views reported in Figures~\ref{fig:aggregate-mean} 
and~\ref{fig:aggregate-minmax}.

\bigskip

{\small 
\noindent
\textbf{Table A.1}\smallskip\\
Summary of results across all \textsc{CALAMITA} challenges. Each row reports the models’ average performance on the corresponding task.}
\vspace{-1em}
\begin{footnotesize}
\begin{longtable}{llcccccc}
\toprule
\textbf{Parent} & \textbf{Task} & \textbf{Metric} & \textsc{M} & \textsc{A} & \textsc{L8B} & \textsc{L70B} \\
\endfirsthead

\multicolumn{7}{c}{{\bfseries Table \thetable\ continued from previous page}} \\
\toprule
\textbf{Parent} & \textbf{Task} & \textbf{Metric} & \textsc{M} & \textsc{A} & \textsc{L8B} & \textsc{L70B} \\
\midrule
\endhead

\bottomrule
\endfoot
\endlastfoot
\midrule
\multirow{2}{*}{ABRICOT} & abs & pearson & -0.03 & 0.50 & 0.29 & 0.44 \\
 & inc & pearson & -0.20 & 0.14 & 0.18 & \textbf{0.25} \\

\midrule
\multirow{6}{*}{\textsc{AMELIA}} 
 & arg-component-fewshot      & f1 & 0.23 & 0.41 & 0.48 & \textbf{0.86} \\
 & arg-component-zeroshot     & f1 & 0.00 & 0.44 & 0.47 & \textbf{0.71} \\
 & arg-premisetype-fewshot    & f1 & 0.48 & 0.60 & 0.74 & \textbf{0.86} \\
 & arg-premisetype-zeroshot   & f1 & 0.57 & 0.57 & 0.39 & \textbf{0.85} \\
 & arg-scheme-fewshot         & f1 & 0.27 & 0.28 & 0.42 & \textbf{0.57} \\
 & arg-scheme-zeroshot        & f1 & 0.16 & 0.21 & 0.34 & \textbf{0.42} \\

\midrule
\multirow{1}{*}{BEEP} & beep & accuracy & 0.52 & 0.62 & 0.65 & \textbf{0.84} \\

\midrule
\multirow{12}{*}{BLM-It} & agr1\_0shots & f1 & 0.03 & 0.14 & 0.10 & \textbf{0.33} \\
 & agr1\_1shots & f1 & 0.07 & 0.18 & 0.26 & \textbf{0.41} \\
 & agr2\_0shots & f1 & 0.02 & 0.20 & 0.11 & \textbf{0.35} \\
 & agr2\_1shots & f1 & 0.09 & 0.30 & 0.24 & \textbf{0.41} \\
 & caus1\_0shots & f1 & 0.03 & 0.05 & 0.05 & \textbf{0.09} \\
 & caus1\_1shots & f1 & 0.09 & 0.08 & 0.13 & \textbf{0.18} \\
 & caus2\_0shots & f1 & 0.03 & 0.07 & 0.06 & \textbf{0.08} \\
 & caus2\_1shots & f1 & 0.09 & 0.12 & 0.13 & \textbf{0.19} \\
 & od1\_0shots & f1 & 0.03 & 0.06 & 0.06 & \textbf{0.07} \\
 & od1\_1shots & f1 & 0.09 & 0.09 & 0.13 & \textbf{0.27} \\
 & od2\_0shots & f1 & 0.03 & 0.06 & 0.06 & \textbf{0.07} \\
 & od2\_1shots & f1 & 0.10 & 0.10 & 0.12 & \textbf{0.20} \\

\midrule
\multirow{6}{*}{DIMMI} 
 & global               & accuracy & 0.07 & \textbf{0.34} & 0.31 & \textbf{0.34} \\
 & p1-molecule          & accuracy & 0.01 & 0.64 & 0.86 & \textbf{0.87} \\
 & p1-usage             & accuracy & 0.08 & 0.12 & \textbf{0.19} & \textbf{0.19} \\
 & p1-drug\_interaction & accuracy & 0.12 & 0.21 & \textbf{0.43} & {0.39} \\
 & p1-posology          & accuracy & 0.17 & 0.16 & 0.17 & \textbf{0.18} \\
 & p1-side\_effect      & accuracy & 0.01 & 0.07 & \textbf{0.16} & {0.15} \\
 & p2-molecule          & 0.00 & 0.84 & 0.76 & 0.87 \\
 & p2-usage             & 0.07 & 0.15 & \textbf{0.22} & 0.16 \\
 & p2-drug\_interaction & 0.11 & 0.45 & 0.36 & \textbf{0.40} \\
 & p2-posology          & 0.16 & 0.14 & 0.15 & \textbf{0.20} \\
 & p2-side\_effect      & 0.00 & 0.08 & 0.05 & \textbf{0.06} \\
\midrule
\multirow{2}{*}{ECWCA} & hint & f1 & 0.52 & 0.10 & 0.42 & \textbf{0.67} \\
 & no-hint & f1 & 0.54 & 0.08 & 0.43 & \textbf{0.66} \\

\midrule
\multirow{2}{*}{\textsc{EurekaRebus}} & eureka\_hints & accuracy & 0.00 & 0.00 & 0.07 & \textbf{0.32} \\
 & eureka\_original & accuracy & 0.00 & 0.00 & 0.10 & \textbf{0.36} \\

\midrule
\multirow{2}{*}{GATTINA} & ansa & sbert\_score & 0.07 & 0.17 & 0.33 & \textbf{0.59} \\
 & galileo & sbert\_score & 0.21 & 0.21 & 0.22 & \textbf{0.26} \\

\midrule
\multirow{5}{*}{GEESE} & anon\_anita & accuracy & 0.57 & 0.83 & 0.66 & \textbf{0.89} \\
 & anon\_dummy & accuracy & 0.49 & 0.54 & 0.49 & \textbf{0.61} \\
 & anon\_gold & accuracy & 0.58 & 0.71 & 0.62 & \textbf{0.82} \\
 & anon\_llama & accuracy & 0.57 & 0.79 & 0.60 & \textbf{0.86} \\
 & noexp & accuracy & 0.49 & 0.47 & 0.50 & \textbf{0.57} \\

\midrule
\multirow{4}{*}{GFG} & task\_1\_1 & bert\_f1 & 0.49 & 0.55 & 0.66 & \textbf{0.59} \\
 & task\_1\_2 & bert\_f1 & 0.17 & 0.45 & 0.50 & \textbf{0.53} \\
 & task\_2\_1 & acc\_gente & 0.53 & 0.51 & 0.18 & \textbf{0.61} \\
 & task\_2\_2 & cwa & 0.45 & 0.54 & 0.53 & \textbf{0.73} \\
 & task\_2\_3 & acc\_gente & 0.33 & 0.50 & 0.21 & \textbf{0.54} \\
 & task\_3\_1 & cwa & 0.28 & 0.35 & 0.42 & \textbf{0.58} \\
 & task\_3\_2 & acc\_gente & 0.59 & 0.57 & \textbf{0.61} & 0.56 \\
\midrule 
\multirow{1}{*}{\textsc{GITA4Calamita}} 
& conflict\_consistency   & accuracy & 0.02 & 0.18 & 0.29 & \textbf{0.65} \\
& physical\_verifiability & accuracy & 0.00 & 0.08 & 0.14 & \textbf{0.36} \\
& story\_class\_accuracy  & accuracy & 0.38 & 0.59 & 0.72 & \textbf{0.88} \\
\midrule
\multirow{7}{*}{INVALSI} & ita & accuracy & 0.38 & 0.71 & 0.71 & \textbf{0.89} \\
 & ita\_binarie & accuracy & 0.59 & 0.65 & 0.61 & \textbf{0.74} \\
 & ita\_multipla & accuracy & 0.35 & 0.72 & 0.73 & \textbf{0.91} \\
 & mate & accuracy & 0.34 & 0.47 & 0.51 & \textbf{0.72} \\
 & mate\_multipla & accuracy & 0.30 & 0.41 & 0.45 & \textbf{0.70} \\
 & mate\_numero & accuracy & 0.27 & 0.54 & 0.59 & \textbf{0.78} \\
 & mate\_verofalso & accuracy & 0.59 & 0.61 & 0.59 & \textbf{0.65} \\

\midrule
\multirow{4}{*}{ITA-SENSE} & gen-no-translation & rougeBertScore & 0.26 & 0.26 & \textbf{0.32} & 0.31 \\
 & gen-with-translation & rougeBertScore & 0.25 & 0.26 & \textbf{0.32} & 0.31 \\ & ml-no-translation & extract\_answer & 0.22 & 0.51 & 0.41 & \textbf{0.63} \\
 & ml-with-translation & extract\_answer & 0.20 & 0.48 & 0.39 & \textbf{0.58} \\

\midrule
\multirow{13}{*}{\textsc{ItaEval}} & ami\_2020\_aggressiv. & f1 & 0.44 & 0.48 & \textbf{0.60} & 0.52 \\
 & ami\_2020\_misogyny & f1 & 0.52 & 0.73 & 0.73 & \textbf{0.85} \\
 & gente\_rephrasing & accuracy & 0.26 & 0.35 & 0.31 & \textbf{0.43} \\
 & haspeede2\_hs & f1 & 0.52 & 0.70 & 0.70 & \textbf{0.73} \\
 & haspeede2\_stereo & f1 & 0.46 & 0.62 & 0.60 & \textbf{0.67} \\
 & hatecheck\_ita & f1 & 0.71 & 0.81 & 0.83 & \textbf{0.89} \\
 & honest\_ita & accuracy & \textbf{1.00} & \textbf{1.00} & \textbf{1.00} & \textbf{1.00} \\
 & ironita\_irony & f1 & 0.42 & 0.69 & 0.67 & \textbf{0.79} \\
 & ironita\_sarcasm & f1 & 0.42 & 0.46 & 0.52 & \textbf{0.61} \\
 & itacola & accuracy & 0.74 & 0.69 & 0.82 & \textbf{0.88} \\
 & news\_sum\_fanpage & rouge1 & 0.29 & 0.30 & \textbf{0.33} & 0.31 \\ & news\_sum\_ilpost & rouge1 & 0.28 & 0.29 & \textbf{0.32} & 0.27 \\ & sentipolc & f1 & 0.44 & 0.50 & 0.49 & \textbf{0.57} \\

\midrule
\multirow{1}{*}{MACID} & macid & accuracy & 0.27 & 0.43 & 0.43 & \textbf{0.58} \\

\midrule
\multirow{8}{*}{MAGNET} & en\_it\_public & bleu & 0.28 & 0.25 & 0.27 & \textbf{0.32} \\
 & IT\_en\_it\_private & bleu & 0.43 & 0.35 & 0.41 & \textbf{0.51} \\
 & it\_en\_public & bleu & 0.33 & 0.31 & 0.35 & \textbf{0.38} \\
 & IT\_it\_en\_private & bleu & 0.47 & 0.33 & 0.47 & \textbf{0.53} \\
 & UK\_en\_it\_private & bleu & 0.42 & 0.31 & 0.40 & \textbf{0.50} \\
 & UK\_it\_en\_private & bleu & 0.47 & 0.32 & 0.49 & \textbf{0.54} \\
 & US\_en\_it\_private & bleu & 0.33 & 0.24 & 0.30 & \textbf{0.34} \\
 & US\_it\_en\_private & bleu & 0.37 & 0.26 & 0.40 & \textbf{0.43} \\

\midrule
\multirow{2}{*}{MULTI-It} & multi-it-a & accuracy & 0.39 & 0.62 & 0.65 & \textbf{0.84} \\
 & multi-it-c & accuracy & 0.50 & 0.63 & 0.66 & \textbf{0.81} \\

\midrule
\multirow{3}{*}{\textsc{PejorativITy}} & misogyny & accuracy & 0.61 & 0.67 & 0.46 & \textbf{0.75} \\
 & misogyny-context & accuracy & 0.66 & 0.79 & \textbf{0.82} & 0.80 \\ & standard & accuracy & 0.43 & 0.51 & 0.44 & \textbf{0.59} \\

\midrule

\multirow{4}{*}{PERSEID} 
 & task\_0 & f1 & 0.32 & 0.34 & 0.49 & \textbf{0.50} \\
 & task\_1 & f1 & 0.38 & 0.29 & 0.50 & \textbf{0.50} \\
 & task\_2 & f1 & 0.35 & 0.31 & 0.50 & \textbf{0.50} \\
 & task\_3 & f1 & 0.39 & 0.23 & 0.50 & \textbf{0.49} \\

\midrule
\multirow{1}{*}{TERMite} & ita-text-to-sql & exec\_accuracy & 0.04 & 0.38 & 0.36 & \textbf{0.46} \\

\midrule
\multirow{1}{*}{\textsc{TRACE-it}} & traceIT & accuracy & 0.63 & 0.70 & 0.72 & \textbf{0.85} \\

\midrule
\multirow{3}{*}{VERYf-IT} & enriched & accuracy & \textbf{0.57} & 0.43 & 0.52 & 0.56 \\
 & full & accuracy & \textbf{0.59} & 0.41 & 0.52 & \textbf{0.59} \\
 & small & accuracy & \textbf{0.57} & 0.43 & 0.52 & 0.56 \\
 \bottomrule
\label{tab:full-results}
\end{longtable}
\end{footnotesize}

\begin{table}[!ht]
\centering
\footnotesize
\caption{Tasks grouped by category, with counts and lists of top-level tasks and subtasks.
\label{tab:category-parent-task}}
\begin{tabular}{p{1.4cm} r p{10.4cm}}
\toprule
\textbf{Ability} & \textbf{\#} & \textbf{Tasks} \\
\midrule
\multirow{2}{*}{\textbf{Common.}} & 10 & \textsc{AMELIA}, ECWCA, \textsc{EurekaRebus}, GEESE, \textsc{GITA4Calamita}, INVALSI, \textsc{ItaEval}, MACID, MULTI-It, \textsc{PERSEID} \\
\cline{2-3}
& 42 & \scriptsize{amelia-arg-component-fewshot, amelia-arg-component-zeroshot, amelia-arg-premisetype-fewshot, amelia-arg-premisetype-zeroshot, amelia-arg-scheme-fewshot, amelia-arg-scheme-zeroshot, ami\_2020\_aggressiveness, ami\_2020\_misogyny, conflict\_consistency, ecwca-hint, ecwca-no-hint, eureka\_hints, eureka\_original, geese\_anon\_anita, geese\_anon\_dummy, geese\_anon\_gold, geese\_anon\_llama, geese\_noexp, gente\_rephrasing, haspeede2\_hs, haspeede2\_stereo, hatecheck\_ita, honest\_ita, invalsi\_ita, invalsi\_ita\_binarie, invalsi\_ita\_multipla, invalsi\_mate, invalsi\_mate\_multipla, invalsi\_mate\_numero, invalsi\_mate\_verofalso, ironita\_irony, ironita\_sarcasm, macid, multi-it-a, multi-it-c, perse\_task\_0, perse\_task\_1, perse\_task\_2, perse\_task\_3, physical\_verifiability, sentipolc, story\_class\_accuracy} \\
\midrule
\multirow{2}{*}{\textbf{Factual}} & 8 & \textsc{AMELIA}, BEEP, DIMMI, ECWCA, \textsc{EurekaRebus}, INVALSI, MULTI-It, \textsc{VeryfIT} \\
\cline{2-3}
& 29 & \scriptsize{amelia-arg-component-fewshot, amelia-arg-component-zeroshot, amelia-arg-premisetype-fewshot, amelia-arg-premisetype-zeroshot, amelia-arg-scheme-fewshot, amelia-arg-scheme-zeroshot, beep, dimmi-global, dimmi-p1-drug\_interaction, dimmi-p1-molecule, dimmi-p1-posology, dimmi-p1-side\_effect, dimmi-p1-usage,dimmi-p2-drug\_interaction, dimmi-p2-molecule, dimmi-p2-posology, dimmi-p2-side\_effect, dimmi-p2-usage, ecwca-hint, ecwca-no-hint, eureka\_hints, eureka\_original, invalsi\_ita, invalsi\_ita\_binarie, invalsi\_ita\_multipla, invalsi\_mate, invalsi\_mate\_multipla, invalsi\_mate\_numero, invalsi\_mate\_verofalso, multi-it-a, multi-it-c, veryfIT\_enriched, veryfIT\_full, veryfIT\_small} \\
\midrule
\multirow{2}{*}{\textbf{Linguistic}} & 14 & ABRICOT, BLM-It, ECWCA, \textsc{EurekaRebus}, GFG, INVALSI, ITA-SENSE, \textsc{ItaEval}, MACID, MULTI-It, \textsc{PejorativITy}, \textsc{PERSEID}, \textsc{TRACE-it}, \textsc{VeryfIT} \\
\cline{2-3}
& 60 & \scriptsize{abricot\_abs, abricot\_inc, ami\_2020\_aggressiveness, ami\_2020\_misogyny,  blm\_agr1\_0shots, blm\_agr1\_1shots, blm\_agr2\_0shots, blm\_agr2\_1shots, blm\_caus1\_0shots, blm\_caus1\_1shots, blm\_caus2\_0shots, blm\_caus2\_1shots, blm\_od1\_0shots, blm\_od1\_1shots, blm\_od2\_0shots, blm\_od2\_1shots, ecwca-hint, ecwca-no-hint, eureka\_hints, eureka\_original, gente\_rephrasing, gfg\_task\_1\_1, gfg\_task\_1\_2, gfg\_task\_2\_1, gfg\_task\_2\_2, gfg\_task\_2\_3, gfg\_task\_3\_1, gfg\_task\_3\_2, haspeede2\_hs, haspeede2\_stereo, hatecheck\_ita, invalsi\_ita, invalsi\_ita\_binarie, invalsi\_ita\_multipla, invalsi\_mate, invalsi\_mate\_multipla, invalsi\_mate\_numero, invalsi\_mate\_verofalso, ironita\_irony, ironita\_sarcasm, ita-sense-gen-no-translation, ita-sense-gen-with-translation, ita-sense-ml-no-translation, ita-sense-ml-with-translation, itacola, macid, multi-it-a, multi-it-c, pejorativITy-misogyny, pejorativITy-misogyny-context, pejorativITy-standard, perse\_task\_0, perse\_task\_1, perse\_task\_2, perse\_task\_3, sentipolc, traceIT, veryfIT\_enriched, veryfIT\_full, veryfIT\_small} \\
\midrule
\multirow{2}{*}{\textbf{Reasoning}} & 7 & ECWCA, \textsc{EurekaRebus}, GEESE, \textsc{GITA4Calamita}, INVALSI,  MULTI-It, \textsc{TRACE-it} \\
\cline{2-3}
& 22 & \scriptsize{conflict\_consistency, ecwca-hint, ecwca-no-hint, eureka\_hints, eureka\_original, geese\_anon\_anita, geese\_anon\_dummy, geese\_anon\_gold, geese\_anon\_llama, geese\_noexp, invalsi\_ita, invalsi\_ita\_binarie, invalsi\_ita\_multipla, invalsi\_mate, invalsi\_mate\_multipla, invalsi\_mate\_numero, invalsi\_mate\_verofalso, multi-it-a, multi-it-c, physical\_verifiability, story\_class\_accuracy, traceIT} \\
\midrule
\multirow{2}{*}{\textbf{Fairness}} & 3 & GFG, \textsc{ItaEval}, \textsc{PejorativITy} \\
\cline{2-3}
& 16 & \scriptsize{ami\_2020\_aggressiveness, ami\_2020\_misogyny, gente\_rephrasing, gfg\_task\_1\_1, gfg\_task\_1\_2, gfg\_task\_2\_1, gfg\_task\_2\_2, gfg\_task\_2\_3, gfg\_task\_3\_1, gfg\_task\_3\_2, haspeede2\_hs, haspeede2\_stereo, hatecheck\_ita, pejorativITy-misogyny, pejorativITy-misogyny-context, pejorativITy-standard} \\
\midrule
\multirow{2}{*}{\textbf{Code}} & 1 & \textsc{TERMITE} \\
\cline{2-3}
& 1 & \scriptsize{ita-text-to-sql} \\
\midrule
\multirow{2}{*}{\textbf{MT}} & 2 & MAGNET, GFG \\
\cline{2-3}
& 15 & \scriptsize{MAGNET\_IT\_en\_it\_private, MAGNET\_IT\_it\_en\_private, MAGNET\_UK\_en\_it\_private, MAGNET\_UK\_it\_en\_private, MAGNET\_US\_en\_it\_private, MAGNET\_US\_it\_en\_private, MAGNET\_en\_it\_public, MAGNET\_it\_en\_public, gfg\_task\_1\_1, gfg\_task\_1\_2, gfg\_task\_2\_1, gfg\_task\_2\_2, gfg\_task\_2\_3, gfg\_task\_3\_1, gfg\_task\_3\_2} \\
\midrule
\multirow{2}{*}{\textbf{Summar.}} & 2 & GATTINA, \textsc{ItaEval} \\
\cline{2-3}
& 4 & \scriptsize{gattina-ansa, gattina-galileo, news\_sum\_fanpage, news\_sum\_ilpost} \\
\bottomrule
\end{tabular}
\end{table}

\clearpage

\begin{table}[!ht]
\centering
\scriptsize
\caption{Challenges and associated abilities.\label{tab:challenge-abilities}}
\resizebox{\linewidth}{!}{%
\begin{tabular}{lcccccccc}
\toprule
Task & \textbf{Commonsense} & \textbf{Factual} & \textbf{Linguistic} & \textbf{Reasoning} & \textbf{Fairness} & \textbf{Code} & \textbf{MT} & \textbf{Summariz.} \\ \midrule
\textsc{abricot} &  &  & $\checkmark$ &  &  &  &  &  \\
\textsc{amelia} & $\checkmark$ & $\checkmark$ &  &  &  &  &  &  \\
\textsc{beep} &  & $\checkmark$ &  &  &  &  &  &  \\
\textsc{blm} &  &  & $\checkmark$ &  &  &  &  &  \\
\textsc{dimmi} &  & $\checkmark$ &  &  &  &  &  &  \\
\textsc{ecwca} & $\checkmark$ & $\checkmark$ & $\checkmark$ & $\checkmark$ &  &  &  &  \\
\textsc{eurekaRebus} & $\checkmark$ & $\checkmark$ & $\checkmark$ & $\checkmark$ &  &  &  &  \\
\textsc{gattina} &  &  &  &  &  &  &  & $\checkmark$ \\
\textsc{geese} & $\checkmark$ &  &  & $\checkmark$ &  &  &  &  \\
\textsc{gfg} &  &  & $\checkmark$ &  & $\checkmark$ &  & $\checkmark$ &  \\
\textsc{gita4calamita} & $\checkmark$ &  &  & $\checkmark$ &  &  &  &  \\
\textsc{invalsi} & $\checkmark$ & $\checkmark$ & $\checkmark$ & $\checkmark$ &  &  &  &  \\
\textsc{ita-sense} &  &  & $\checkmark$ &  &  &  &  &  \\
\textsc{ItaEval} & $\checkmark$ & &  $\checkmark$ & & $\checkmark$ &  &  & $\checkmark$ \\
\textsc{macid} & $\checkmark$ &  & $\checkmark$ &  &  &  &  &  \\
\textsc{magnet} &  &  &  &  &  &  & $\checkmark$ &  \\
\textsc{mult-it} & $\checkmark$ & $\checkmark$ & $\checkmark$ & $\checkmark$ &  &  &  &  \\
\textsc{pejorativITy} &  &  & $\checkmark$ &  & $\checkmark$ &  &  &  \\
\textsc{perseid} & $\checkmark$ &  & $\checkmark$ &  &  &  &  &  \\
\textsc{termite} &  &  &  &  &  & $\checkmark$ &  &  \\
\textsc{traceIT} &  &  & $\checkmark$ & $\checkmark$ &  &  &  &  \\
\textsc{veryfIT} &  & $\checkmark$ & $\checkmark$ &  &  &  &  &  \\
\bottomrule
\end{tabular}
}
\end{table}


\bigskip

\begin{table}[h]
\centering
\begin{footnotesize}
\caption{Results on the six \textsc{ItaEval} tasks which have been translated from English and are not part of the official CALAMITA benchmark; see Section~\ref{sec:results}, and in particular Section~\ref{sec:itaeval} for details.\label{tab:itaeval-english-results}}
\begin{tabular}{llccccc}
\toprule

\textbf{Task} & \textbf{Metric} & \textsc{M} & \textsc{A} & \textsc{L8B} & \textsc{L70B} \\
\midrule

 arc\_challenge\_ita & accuracy & 0.37 & \textbf{0.53} & 0.41 & \textbf{0.53} \\
 belebele\_ita & accuracy & 0.42 & 0.84 & 0.86 & \textbf{0.92} \\
 hellaswag\_ita & accuracy & 0.45 & 0.51 & 0.45 & \textbf{0.53} \\
 squad\_it & squad\_em & 0.00 & 0.53 & 0.65 & \textbf{0.66} \\
 truthfulqa\_mc2\_ita & accuracy & 0.41 & \textbf{0.68} & 0.51 & 0.54 \\ 
 xcopa\_it & accuracy & 0.74 & 0.74 & 0.72 & \textbf{0.82} \\

\bottomrule
\end{tabular}
\end{footnotesize}
\end{table}

\end{document}